\documentclass[11pt, a4paper, onecolumn, copyright, gdm]{google}

\usepackage{mathmacros}
\usepackage{xspace}

\usepackage{xcolor}
\usepackage{subcaption}

\usepackage[authoryear, sort&compress, round]{natbib}
\uselogo{} 
\newcommand{\schedulename}[1]{\textsc{#1}\xspace}
\newcommand{\tps}{\schedulename{Two-Point Spline}}
\newcommand{\tpl}{\schedulename{Two-Point Linear}}
\newcommand{\cosstd}{\schedulename{Cosine}}
\newcommand{\cosgen}{\schedulename{Generalized Cosine}}
\newcommand{\con}{\schedulename{Constant}}
\newcommand{\snm}{\schedulename{Smooth Non-Monotonic}}
\newcommand{\sqrtdecay}{\schedulename{Square-root Decay}}
\newcommand{\rex}{\schedulename{Generalized Rex}}

\newcommand{\shortschedulename}[2]{\textcolor{#1}{\textsc{#2}}\xspace}
\newcommand{\shortcon}{\shortschedulename{SeaGreen}{con}}
\newcommand{\shortcosstd}{\shortschedulename{orange}{cos-std}}
\newcommand{\shortcosy}{\shortschedulename{orange}{cos-y}}
\newcommand{\shortcosgen}{\shortschedulename{CadetBlue}{cos-gen}}
\newcommand{\shorttps}{\shortschedulename{Thistle}{tps}}
\newcommand{\shorttpsy}{\shortschedulename{Thistle}{tps-y}}
\newcommand{\shorttpl}{\shortschedulename{LimeGreen}{tpl}}
\newcommand{\shortsqrtdecay}{\shortschedulename{Goldenrod}{sqrt}}
\newcommand{\shortrex}{\shortschedulename{Tan}{rex}}
\newcommand{\shortsnm}{\shortschedulename{gray}{snm}}

\newif\ifcomments
\commentstrue %

\ifcomments
\newcommand{\UPDATE}[1]{{\color{magenta} #1}}
\newcommand{\FIX}[1]{{\color{blue}[FIX] #1}}
\newcommand{\TODO}[1]{{\color{magenta}[TODO] #1}}
\newcommand{\aga}[1]{{\color{teal}[AA] #1}}
\newcommand{\PK}[1]{\textcolor{Green}{[PK]: #1}}
\newcommand{\ged}[1]{{\color{Plum}[GD]: #1}}
\newcommand{\hn}[1]{{\color{orange}[HN]: #1}}

\else
\newcommand{\UPDATE}[1]{}
\newcommand{\FIX}[1]{}
\newcommand{\TODO}[1]{}
\newcommand{\aga}[1]{}
\newcommand{\PK}[1]{}
\newcommand{\ged}[1]{}
\newcommand{\hn}[1]{}

\fi

\newcommand{\D}{D}
\newcommand{\T}{T}
\renewcommand{\S}{S} %

\newcommand{\y}{\m{y}}
\newcommand{\z}{\m{z}}

\newcommand{\J}{\m{J}}

\renewcommand{\th}{\sm{\theta}}

\newcommand{\Lo}{\mathcal{L}}
\renewcommand{\P}{P}

\newcommand{\U}{\m{U}}

\newcommand{\ntk}{\hat{\Theta}}

\newcommand{\B}{B}

\newcommand{\Id}{\m{I}}

\newcommand{\lam}{\lambda}

\newcommand{\diag}{{\rm diag}}

\newcommand{\bfr}{\beta}
\newcommand{\pmat}{\m{P}}

\newcommand{\pvec}{\m{p}}
\newcommand{\lmat}{\sm{\Lambda}}

\renewcommand{\S}{\m{S}}

\newcommand{\cifar}{\textsc{CIFAR-10}\xspace}
\newcommand{\wikitext}{\textsc{\mbox{WikiText-103}}\xspace}
\newcommand{\cnn}{\textsc{CNN}\xspace}
\newcommand{\transformer}{\textsc{\mbox{Transformer}}\xspace}

\newcommand{\adamw}{\textsc{AdamW }}

\usepackage{amsmath,amsfonts,bm}

\def\eqref#1{equation~\ref{#1}}

\def\1{\bm{1}}

\DeclareMathAlphabet{\mathsfit}{\encodingdefault}{\sfdefault}{m}{sl}
\SetMathAlphabet{\mathsfit}{bold}{\encodingdefault}{\sfdefault}{bx}{n}

\newcommand{\lr}{\alpha}

\title{What do near-optimal learning rate schedules look like?}

\correspondingauthor{gdahl@google.com}

\reportnumber{} %

\renewcommand{\today}{2026-03-10}

\author[1,2, $\dagger$]{Hiroki Naganuma}
\author[3]{Atish Agarwala}
\author[3]{Priya Kasimbeg}
\author[3]{George E.~Dahl}

\affil[1]{Mila}
\affil[2]{Université de Montréal}
\affil[3]{Google DeepMind\newline $\dagger$ This work was partially done when H.Naganuma was a Student Researcher at Google DeepMind}

\begin{abstract}
A basic unanswered question in neural network training is: what is the best learning rate schedule shape for a given workload? The choice of learning rate schedule is a key factor in the success or failure of the training process, but beyond having some kind of warmup and decay, there is no consensus on what makes a good schedule shape. To answer this question, we designed a search procedure to find the best shapes within a parameterized schedule family. Our approach factors out the schedule shape from the base learning rate, which otherwise would dominate cross-schedule comparisons. We applied our search procedure to a variety of schedule families on three workloads: linear regression, image classification on CIFAR-10, and small-scale language modeling on Wikitext103. We showed that our search procedure indeed generally found near-optimal schedules. We found that warmup and decay are robust features of good schedules, and that commonly used schedule families are not optimal on these workloads. Finally, we explored how the outputs of our shape search depend on other optimization hyperparameters, and found that weight decay can have a strong effect on the optimal schedule shape. To the best of our knowledge, our results represent the most comprehensive results on near-optimal schedule shapes for deep neural network training, to date.
\end{abstract}

\begin{document}
\maketitle

\section{Introduction}

Modern deep neural networks are almost exclusively trained via variants of gradient descent. In these methods, the update to the parameters at each step consists
of the gradient on the current batch (possibly transformed and accumulated with moving averages) multiplied by a scalar known as the \emph{learning rate}. 
Setting an appropriate learning rate is essential for rapid and successful training. At best, an inappropriate learning rate will slow down the training process, and at worst it can cause it to fail entirely.

Varying the learning rate during training according to a schedule can be surprisingly beneficial. A good schedule allows models to reach much lower loss values than any constant value of the learning rate can achieve in a similar number of updates.
Effective schedules in current practice consistently exhibit two high-level commonalities: an initial \emph{warmup} stage where the
learning rate increases from zero (or near-zero) to its peak value, and a later \emph{decay} stage where the learning rate decreases back down to a much smaller value, often near zero.
Analyzing the curvature of the non-linear loss surface and the resulting training dynamics can mostly explain the value of the warmup phase \citep{gilmer2021loss,cohen2024adaptivegradientmethodsedge}, if not give a presciption for the exact shape of the warmup. The need to anneal noise later in training justifies the use of some kind of final decay phase \citep{lee2022trajectory, qiu2025scaling}, but also falls short of a precise characterization of the shape of the decay, or exactly when it should start. More generally there is the question of how long each phase should be, and exactly how to construct a good schedule for a given workload and number of training steps.

Despite the relative consensus on warmup and decay phases being broadly beneficial schedule components, very little is known about exactly what \emph{shape} the learning rate schedule should take.
If the schedule is even tuned at all, in many cases (for example \citet{goyal2017accurate,vaswani2017attention,devlin2019bert}) researchers will only tune a handful of schedule parameters using a fixed functional form for the warmup and decay phases: warm-up duration, peak learning rate, decay start, and final learning rate. Popular choices for the functional form of the decay are linear, inverse square root, or cosine.
It is hard to imagine the optimal shape not being at least somewhat workload dependent, making  guidance on how to find shapes tailored to a given workload desperately needed.

In this work, we take a first step towards characterizing how schedule shapes relate to training outcomes. Towards that end, we defined several (parameterized) learning rate schedule families (Section \ref{sec:schedule_families}), including relatively flexible spline-based families that are capable of covering and extending the typical cosine, linear, and inverse-square-root curves. In Section \ref{sec:search_procedure}, we develop a methodology to search these schedule families on computationally-inexpensive neural network training workloads to find near-optimal shapes within each family. This approach allows us to understand the differences between different schedule families, the benefits of different schedule shapes, and the relationship between the best-performing shapes and specific details of the workload, including other optimizer hyperparameters. In particular, this work makes the following key contributions:
\begin{itemize}
\item We provide the first known optimal schedule for linear regression trained using stochastic gradient descent, and use it to benchmark the efficacy of our search procedure (Section \ref{sec:lin_reg_results}).
\item We provide near-optimal schedules for the different families we defined on two neural network training workloads, a convolutional neural network (\cnn) trained to classify images and a small \transformer language model workload.
The best schedules from all families benefit from warmup and gradual decay, even for families such as \snm that do not enforce these properties
(Section \ref{sec:near_optimal_cifar10_wikitext}).
These results contrast with the linear
regression case where the optimal schedule has no warmup and a sharp decay.
\item We provide multiple lines of evidence that our search experiments were able to adequately explore all but one of our schedule families (Section \ref{sec:validating_near_optimal}).
\item Finally, we show the relationship between optimal schedule shape on our workloads and other hyperparameters, specifically the AdamW $\beta_1$, $\beta_2$, and weight-decay hyperparameters, and that---in particular---weight decay strongly affects the optimal learning rate schedule shape (Section \ref{sec:workload_variations}).
\end{itemize}
Our results represent an important step towards a better understanding of what makes an effective learning rate schedule for neural network training.

\section{Related Work}

From the early days of backpropagation in the 1980s and early 1990s, learning rate tuning, including schedules (although schedules with a warmup phase only became popular later), has been a nuisance for anyone using neural networks. Reviewing all of the various techniques proposed to alleviate this burden is beyond the scope of this work. However, one point bears mentioning. Arguably, removing the learning rate from consideration as a hyperparameter motivated the line of research on per-parameter learning rate algorithms in the early 2010s that eventually culminated with the overwhelmingly popular Adam algorithm \citep{kingma2014adam} (with new variations and extensions still proliferating to this day). Unfortunately, despite the popularity of Adam, it does not obviate the need to tune its global learning rate hyperparameter. Furthermore, the best results with Adam and similar algorithms generally require selecting some kind of non-constant schedule.

As has been common practice for decades when training neural networks, if \emph{any} training hyperparameters are tuned, then likely at least the learning rate will be tuned (for example, \citet{bengio2012practical} describes the learning rate as ``often the single most important hyperparameter''). However, it is important to distinguish the overall scale of the learning from the \emph{shape} of the learning rate schedule. In the presence of a non-constant schedule, the unqualified ``learning rate'' generally refers to the peak, or ``base'' learning rate, and serves as a step-independent scaling factor that, when combined with a shape, gives rise to the absolute learning rate schedule (the function mapping step number to step size). Like any other hyperparameters of the training procedure, base learning rates and schedule shape parameters can, in principle, be tuned using (quasi-)random search \citep{bergstra2012random,bousquet2017critical} or Bayesian optimization \citep{mockus1978bayesopt,snoek2012practical}. In practice, the vast majority of the time, researchers apply a tuned base learning rate to a fixed warmup and decay template, at most tuning the duration of the schedule phases or trying a handful of different discrete functional forms, with the exact details varying from paper to paper. One representative example is \citet{shallueEtAlBatchScience2019}, who predominantly used a linear decay schedule followed by a constant phase (tuning the transition point between the two schedule phases, the base learning rate, and the final learning rate), but also reported trying a handful of other shapes including cosine, exponential, and ``inverse exponential polynomial'' schedules. Some papers have argued for particular shapes, even including periodic ones in \citet{loshchilov2016sgdr} and \citet{smith2017cyclical}, but they tend to be the exception. More commonly, researchers assume a particular shape and study the base learning rate more carefully. For example, \citet{goyal2017accurate} and \citet{shallueEtAlBatchScience2019} are both examples of work that looked at how the base learning rate might depend on batch size and do not investigate any relationship between the optimal schedule shape and the batch size.

When it comes to the schedule \emph{shape} specifically, there are a handful of works that attempt to discover well-performing shapes in various ways, either offline or during the process of training a particular model. Attempts at automating LR shape optimization include AutoLRS \citep{jin2021autolrs}, which dynamically sets learning rate values through Bayesian optimization with a build-in time series model to forecast near-term training progress. \citet{maclaurin2015gradient} proposed an intriguing gradient-based hyperparameter optimization scheme, but, in addition to facing meta-optimization stability and computational cost challenges, their method required careful co-design with the particular optimizer update rule and would require modifications to apply to new optimizers (including Adam). Another gradient-based approach is Hypergradient Descent \citep{baydin2017online}, which uses gradients to adaptively update the learning rate online, during training. Population-Based Training (PBT) \citep{jaderberg2017population} evolves hyperparameters, including the learning rate by pruning and mutating a population of training instances, uncovering retrospectively effective schedules for the particular starting weights. Techniques that discover schedules online during training might find schedules that only perform well on the particular training trajectory they are derived from, instead of reusable schedules that perform well on average over many different initial random seeds or other sources of training pipeline variation. Depending on the circumstances, schedules that depend on a particular training trajectory to be effective could be a positive or negative.

\section{Methods}
\label{sec:methods}

\begin{table}[htbp]
\centering
\caption{Learning Rate Schedule (Shape) Families used in our experiments}
\resizebox{\linewidth}{!}{%
\label{tab:lr_schedule_families}
\begin{tabular}{lll}
\toprule
\textbf{Schedule Family} & \textbf{Short Name} & \textbf{Description} \\
\midrule
\con & \shortcon & Warmup to base LR, followed by a constant LR. \\
\cosstd & \shortcosstd & Warmup to base LR, followed by cosine decay with a fixed exponent = 1. \\
\cosgen & \shortcosgen & Warmup to base LR, followed by cosine decay with a parameterized exponent. \\
\sqrtdecay & \shortsqrtdecay & Warmup to base LR, followed by inverse square root decay. \\
\rex & \shortrex & Warmup to base LR, followed by decay parameterized through REX \citep{chen2022rex}. \\
\tps & \shorttps & Warmup to base LR, followed by decay parameterized through two spline interpolation points. \\
\tpl & \shorttpl & Warmup to base LR, followed by decay parameterized through two linear interpolation points. \\
\snm & \shortsnm & Parameterized by initial LR, step of peak LR, two spline interpolation points, and final LR. \\
\bottomrule
\end{tabular}
}
\end{table}

For our purposes, a learning rate schedule is a function $s(t)$ where $s$ is the learning rate at step $t$. Our goal is to find schedules that are empirically \emph{near-optimal} within some parameterized family of schedule functions, or, in other words, as close to the true optimum $s^*$ as possible while still being within the schedule family.

In practice, the learning rate schedule is often defined as 
$s(t) = \alpha \cdot \phi(t/T)$ where $\alpha$ is the \emph{base learning rate}, $T$ is the training horizon (total number of steps), and $\phi$ is a schedule \emph{shape}\footnote{We will sometimes use the term ``relative schedule'' to refer to the schedule shape in contrast to the actual, or ``absolute'' schedule.}, a function from $[0,1]$ to $[0,1]$.
Note that there are some schedules which are defined without a horizon; we will not consider those here.

\subsection{Learning Rate Schedule Families}
\label{sec:schedule_families}

\begin{figure}[h!]
    \centering
    \begin{subfigure}[b]{\linewidth}
      \includegraphics[width=0.48\linewidth]{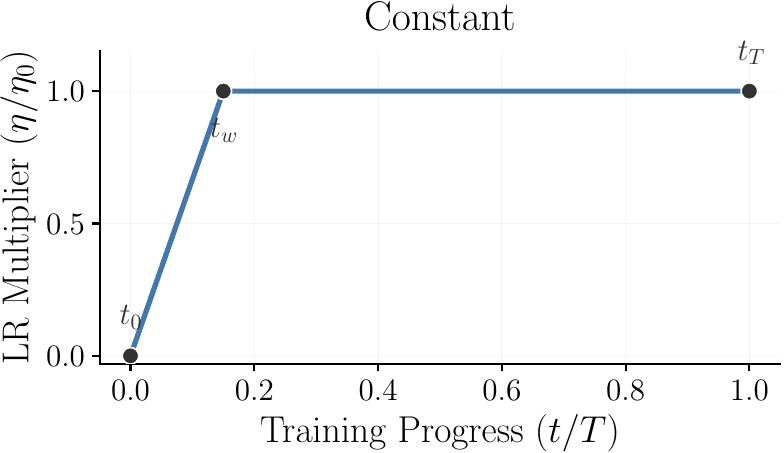}
      \includegraphics[width=0.48\linewidth]{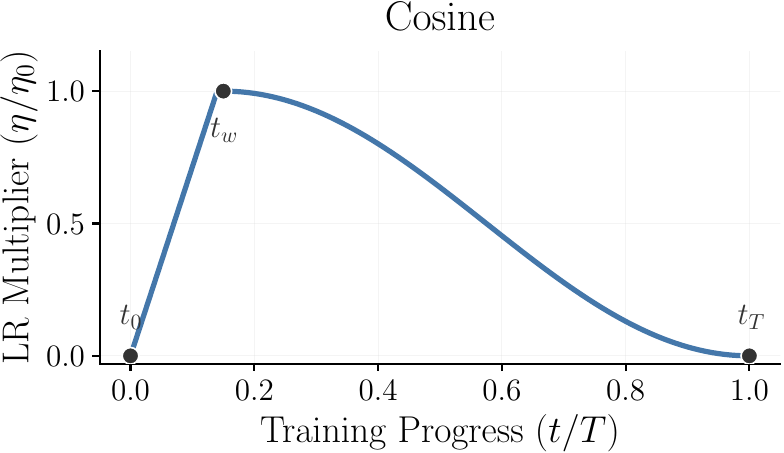}
    \end{subfigure}
    \begin{subfigure}[b]{\linewidth}
      \includegraphics[width=0.48\linewidth]{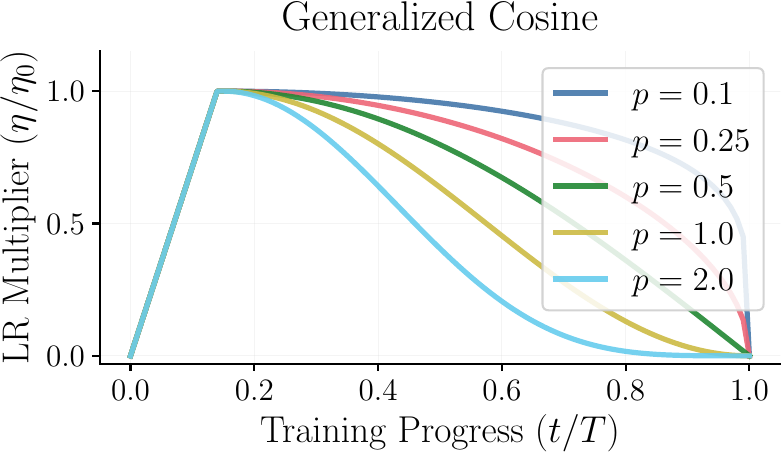}
      \includegraphics[width=0.48\linewidth]{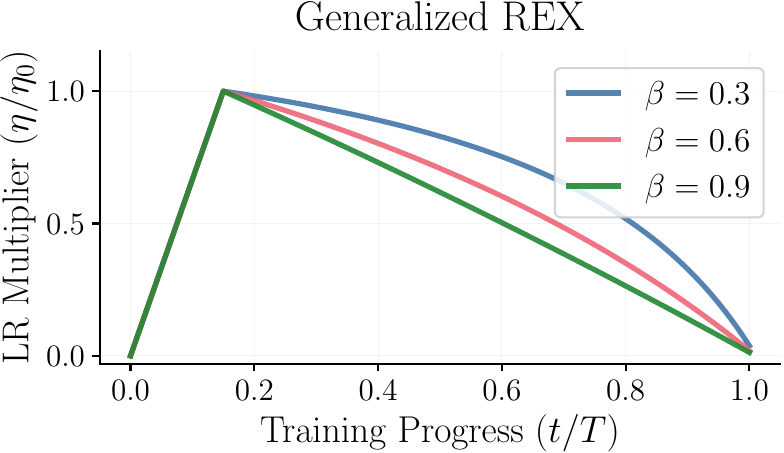}
    \end{subfigure}
    \begin{subfigure}[b]{\linewidth}
      \includegraphics[width=0.48\linewidth]{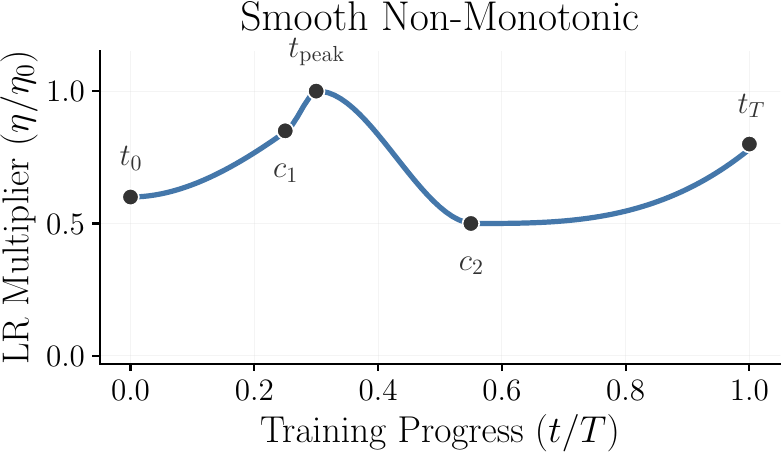}
      \includegraphics[width=0.48\linewidth]{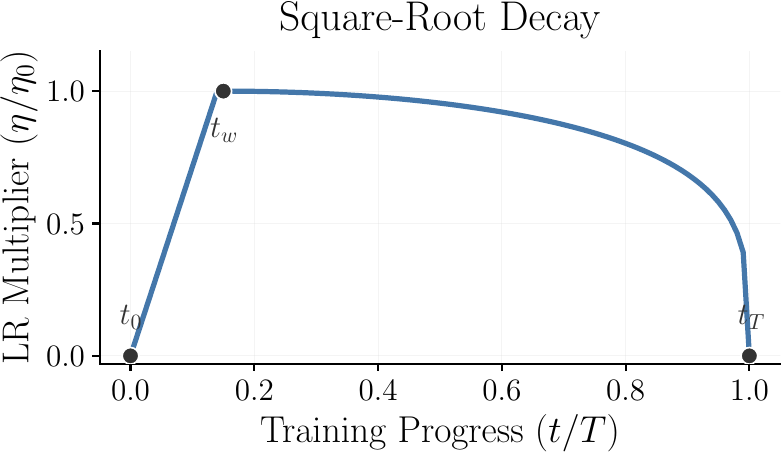}
    \end{subfigure}
    \begin{subfigure}[b]{\linewidth}
      \includegraphics[width=0.48\linewidth]{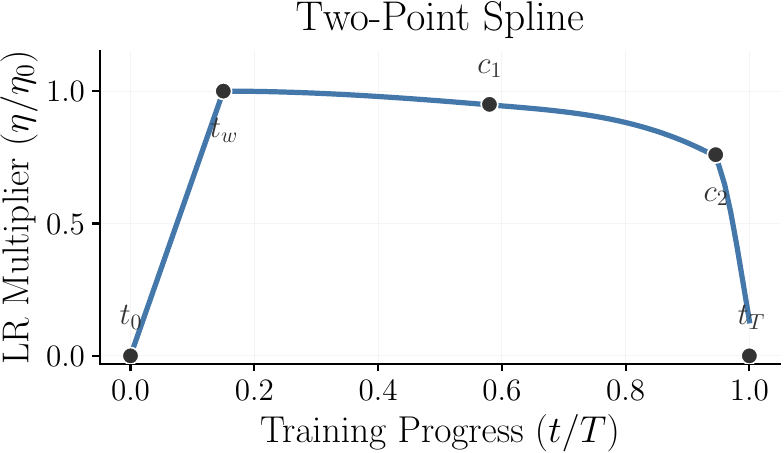}
      \includegraphics[width=0.48\linewidth]{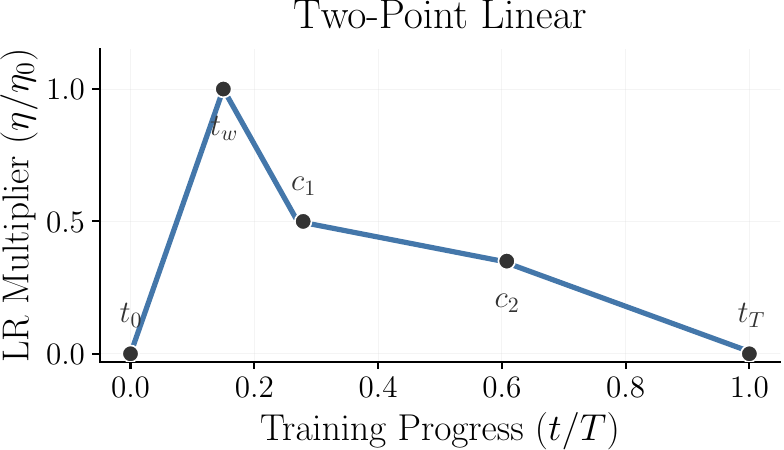}
    \end{subfigure}
    \caption{
    Learning rate schedule families used in our experiments: \con, \cosstd, \cosgen, \rex, \snm, \sqrtdecay, \tps, and \tpl. Markers identify key points such as initial learning rate, warmup completion, intermediate control points, and end of training. Numerical annotations specify parameters particular to each schedule. Peak of \snm can occur in any order compared to control points unlike other schedules.
    }
    \label{fig:schedule_families_v2}
\end{figure}

To constrain the space of learning rate schedules, we defined various
\emph{learning rate schedule (shape) families} (Table \ref{tab:lr_schedule_families}). These are parameterized families of functions from $[0,1]$ to $[0,1]$ (see Appendix \ref{app:schedule_families} for details).
Our \con (\shortcon) schedule family represents shapes without any learning rate decay (but potentially with linear warmup).
The \cosstd (\shortcosstd) schedule family represents the extremely popular decay shape given by $\phi(t/T) = (\cos(\pi f)+1)/2$. To allow for a larger variety of shapes, we also considered the \cosgen (\shortcosgen) family which adds a tunable power $p$ so that $\phi(t/T) = \left[(\cos(\pi f)+1)/2\right]^{p}$.
We also studied some additional monotonically decaying schedules including the \sqrtdecay
(\shortsqrtdecay) and \rex (\shortrex) schedules, which have been studied in  \citet{chen2022rex}.
The most flexible monotonic decay schedules we considered were \tps (\shorttps) and \tpl (\shorttpl), two new families with a decay profile defined using spline interpolation with two control points (not counting the initial peak and the endpoint).

All schedule families mentioned so far also included a linear warmup, since warmup can be useful on many training workloads across many different optimizers, especially when accessing higher peak learning rates is beneficial.
For completeness, we also included the \snm (\shortsnm) schedule family, which can express warmup or decay, but does \emph{not} guarantee it. This schedule family is a completely general two-control-point spline (once again not counting the peak or endpoints) that starts and ends at tunable, non-zero learning rates and has a peak at an arbitrary horizontal position independent of the control points.

\subsection{Workloads and Experimental Setup}

\paragraph{Workloads}
We evaluated learning rate schedules on three different  workloads, where each workload corresponds to a machine learning task defined by a dataset, model, and training objective:

\begin{itemize}
    \item Linear Regression: minimize mean squared error (MSE) on a linear regression problem with prescribed covariance. See Appendix \ref{app:lin_reg_workload_def} for details.
    \item Image Classification: Train a small convolutional neural network (\cnn) on the \cifar dataset \citep{krizhevsky2009learning}.
    \item Small Transformer-based Language Model: Train an 8 million parameter \transformer model \citep{vaswani2017attention} on \wikitext \citep{merity2016pointer}.
\end{itemize}

Details of the model architectures for the \cnn and \transformer are in Appendix \ref{app:model_arch}.
In order to enable high experimental throughput, we purposely selected small, computationally-inexpensive workloads where multiple training runs could fit on a single accelerator.

\paragraph{Optimization-limited Regime and Training Setup}

Given a sufficiently large step budget, most (appropriately) bounded and reasonable schedules will reach a similar minimal loss, erasing the empirical differences in schedule shape performance that we are interested in studying. Even at smaller---but still large---numbers of training steps, our ability to \emph{detect} clear differences between how different schedule shapes performed is limited. Furthermore, a natural goal of studying learning rate schedules is to train \emph{faster} by requiring \emph{fewer} steps to reach the same loss, making it imperative to study regimes where the number of training steps is not quite enough to get the best results.
To highlight more relevant effects and to reduce these measurement issues, we purposefully selected step budgets that afford enough time to make progress on the training objective, but not enough time that most runs saturate.
Training in this \emph{optimization-limited} regime allows us to
distinguish between effective schedules, which achieve better training metrics, and ineffective schedules, which produce worse training metrics.
The total number of training steps and batch sizes were:
\begin{itemize}
    \item Linear regression: 1000 steps, batch size 32.
    \item \cifar: 1000 steps, batch size 256.
    \item \wikitext: 1600 steps, batch size 512.
\end{itemize}

These training step budgets, informed by preliminary experiments, ensure that constant schedules performed somewhat poorly, allowing other schedules to demonstrate clear improvements.
For linear regression experiments, we used SGD. For \cifar and \wikitext experiments,
unless stated otherwise, we used \adamw \citep{kingma2014adam} with standard beta values as $\beta_1 = 0.9$ and $\beta_2 = 0.999$ and weight decay $\lambda_{WD} = 0$. 
Please see Appendix \ref{app:experimental_details} for the details of other hyperparameter settings.

\subsection{Search Procedure}

\label{sec:search_procedure}

At a high level, in order to answer questions about near-optimal schedules, we need a procedure to find the best (or one of the best) schedule shapes within any one of our schedule shape families.
Our search procedure tries to find the schedule shape with the best (lowest) score, which we define as follows. Let $L_{\rm train}^{(r)}(\theta, \alpha, t)$ be the training loss after $t$ steps of training with parameterized shape $\phi_{\theta}(t/T)$, base learning rate
$\alpha$, where $r$ represents sources of randomness in the training process (e.g. weight initializations or data orderings). We define the optimal training loss by:
\begin{align}
\mathcal{J}(\theta,\alpha)
  &:=\operatorname*{median}_{r\sim\mathcal{R}}
      \Bigl[\!\operatorname*{min}_{0\le t\le T}
        L_{\text{train}}^{(r)}(\theta,\alpha,t)\Bigr]
\end{align}
where the median is taken over the distribution $\mathcal{R}$. By using a median, our search will prefer schedule shapes that are likely to perform well across multiple training runs with different initial weights, and tend to avoid any schedules that are hyper-specific to a particular starting point or data ordering.
The optimal shape is the shape $\phi_{\theta^\star}(t/T)$ whose parameters are given by minimizing over all $\theta$ and $\alpha$:
\begin{equation}
\theta^\star
  :=\operatorname*{arg\,min}_{\theta}[\min_{\alpha}
    \mathcal{J}\!\bigl(\theta,\alpha\bigr)].
\end{equation}
We approximate $\theta^\star$ using the two step procedure described below.

\paragraph{Search step.} Our search decouples the search over the schedule parameters from the search over the optimal base learning rate $\alpha$ for a particular shape.
The parameters for each learning rate schedule family were randomly sampled according to distributions detailed in Appendix \ref{app:hyperparameter_search_space}. 
For each parameter setting, we swept over 16 base learning rates from a logarithmically-spaced grid from $10^{-4}$ to $10^{-1}$.
For the \cifar workload, we generated $3600$ shapes for each family except the \snm schedule family where, in some experiments, we generated an additional $36000$ shapes.
For the \wikitext workload, computational constraints limited exploration to around 600 shapes per family. To score a particular schedule during the search, we used $10$ PRNG seeds per schedule on \cifar, and $5$ different seeds per schedule for the initial search on \wikitext.
Each shape was evaluated with 16 base learning rates.
In total, the searches in our experiments involved over $10^{7}$ individual training runs for \cifar, and over $10^{6}$ for \wikitext
---large numbers
enabled by our choice of small, inexpensive workloads.

\paragraph{Evaluation step.} After the search, we performed an additional evaluation round on promising
schedules to better rank them. We first took the top $k$ schedules ranked by median score from the search ($k=100$ for \cifar, and $k=50$ for \wikitext),
and re-trained them with $100$ seeds (all pairwise combinations of $10$ unique initializations and $10$ unique data orderings). We again took the median for final scores,
and used the DKW method \citep{dvoretzky1956} to compute confidence intervals when we needed to.

This two part search and evaluation strategy allowed us to focus computational resources on the most promising schedules. More details on the noise characteristics of this procedure can be found in Appendix \ref{app:noise_character}.

The experimental protocol is described in detail in Appendix \ref{app:experimental_details}; we go over the key points here.

\section{Results}

\begin{figure}[tb]

    \centering
    \begin{subfigure}[b]{0.48\linewidth}
      \includegraphics[height=0.5\linewidth]{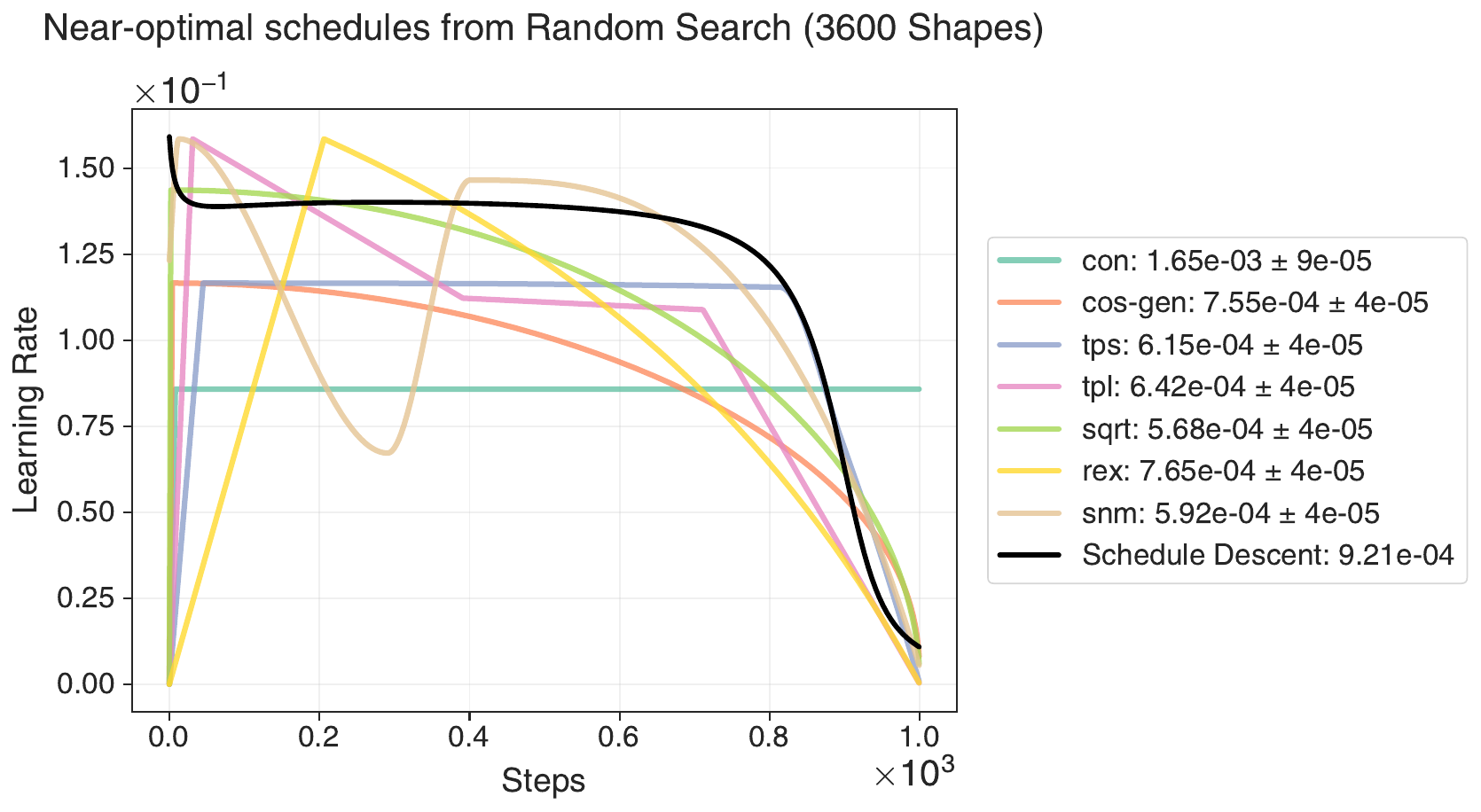}
    \end{subfigure}
    \begin{subfigure}[b]{0.48\linewidth}
      \includegraphics[height=0.5\linewidth]{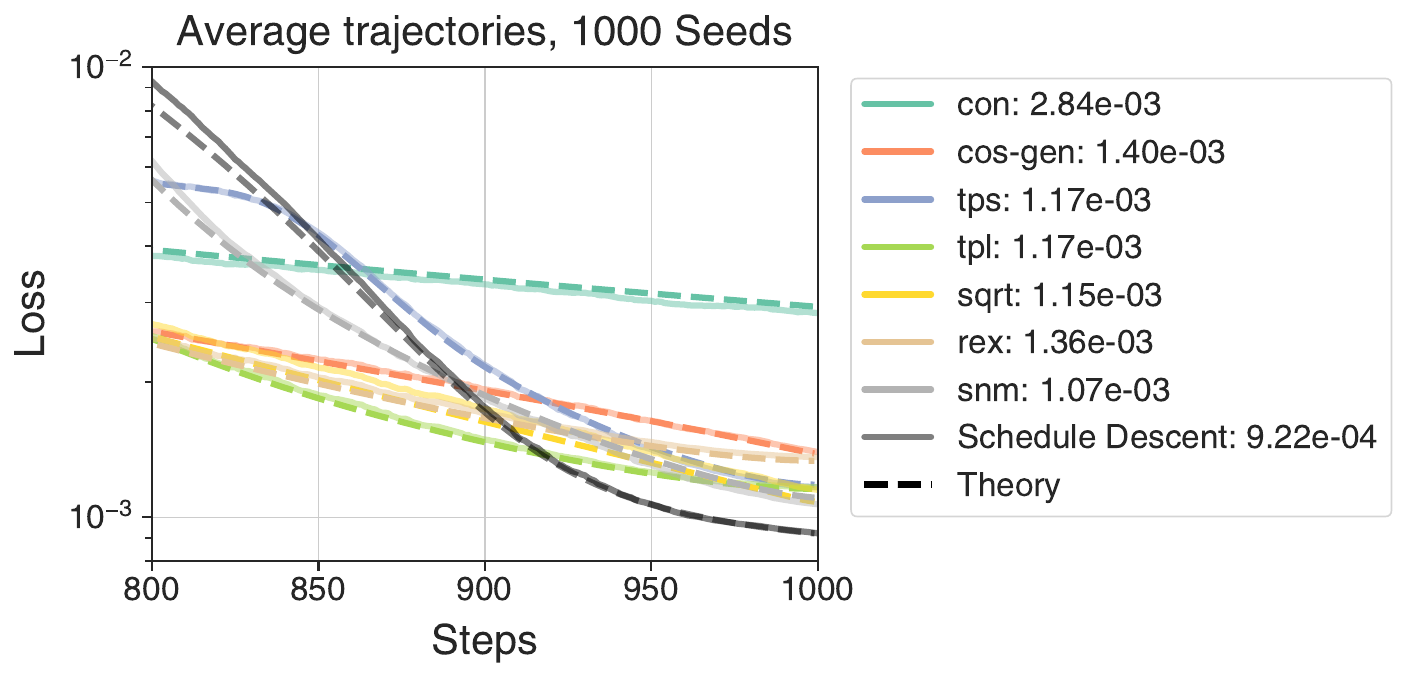}
    \end{subfigure}
    \caption{For a linear regression workload, schedules found via random search capture some of the features of the theoretically optimal schedule but fail to match if completely (left). Average losses appear
    to be better than theoretically optimal; however, when re-evaluating with $1000$ seeds, searched schedules match theoretical prediction and
    are slightly worse than optimal (right).
    }
    \label{fig:best_shapes_lin}
\end{figure}

\subsection{Linear regression: test case with ground truth}

\label{sec:lin_reg_results}

As a test of our experimental protocol, we first evaluated our search methodology on a synthetic linear regression problem with MSE loss.
In this setting (described in detail in Appendix \ref{app:lin_reg_workload}) we define the number of datapoints $\D$, the number of training steps $\T$,
and the data covariance spectrum $\S$, a $\D$ dimensional vector of non-negative numbers. We chose $\S_{k} = \frac{2k}{(\D+1)}$, which corresponds
in the limit of large $D$ to a uniform spectrum $U(0, 1)$, normalized so that $\langle \S_{k}^{2}\rangle = 1$ for convenience (the normalization just
affects the base learning rate of the optimal schedule).

One advantage of this workload is that the final training loss, averaged over random initializations,
can be computed as a relatively simple function of the learning rate schedule in the limit of large $\D$ \citep{agarwala2024high, lee2022trajectory}.
By optimizing this function, we can find the ground truth optimal learning rate schedule which minimizes the average loss at the end of training.
We describe a numerical procedure for finding the optimal schedule in Appendix \ref{app:sched_descent}. The result is an
optimal schedule with decay and no warmup (Figure \ref{fig:best_shapes_lin}, left, black). The schedule is remarkably smooth given that we directly
optimized the $1000$ individual learning rates at each step with no constraints beyond minimizing the final training loss.

We carried out the search procedure across all our schedule families using random search with $3600$ schedule shapes, evaluated with $100$ seeds.
The resulting schedules show some similarities in shape to the ground truth optimal schedule, with very little warmup and decay to near zero
(Figure \ref{fig:best_shapes_lin}, left). This is especially remarkable for the \snm schedule which has no built-in decay.
The more flexible schedule families give better average final loses than the \con family, and at first glance
seem to have better metrics than the theoretically predicted optimal schedule (average and standard error reported in the figure).

\begin{figure}[tb]
    \centering
    \begin{subfigure}[b]{0.31\linewidth}
      \includegraphics[width=\linewidth]{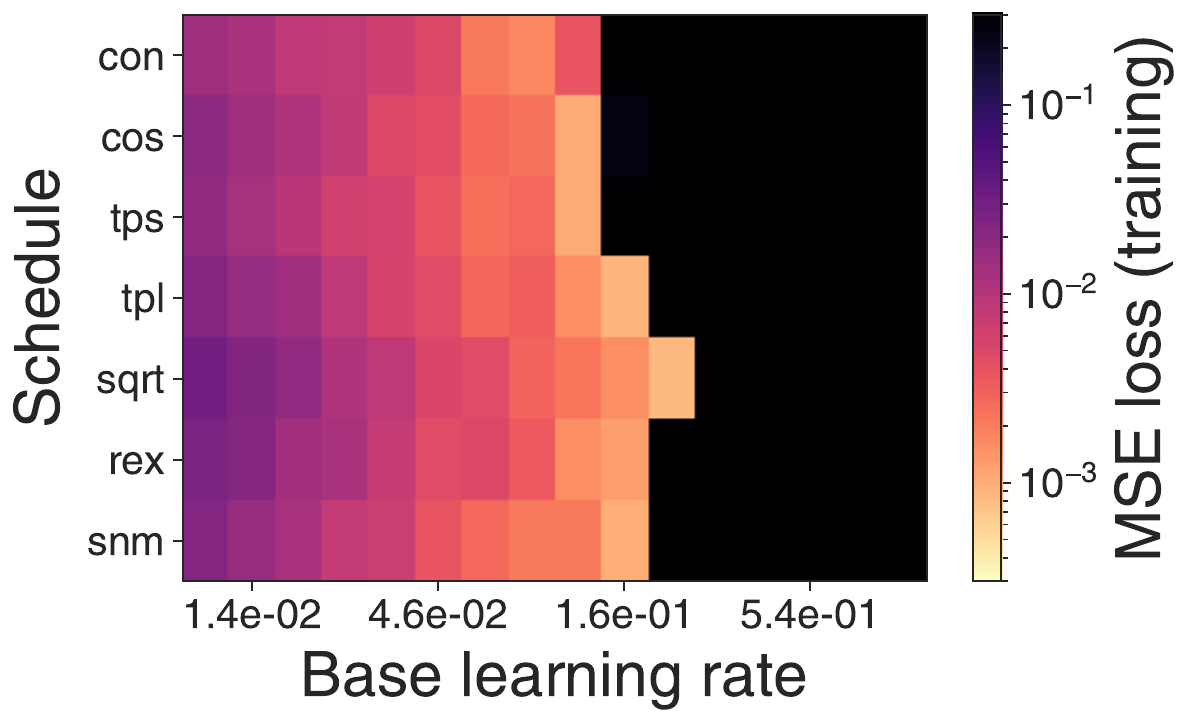}
    \end{subfigure}
    \begin{subfigure}[b]{0.32\linewidth}
      \includegraphics[width=\linewidth]{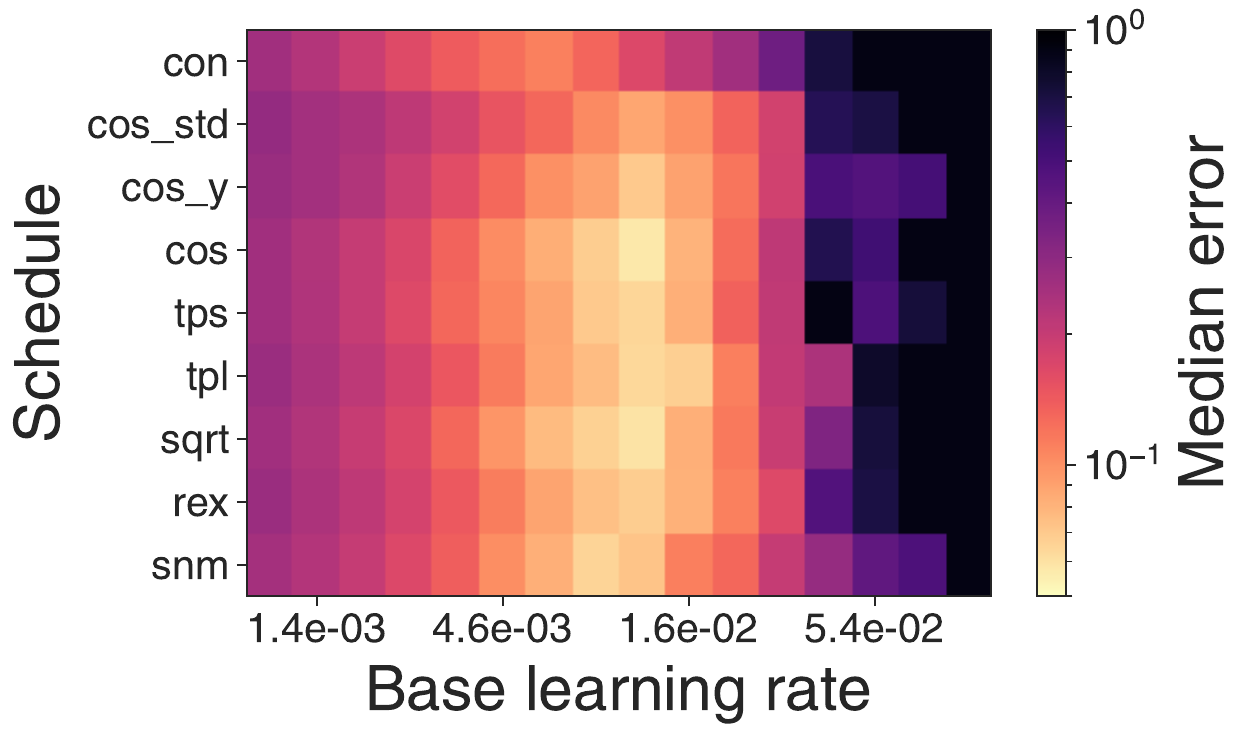}
    \end{subfigure}
    \begin{subfigure}[b]{0.33\linewidth}
      \includegraphics[width=\linewidth]{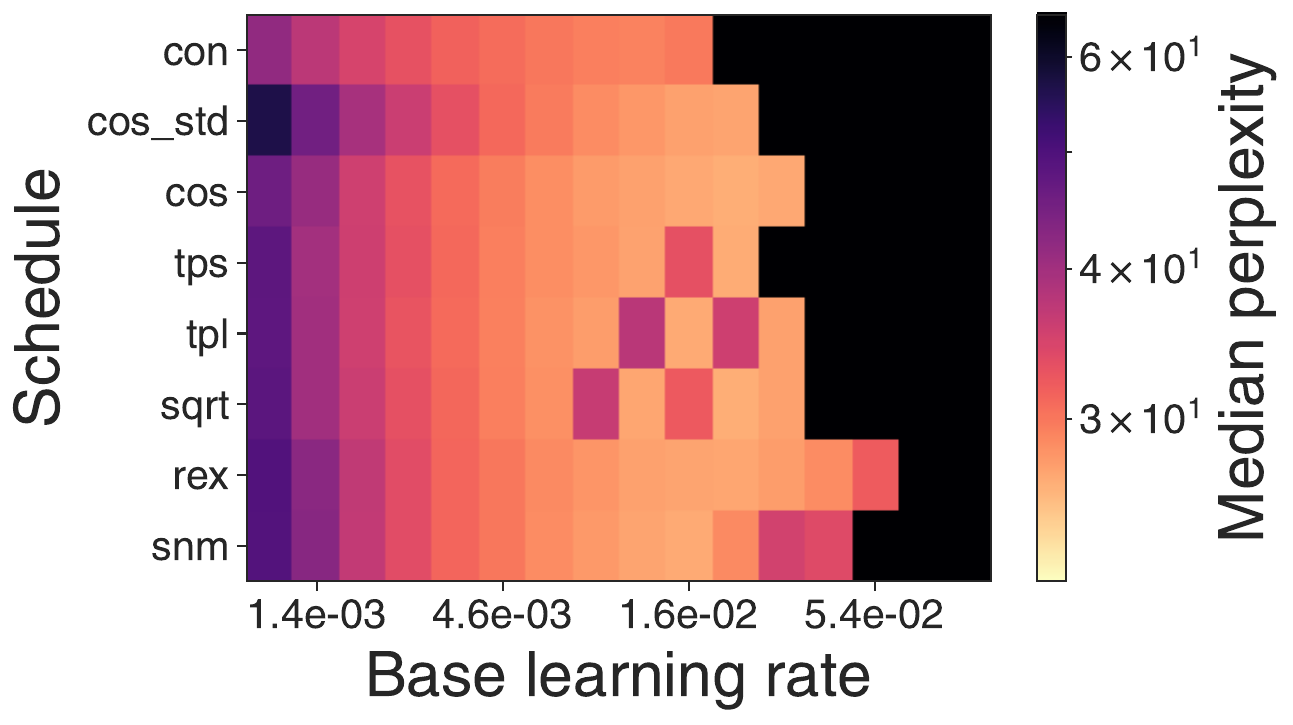}
    \end{subfigure}
    \caption{Training metrics versus base learning rate for best schedules in each family for linear regression (left), \cifar (middle), and \wikitext (right). Each row corresponds to a schedule family, each column to a learning rate, and lighter colors correspond to better performance. Base learning rate is far more important for success than schedule identity, with the exception of the \con schedule which performs worse in all cases.
    }
    \label{fig:base_lr_heatmap}
\end{figure}

However, re-evaluating with more seeds shows that the true average performances are worse than the initial selection experiments suggested,
and that indeed the theoretically optimal learning rate schedule is better than all the others tested (Figure \ref{fig:best_shapes_lin}, right). The relative ordering between the families remains similar, but the absolute performance
is slightly worse than expected. We
hypothesize that this is due to a combination of selection bias (we sampled the best of $3600$ schedules) and the fact that this linear regression
problem is known to have a log-normal distribution of the final loss \citep{agarwala2024high}, making characterization of the deviations more difficult.

We found that the base learning rate is the most important predictor of the success of an (absolute, not relative) schedule (Figure \ref{fig:base_lr_heatmap}, left).
One interesting feature of the linear regression workload is that best performance occurs very close to the edge of stability. We analyze this
phenomenon further in Appendix \ref{ref:app:lin_reg_compare}; by studying the learning curves we can see that the optimal schedules use a large learning
rate at early times that gives short term instability in large eigendirections, but help make progress in small eigendirections; the
learning rate decay at the end of training is used to converge the large eigenmodes.

\begin{figure}[tb]
    \centering
    \begin{subfigure}[b]{0.4\linewidth}
      \includegraphics[height=0.6\linewidth]{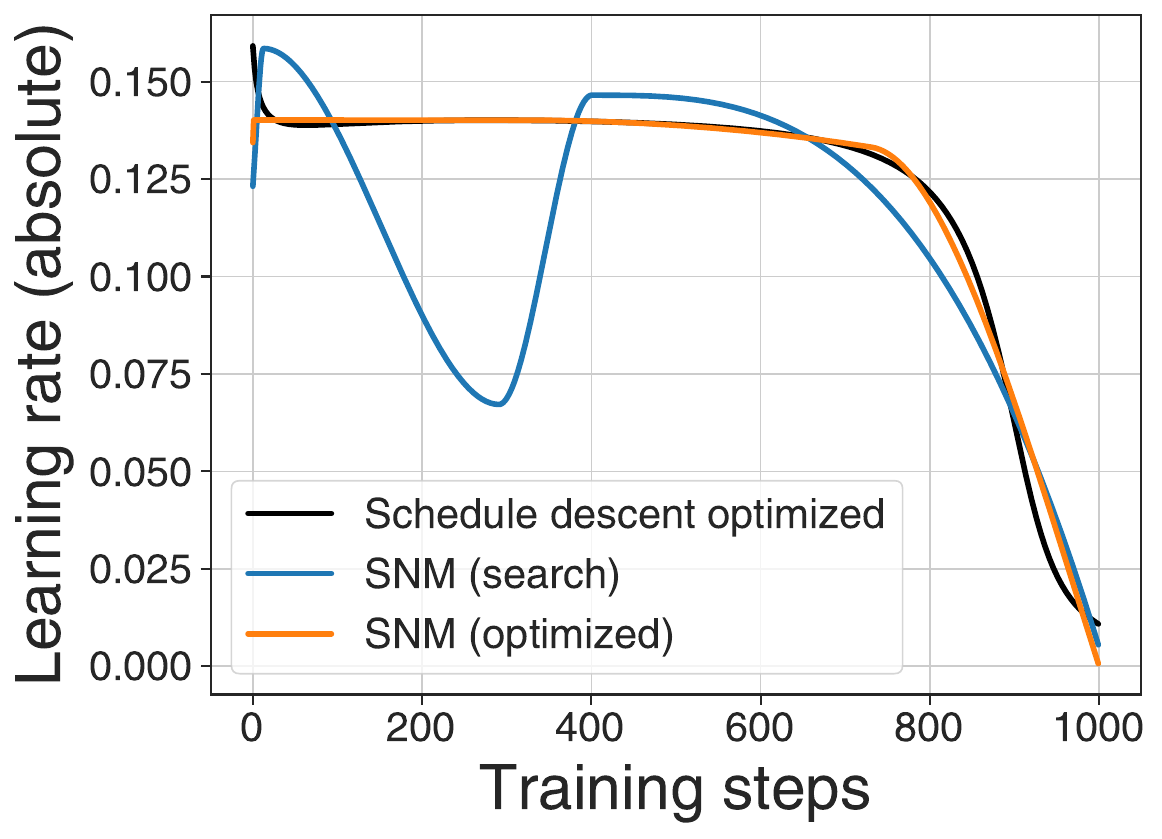}
    \end{subfigure}
    \begin{subfigure}[b]{0.4\linewidth}
      \includegraphics[height=0.6\linewidth]{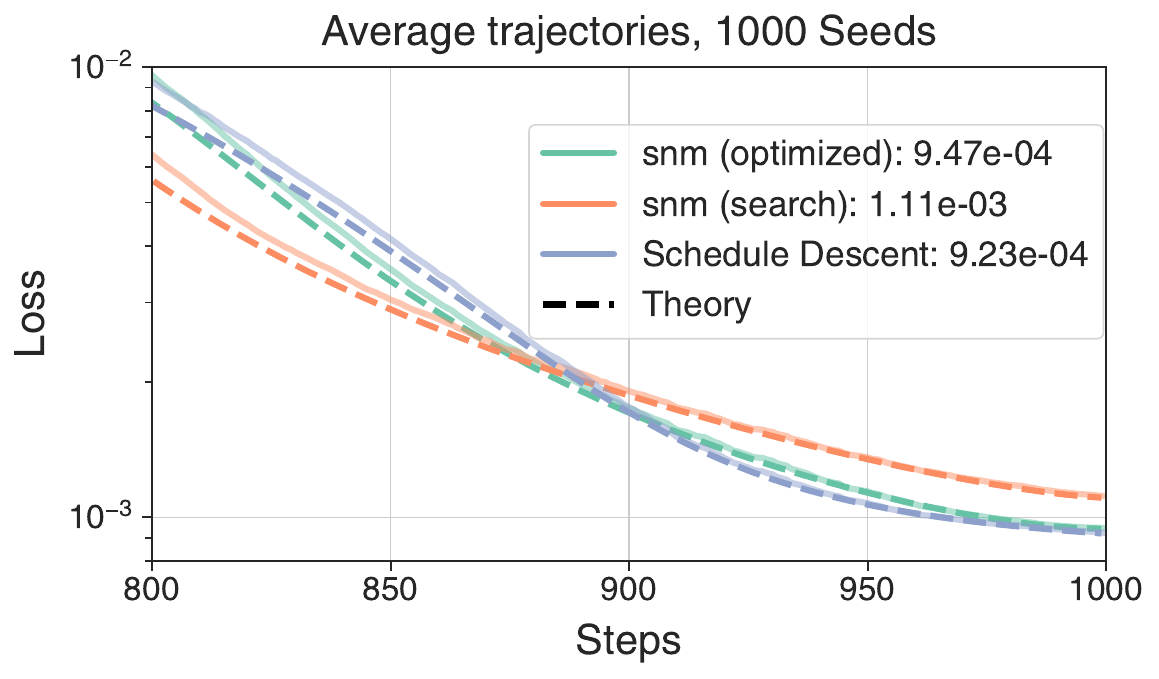}
    \end{subfigure}
    \caption{Best \snm family member does not match optimal schedule (left, blue). We can numerically solve for the best fit curve in the family (left, orange). This obtains very similar performance to the theoretical optimal schedule (right).
    }
    \label{fig:snm_comparison_lin}
\end{figure}

For the linear regression workload our random search methodology retrieves schedules which perform well overall and have some of the features
of the true optimal schedule, but does not actually obtain the best schedules in each family. This can be seen most clearly with the \snm family.
We found the \snm family member that is the closest to the ground truth via numerical optimzation (Figure \ref{fig:snm_comparison_lin}, left). The resulting
schedule is worse than the true optimal schedule by $2\cdot10^{-5}$ only---much less than the difference of $8.5\cdot10^{-5}$ found from the
random search (Figure \ref{fig:snm_comparison_lin}, right).

Overall, our exploration of the linear regression workload suggest that:
\begin{itemize}
    \item Our search procedure can generate schedules that obtain loss values close to optimum.
    \item These shapes capture qualitative and quantitative features of the optimal shape, although may not exactly match the ground truth in every respect.
    \item It is important to validate the efficiency of the search methods for the more flexible schedule families.
\end{itemize}
We use these lessons to guide our analysis of the \cifar and \wikitext workloads in the remainder of this paper.

\subsection{Near-optimal schedules for \cifar and \wikitext workloads}

\label{sec:near_optimal_cifar10_wikitext}

\begin{figure}[tb]
    \centering
    \begin{subfigure}[b]{0.48\linewidth}
    \centering
      \includegraphics[height=0.58\linewidth]{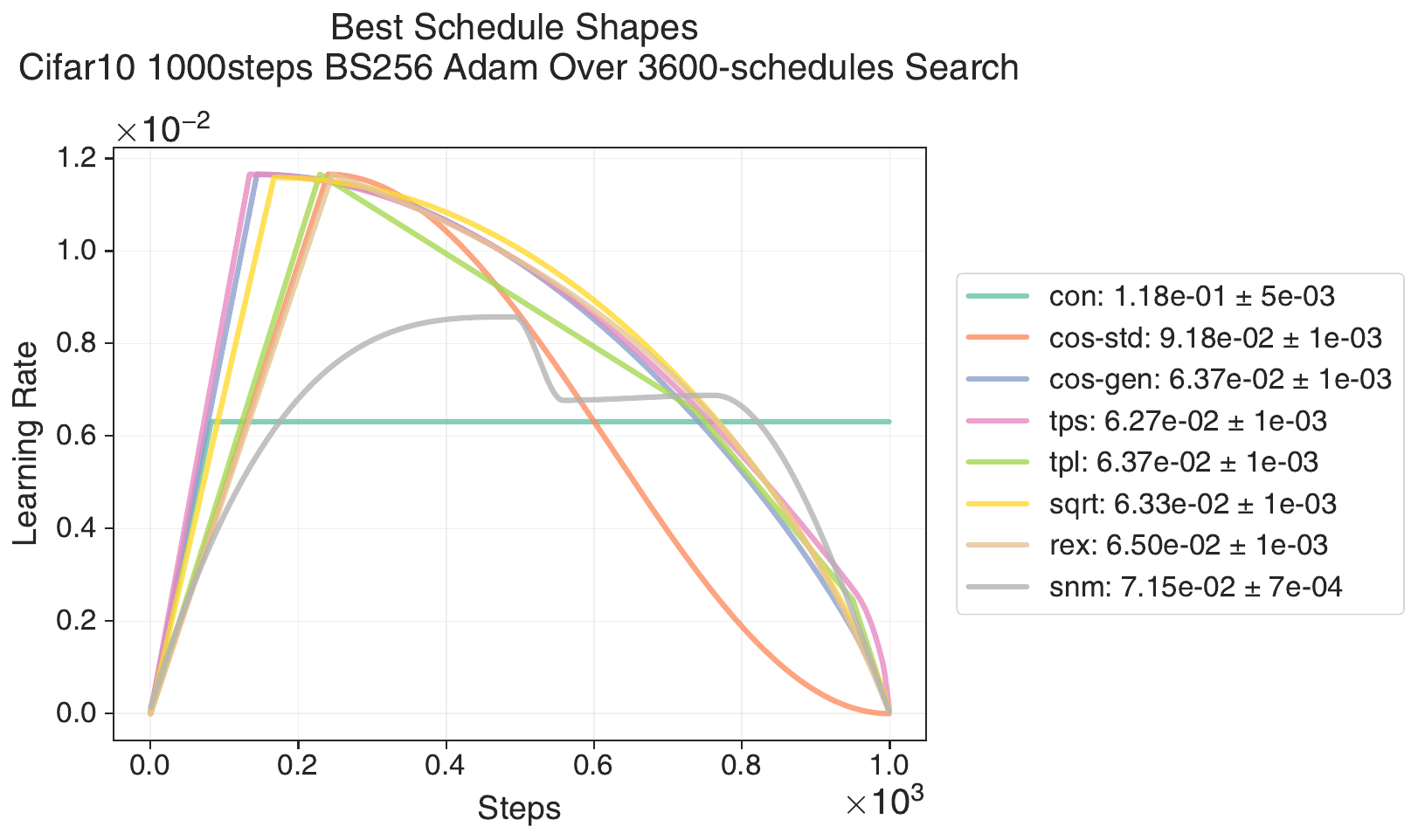}
    \end{subfigure}
    \begin{subfigure}[b]{0.48\linewidth}
    \centering
      \includegraphics[height=0.58\linewidth]{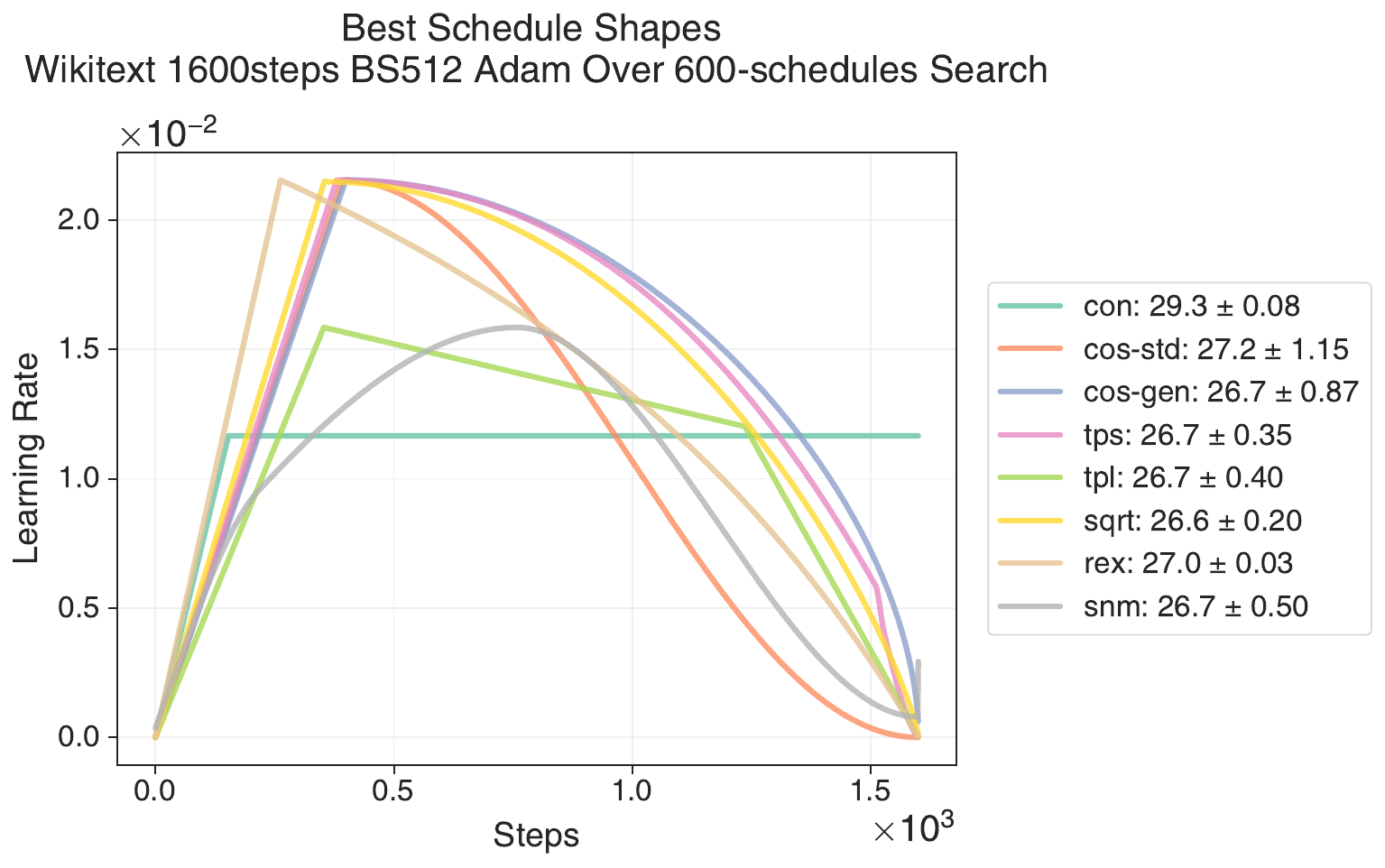}
    \end{subfigure}
    \caption{
    Near optimal learning rate schedules for \cifar (left) and \wikitext (right). The curves represent the best absolute learning rate schedules for each family found in our search procedure which minimized final train error (\cifar) or train perplexity (\wikitext). All curves show similar warmup and decay patterns in each workload,
    including \snm family which does not guaranteed to have those properties. More flexible families perform better than \con and \cosstd.
    }
    \label{fig:best_shapes}
\end{figure}

We carried out the experimental protocol described in Section \ref{sec:methods} on our \cifar and \wikitext workloads, over the schedule families described in Section \ref{sec:schedule_families}. This leads to near-optimal schedules
which have better performance than both warmup+constant and standard cosine (Figure \ref{fig:best_shapes}). We note that the standard error of the
\wikitext experiments tends to be larger than those of the \cifar experiments, in part because some percentage of training runs using
near-optimal schedules become unstable and give very large training perplexities.

We note the following important trends:

\paragraph{Base learning rate is the most important factor for a good schedule.}
Much like in the linear regression case, once a schedule has both warmup and decay,
the base learning rate is a much more important factor in obtaining good performance as compared to the
specific family used (Figure \ref{fig:base_lr_heatmap}, plotted for best schedules for each family).
Note that unlike the linear regression case, for non-linear models it appears that there is a non-trival range of learning rates larger than the optimal, which don't lead to divergent training.
For all families except \con, extra tuning
budget is generally better spent on finetuning base learning rate as opposed to more detailed tuning of schedule hyperparameters.

\paragraph{Warmup and monotonic decay are both crucial.} For both workloads warmup fractions were non-trivial (from $10-30\%$ of total training time), and the decay took up the rest of the time. Most
strikingly, this is true for  the \snm schedule, which does \emph{not have warmup or decay built in} but nontheless ``discovers'' these techniques
via random search. This suggests that warmup and monotonic decay are not just the result of our limited ability to optimize schedules in
expensive workloads, but a fundamental feature of good schedules in deep learning problems.

Indeed, if researchers had carried out a search like ours on a family like \shortsnm in the early days of machine learning research, \emph{they may have discovered warmup and decay earlier}. In particular our \cifar workload is 
similar to training workloads that have existed for decades.

\paragraph{Optimal schedules for deep learning workloads can differ significantly than those for linear regression workloads.}
The optimal schedule for our linear regression workload has no warmup. Additionally, it has a flat, large learning rate for most of training
followed by a sharp decay. This is very different from our two deep learning workloads which benefit from nontrivial warmup and favor gentler
decay.

\paragraph{Flexible families provide small but significant gains.} For \cifar, we found that the more flexible families generally gave shapes that
were significantly different than \cosstd, and achieved lower training error as well: the best flexible families (\tps, \sqrtdecay, \cosgen, \tpl) reached median training errors of $0.063$--$0.064$, compared to $0.092$ for \cosstd (Figure \ref{fig:best_shapes}). For \wikitext, the optimal schedule shapes are different from
\cosstd, with smaller median perplexities ($26.6$--$26.7$ vs.\ $27.2$ for \cosstd), though the larger standard errors make more precise statistical statements difficult (Figure \ref{fig:best_shapes}).

\begin{figure}[tb]
    \centering
    \begin{subfigure}[b]{0.48\linewidth}
    \centering
      \includegraphics[height=0.58\linewidth]{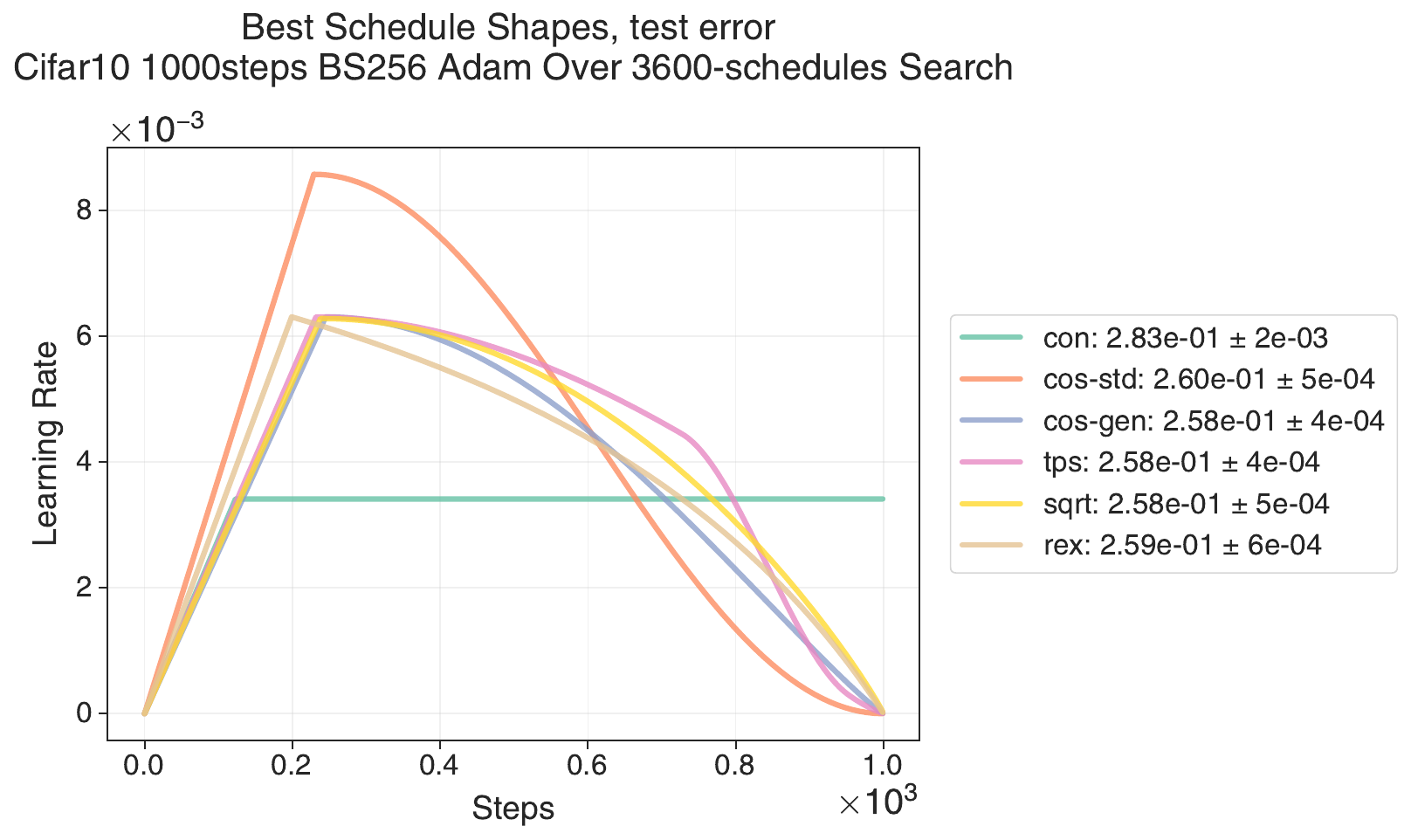}
    \end{subfigure}
    \begin{subfigure}[b]{0.48\linewidth}
    \centering
      \includegraphics[height=0.58\linewidth]{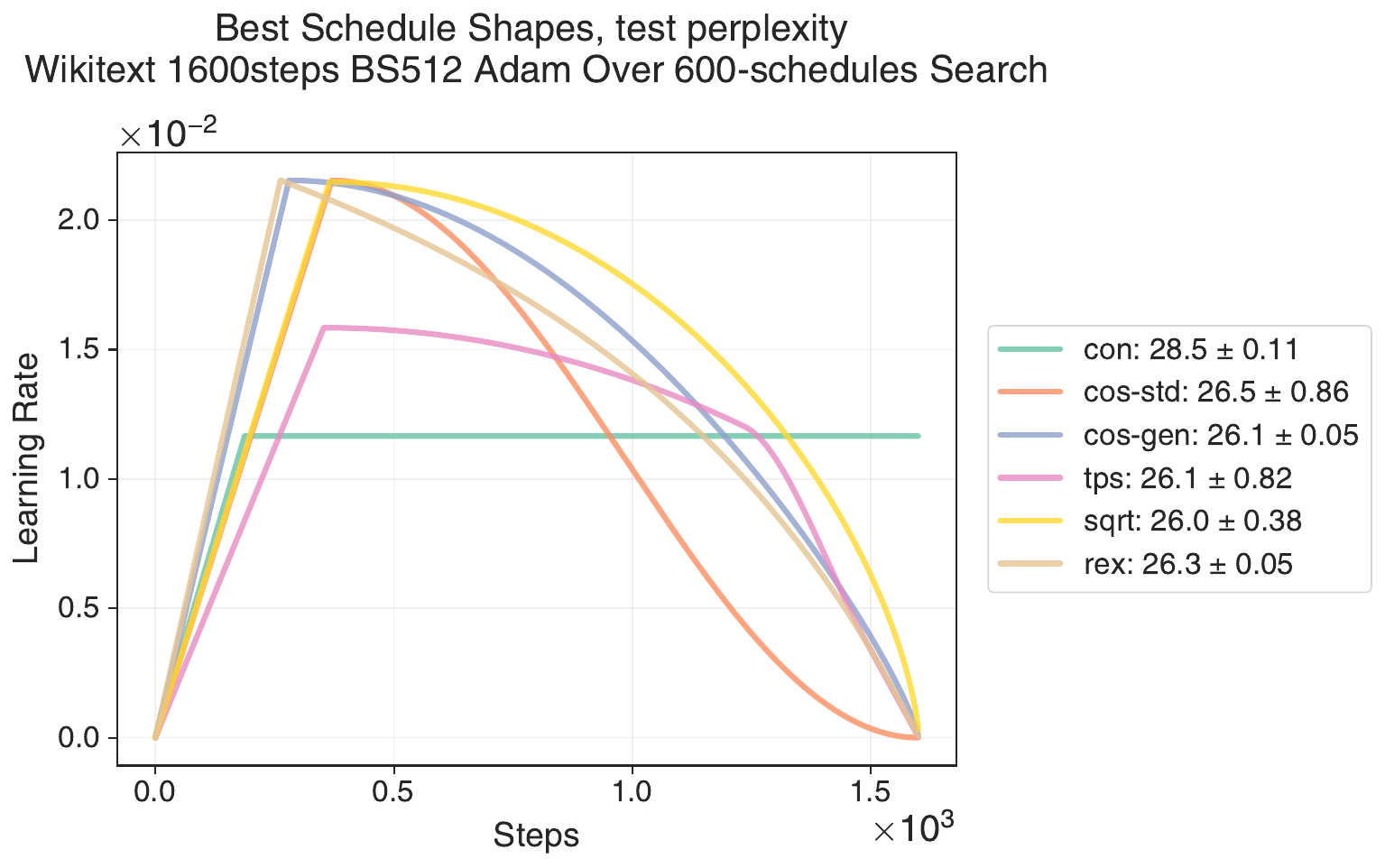}
    \end{subfigure}
    \caption{
     Near optimal learning rate schedules for \cifar (left) and \wikitext (right), selected on best median test  error (\cifar) and test perplexity (\wikitext). Similar trends to selecting on training metrics, with some quantitative differences. \cifar favors training curves with smaller absolute base learning rate when selecting on test error vs train error. \shortsnm and \shorttpl were omitted for clarity.
    }
    \label{fig:best_shapes_test}
\end{figure}

We found similar results when selecting schedules using test metrics (error and perplexity for \cifar and \wikitext respectively),
with somewhat increased variation in schedule shapes and smaller base learning rates in the case of \cifar (Figure \ref{fig:best_shapes_test}).

\subsection{Validating the ``near-optimal'' nature of the search}

\label{sec:validating_near_optimal}

In order for the analysis of the previous section to be meaningful, we must understand: to what extent is our search procedure really obtaining near
optimal schedules? In this section we provide evidence that most of our families are indeed searched well, and that we expect our basic conclusions
to hold even with a better search procedure.

\begin{figure}[tb]
    \centering

    \begin{subfigure}[b]{0.48\linewidth}
    \centering
      \includegraphics[height=0.58\linewidth]{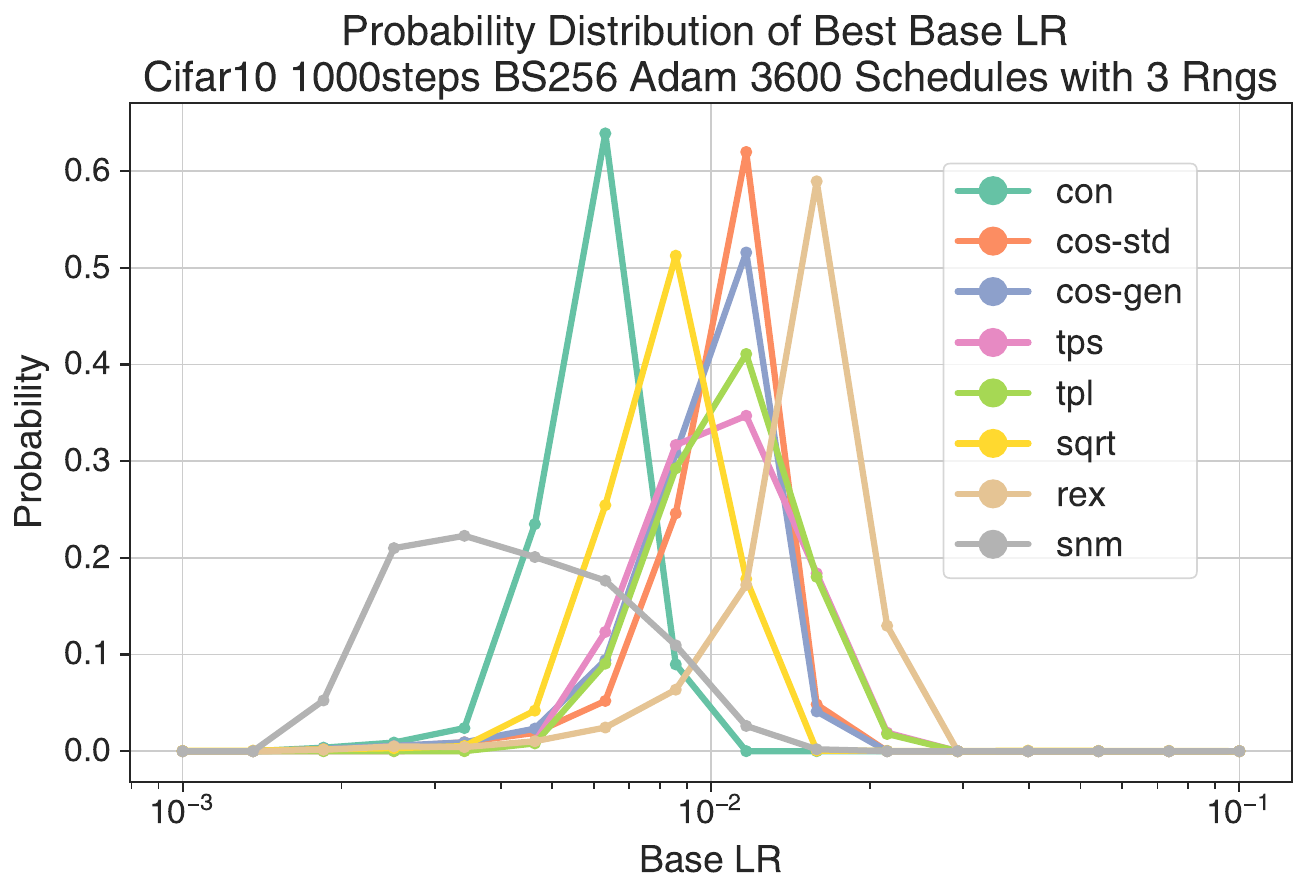}
    \end{subfigure}
    \begin{subfigure}[b]{0.48\linewidth}
    \centering
      \includegraphics[height=0.58\linewidth]{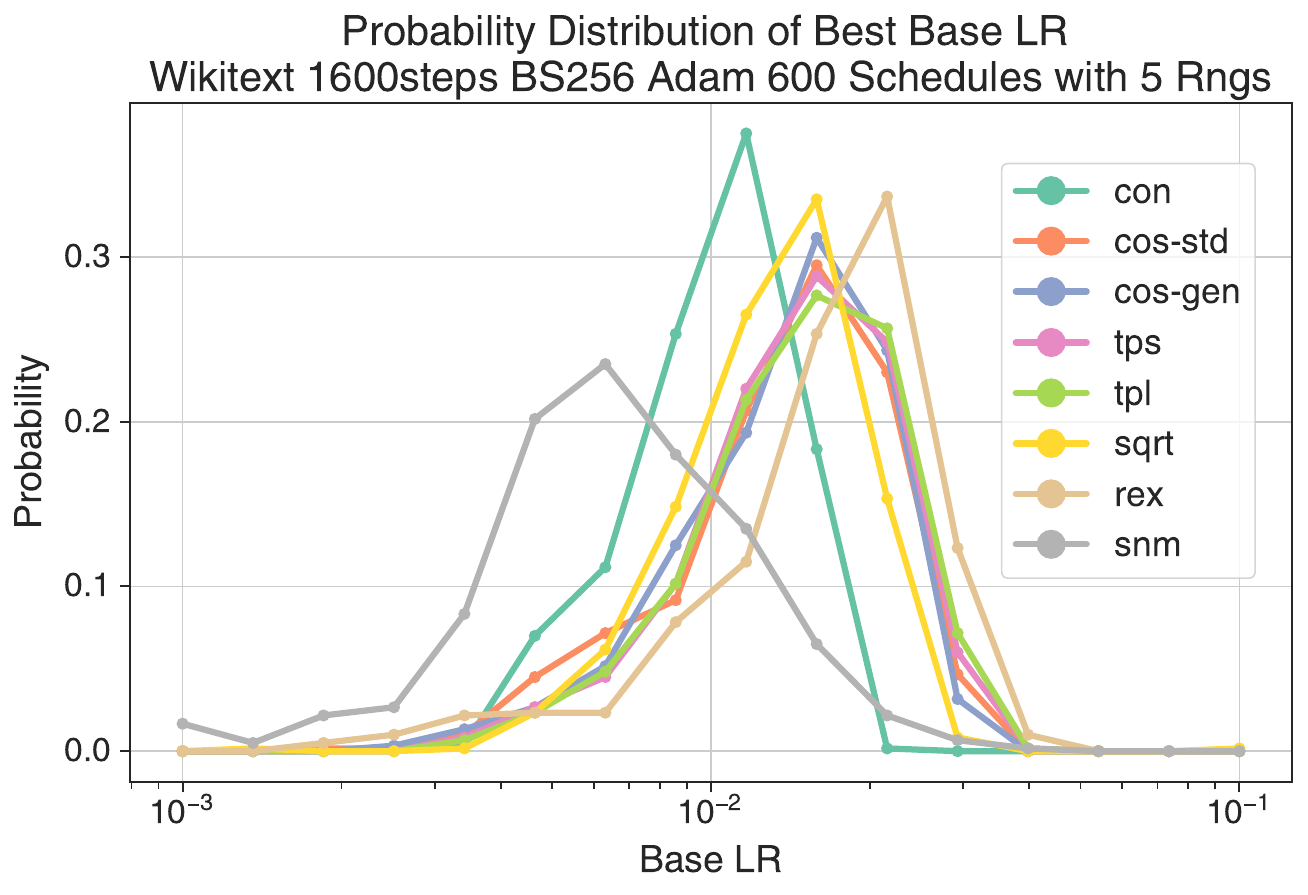}
    \end{subfigure}
    
    \caption{Distribution of best base learning rates for shape search points on \cifar (left) and \wikitext (right). Each curve represents a
    different schedule family. Support for the distributions is mostly contained inside the search grid, which suggests that the base learning rate range was well-chosen. Optimal base learning rate distributions are similar across families, which the exception of \con and \snm which are significantly smaller than the others.
    }
    \label{fig:best_base_lr}
\end{figure}

We noted that the base learning rate is the most important factor in predicting the performance of a learning rate schedule (Figure \ref{fig:base_lr_heatmap}); this is what motivated our decision to
optimize the base learning rate for each family individually. The distribution of optimal learning rates suggests that the range of our
grid search was chosen appropriately for each schedule shape
(Figure \ref{fig:best_base_lr}).

\begin{figure}[tb]
    \centering
    \begin{subfigure}[b]{0.48\linewidth}
    \centering
      \includegraphics[height=0.5\linewidth]{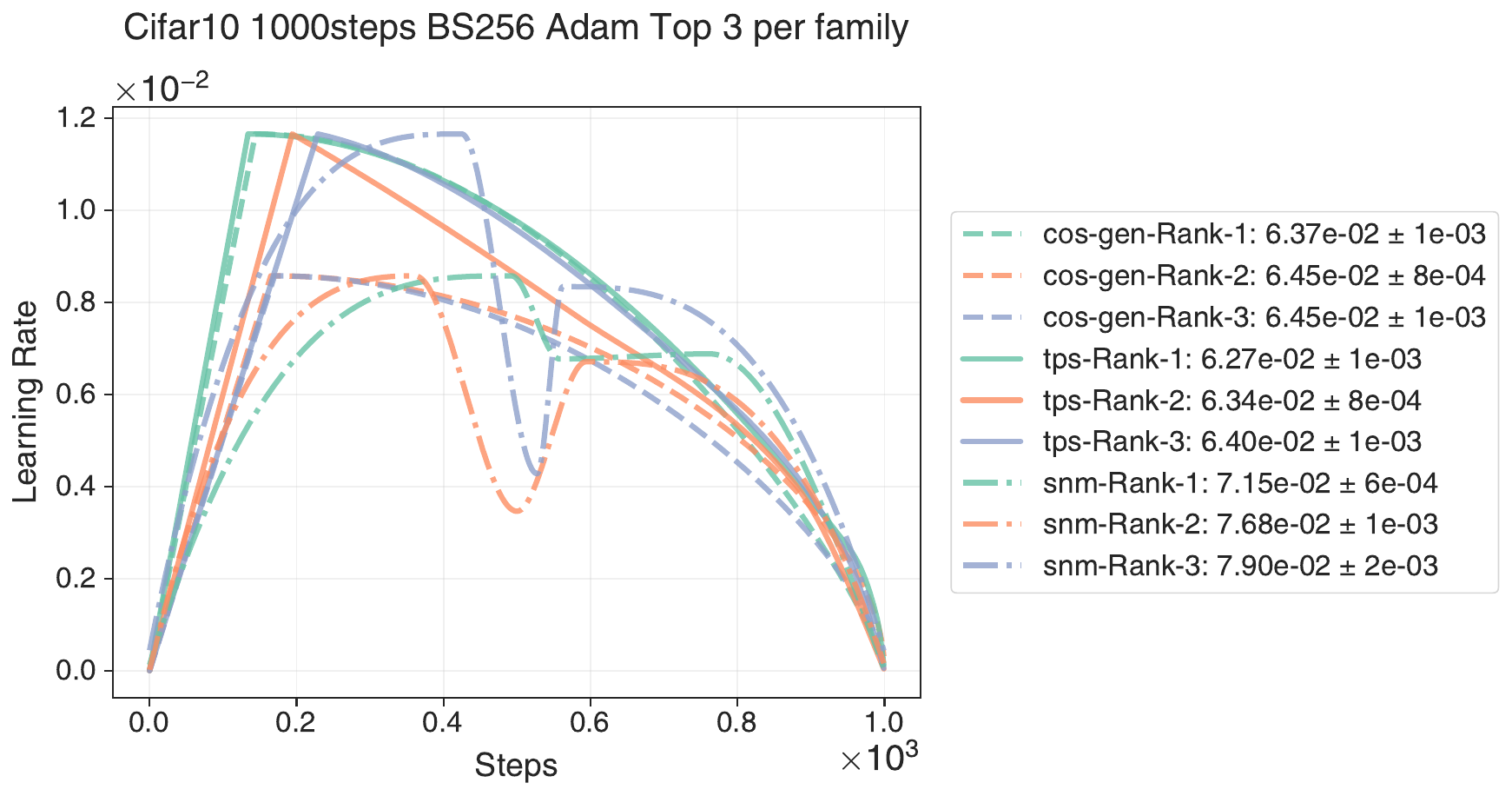}
    \end{subfigure}
    \begin{subfigure}[b]{0.48\linewidth}
    \centering
      \includegraphics[height=0.5\linewidth]{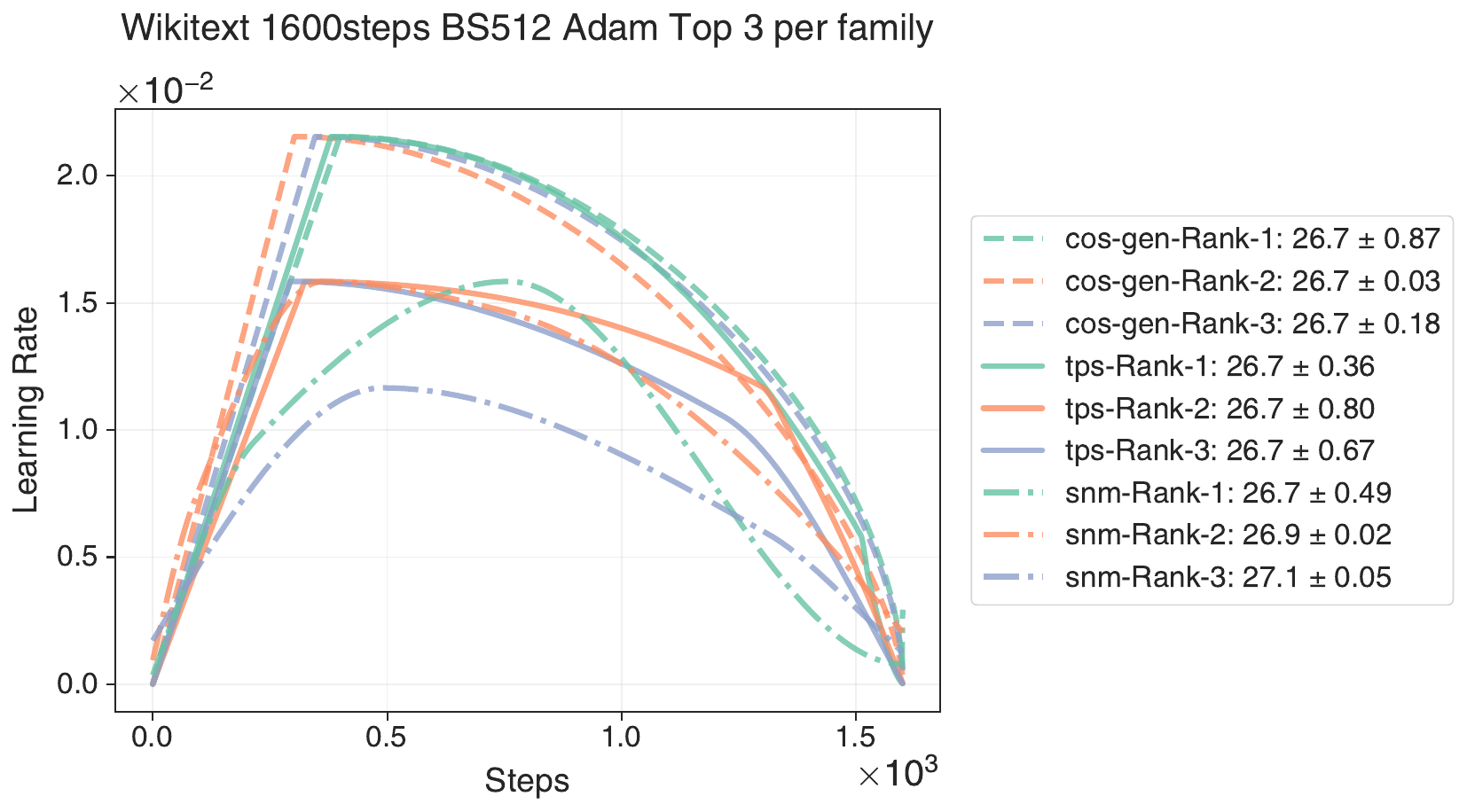}
    \end{subfigure}
    \caption{Top $3$ schedules for \cifar and \wikitext for \cosstd,
    \tps, and \snm families. For \cosstd, schedules are very similar within workload which suggests that \cosstd was searched well. For \cifar, all \tps schedules are similar to each other, while for \wikitext all but one \tps schedules have similar warmup, base learning rate, and decay. This provides evidence that \tps is also generally searched well, but that the \wikitext workload admits a wider variety of learning rate schedules with similar final training metrics. \snm schedules show high variability, suggesting they are not well-searched.
    }
    \label{fig:top_k_family}
\end{figure}

Another relevant question is: how meaningful are the schedule shapes themselves? If the random search was undersampled, then the top shapes per
family may in fact be highly variable. For \cifar, we found that the top shapes were in fact relatively stable; this was true across families
(Figure \ref{fig:best_shapes}, left), but also for the top $3$ shapes within a family (Figure \ref{fig:top_k_family}, left, for \cosstd and \tps).
For the \wikitext workload there is more variability between the best schedule per family (Figure \ref{fig:best_shapes}, right), though there
is broad consistency; there is more consistency within family (Figure \ref{fig:top_k_family}, right, for \cosstd and \tps). The exception in
both workloads is the \snm family, which both deviates from the other families, as well as has high within-family deviation even for the top
schedules (Figure \ref{fig:top_k_family}).

\begin{figure}[tb]
    \centering
    \begin{subfigure}[b]{0.9\linewidth}
      \includegraphics[width=\linewidth]{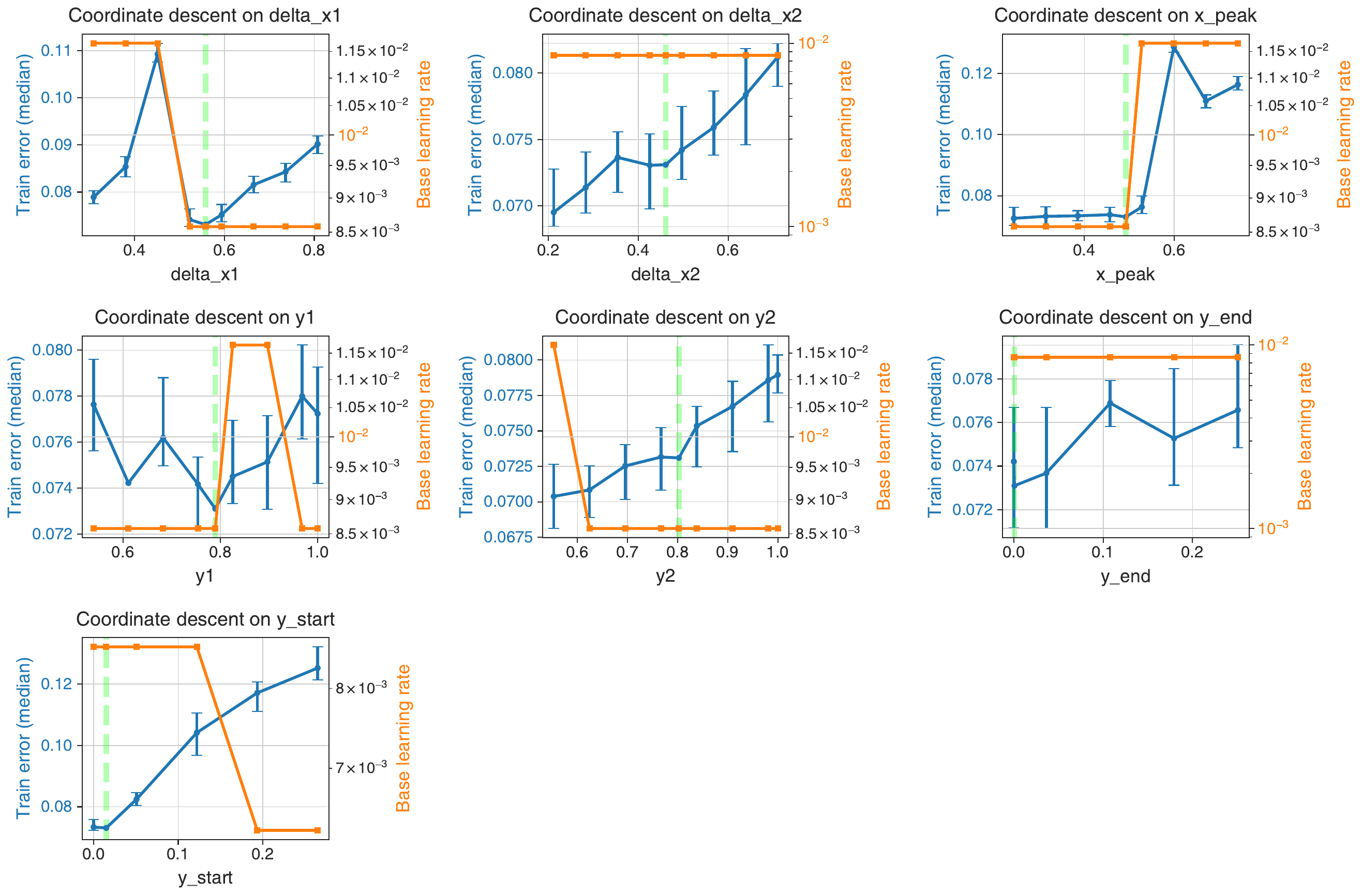}
        \caption{\cifar}
    \label{fig:snm_coordinate_descent_cifar}
    \end{subfigure}
    \begin{subfigure}[b]{0.9\linewidth}
      \includegraphics[width=\linewidth]{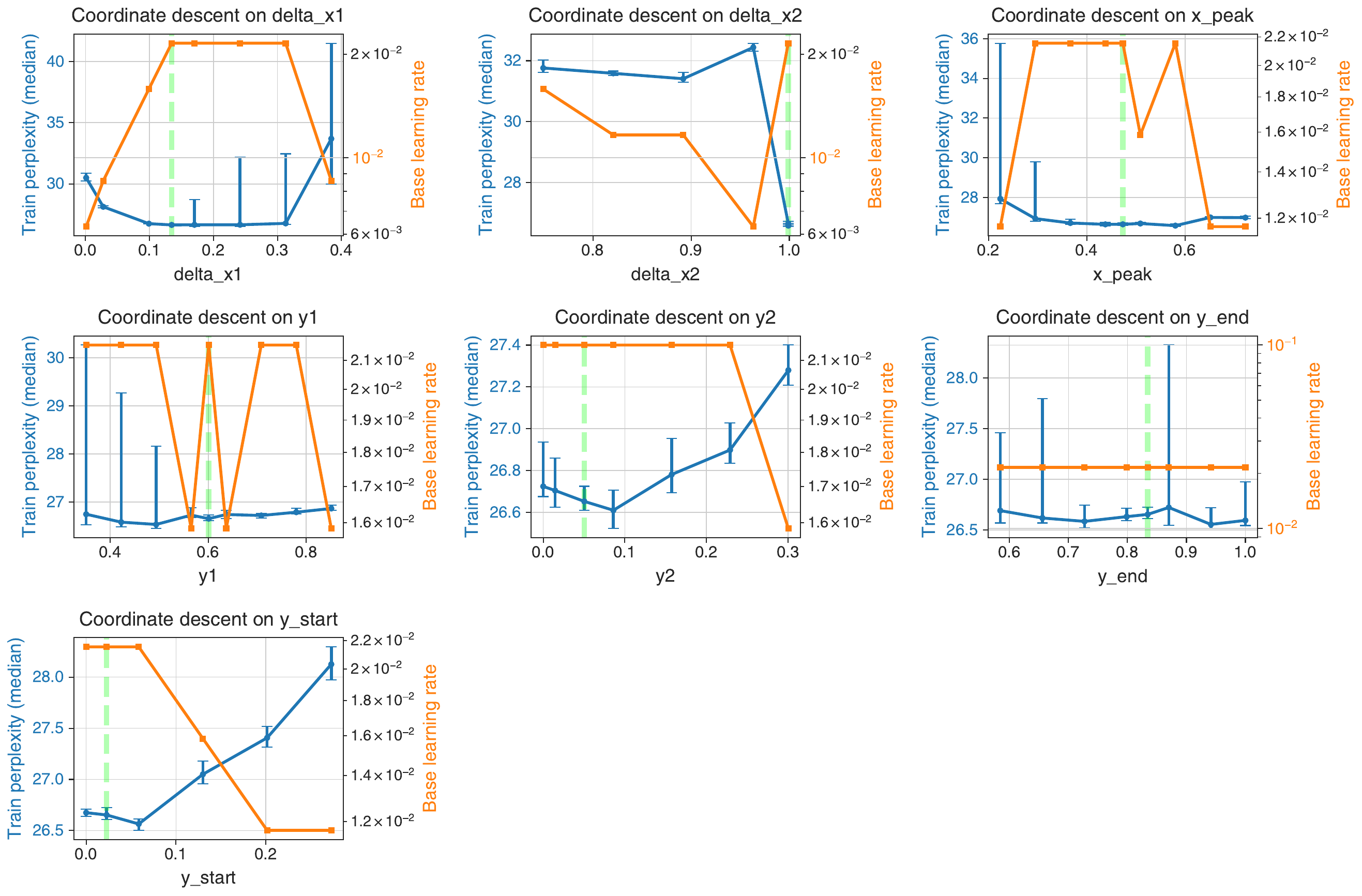}
        \caption{\wikitext}
    \label{fig:snm_coordinate_descent_wikitext}
    
    \end{subfigure}
    \caption{Coordinate-wise linesearch from optimal \snm schedule shape for \cifar (top 3 rows) and \wikitext (bottom 3 rows). For each
    coordinate, we vary the coordinate and remeasure the appropriate training metric (blue, 95\% CI plotted), with a new base learning rate optimization (best base LR in orange). Green dashed line marks the
    original value of the coordinate from the search. Varying some coordinates, particularly $y_{2}$, can significantly improve
    training metrics suggesting that \snm is not well-optimized.
    }
    \label{fig:snm_coordinate_descent}
\end{figure}

\begin{figure}[tb]
    \centering
    \begin{subfigure}[b]{0.9\linewidth}
      \includegraphics[width=\linewidth]{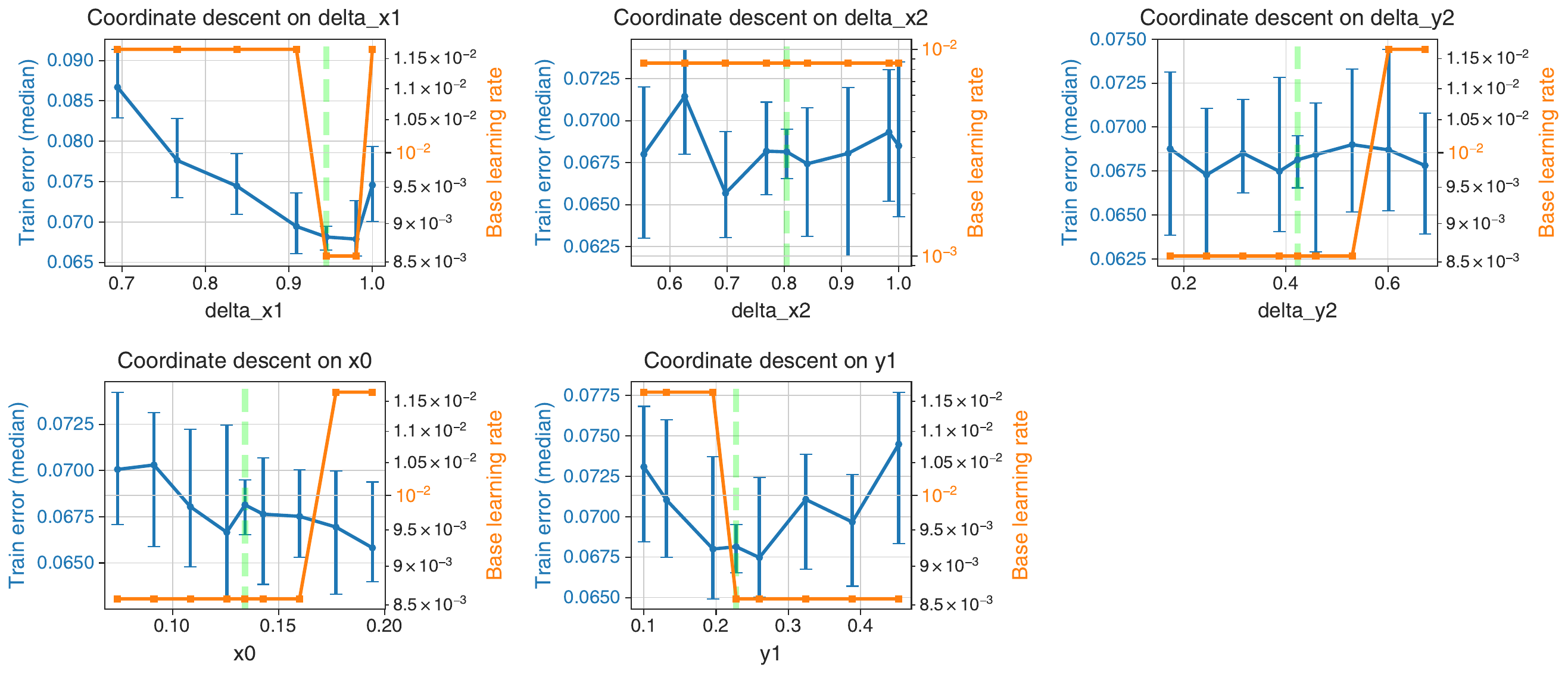}
      \caption{\cifar}
    \end{subfigure}
    \begin{subfigure}[b]{0.9\linewidth}
      \includegraphics[width=\linewidth]{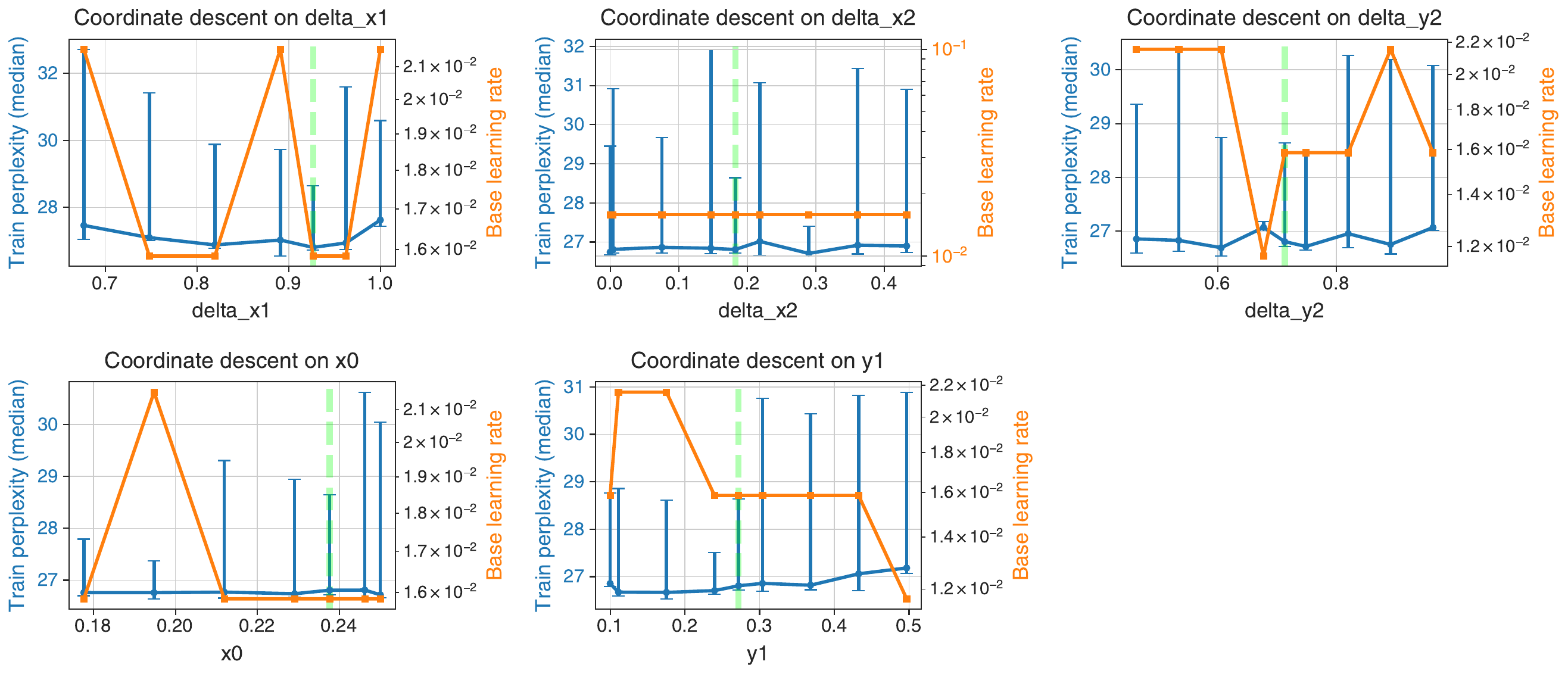}
      \caption{\wikitext}
    \end{subfigure}
    \caption{Coordinatewise linesearch for \tps schedule. Methodology and labeling are the same as Figure \ref{fig:snm_coordinate_descent}. Gains
    from linesearch are generally smaller than measurement uncertainty, suggesting \tps is well-optimized at the resolution of our experiments --- unlike \snm.
    }
    \label{fig:tps_coordinate_descent}
\end{figure}

The poor optimization of the \snm family can be seen directly by performing linesearches of schedules in different coordinate directions. For a
near-optimal schedule, perturbing any individual parameter would not give improvements to the training metrics; for non-optimal schedules, at least one
perturbation direction would give clear improvements.
We carried out such a search on the best \snm and \tps schedules for both our workloads. We varied one schedule parameter at a time, reoptimized base learning rate, using $100$ seeds for each coordinate searched.
We found that for the \snm schedule, for \cifar there were two coordinates -- $\Delta x_{2}$ and $y_{2}$---which could be varied to significantly improve the final train error (Figure \ref{fig:snm_coordinate_descent} (a)). For \wikitext, $y_2$ was also likely suboptimal, and $y_{start}$ may be as well;
detailed analysis is made more difficult by the larger confidence intervals and smaller improvements available for \wikitext (Figure \ref{fig:snm_coordinate_descent} (b)). In contrast for the best \tps schedule, only the coordinate $x_{0}$ was measurable suboptimal, and even
there the best value found was within the $95\%$ CI of the original point
(Figure \ref{fig:tps_coordinate_descent}). This provides a more quantitative insight into how ``near-optimal'' our best \tps schedule is, and show
directly that the \snm schedule is far less optimized than \tps and other schedules.

\begin{figure}[tb]

    \begin{subfigure}[b]{0.49\linewidth}
    \centering
      \includegraphics[height=0.5\linewidth]{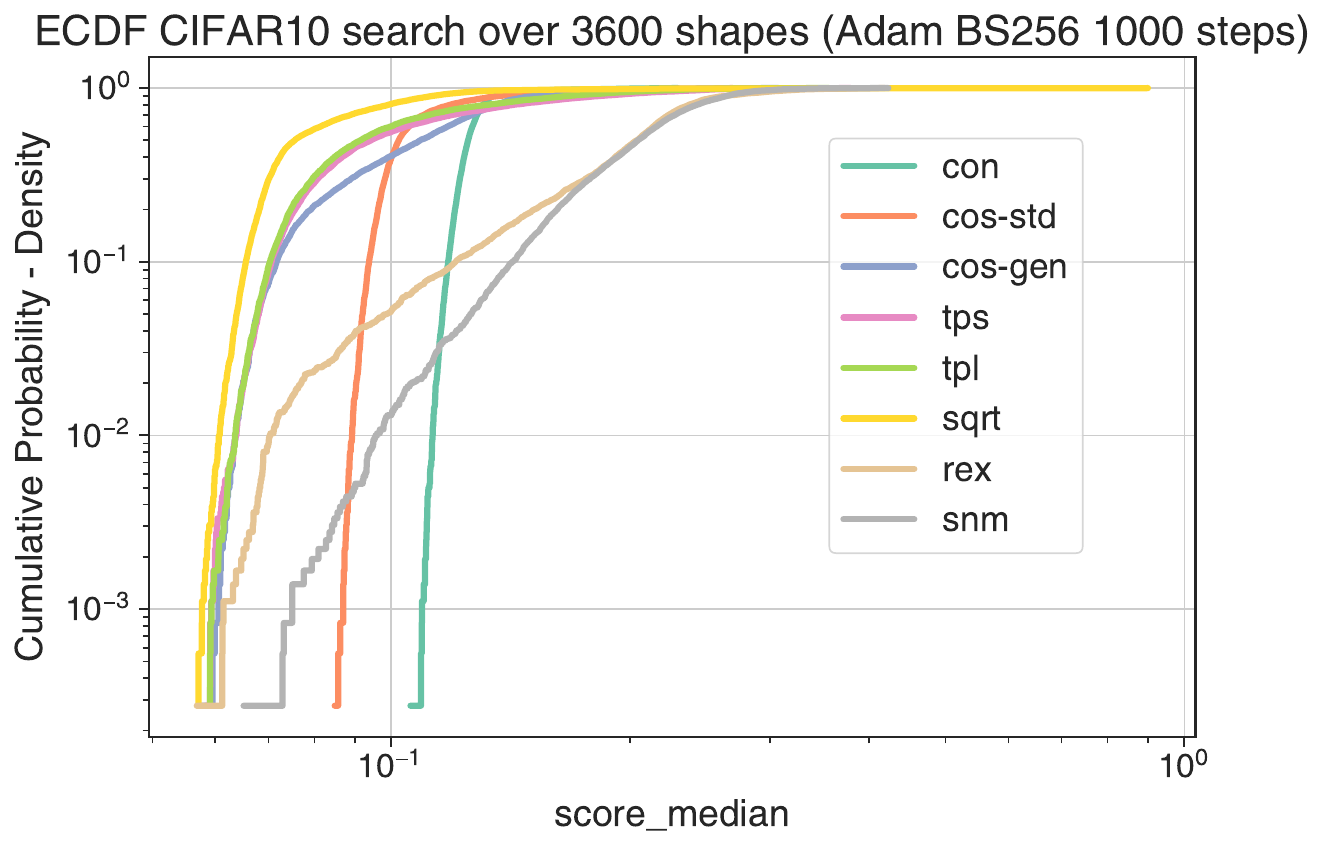}
    \end{subfigure}
    \begin{subfigure}[b]{0.49\linewidth}
    \centering
      \includegraphics[height=0.5\linewidth]{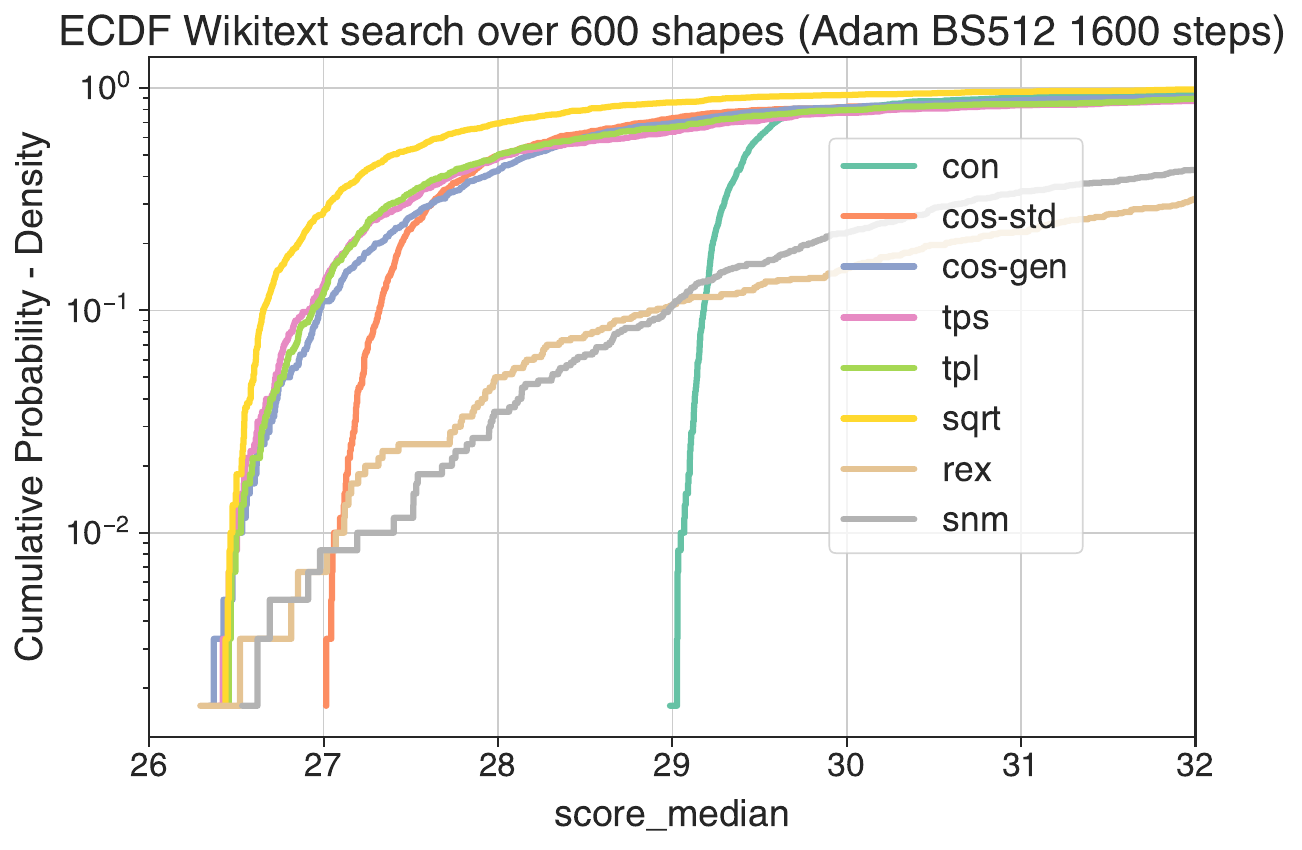}
    \end{subfigure}
    \caption{
     ECDF (Empirical cumulative distribution functions) of the score reached by random search within each learning rate schedule family. The horizontal axis reports the median training error for \cifar or the median training perplexity for \wikitext, where lower values mark better learning. The vertical axis shows the chance of reaching or beating each score. On \cifar, the constant schedule and the cosine schedule with exponent 1.0 reduce error with high probability yet stall near 0.05. \sqrtdecay schedules move the curve left most sharply. \tps, \tpl and \cosgen trace a similar path. \rex and \snm attain lower scores at lower CDF probability (alternatively, after more samples). \snm in particular lags because its large parameter space slows the search.
    }
    \label{fig:ecdf}
\end{figure}

As we discussed in Section \ref{sec:lin_reg_results} in the context of the linear regression workload, the \snm family is not due to some fundamental limitation of its
representation, but by the difficulty of optimizing a higher dimensional family using random search alone.
We can see this by plotting the empirical cumulative distribution function (ECDF) of the training metrics (Figure \ref{fig:ecdf}). The ECDF plots the selection score on the $x$-axis and the normalized rank on the $y$-axis;
in other words, it is the inverse of the function $s(r)$ going from normalized rank to the score at that rank in a dataset.

For random search, given a true CDF function $C(s)$ the expected best score of a search over $k$ random samples is given by $s^*$ such that $C(s^*) = k$.
Therefore the ECDFs give us a way to estimate the sample efficiency of random search; ECDFs for efficient methods will reach lower score values
for equivalent normalized ranks (efficient methods generate ECDFs that reach the lower left of the plots). We can see that the \con and \cosstd ECDFs
converge to a point to the right of the other families; \snm in contrast has an ECDF curve that is still traveling to the left (non-zero slope until the final few ranks), but does not reach the other schedules. This gives us additional evidence that the sample efficiency of the search is a problem.
Indeed, increasing the number of samples for \snm specifically by a factor of
$10$ still does not close the gap (Appendix \ref{app:snm_ecdf_extra}).

The difficulty of optimizing \snm schedules with random search is due to the fact that warmup from (near) zero and decay to (near) zero learning rate is a key feature of the best schedules. With our
random search procedure, the prior corresponding to these features is small for the \snm schedule. A better search procedure (either random search with
a better prior, or some adaptive search method) would likely greatly improve our ability to find near-optimal \snm schedules.

Overall our results suggest that all our schedules save the \snm can be reasonably considered to be ``near-optimal''. This suggests that for small
workloads, our random search procedure can be used to reasonably optimize schedule families which naturally incorporate warmup and decay.

\subsection{Workload variations}

\label{sec:workload_variations}

In our workload definitions, we focused on the case where all optimizer hyperparameters are fixed except for the learning rate at each step.
In practice, these hyperparameters may also need to be tuned, raising the question of how---and to what extent---the optimal schedule shape depends on the exact values of these other training hyperparameters?

To answer this question, we conducted a series of experiments where we independently varied the \adamw hyperparameters $\beta_{1}$, $\beta_{2}$,
and $\lambda_{WD}$ (the decoupled weight decay parameter). We focused on the \tps family, which gave strong results in our previous experiments due to
its balance of flexibility and searchability. For each hyperparameter, we selected a grid of values, which defined a set of experimental conditions. For each condition, we performed our search and evaluation protocols to select the best schedule for that particular hyperparameter value.
In general, our search procedure will almost always return a ``best'' shape for a given setting of (say) $\beta_{1}$ that looks at least slightly different from the shape returned for another value, even if the best shape has no meaningful dependence on $\beta_{1}$. 
However, if $\beta_{1}$ has little effect on which shapes perform well, then a shape selected with $\beta_{1}=0.8$ should perform just as well when re-evaluated at $\beta_{1}=0.9$ as the shape originally found using $\beta_{1}=0.9$. To rule out the case where superficially different optimal shapes do not lead to detectably different training outcomes, we re-evaluated the optimal shape found in each condition under every other condition.

All experiments were done on both the \cifar and the \wikitext workloads. We describe the results in the remainder of this section.

\begin{figure}[tb]
    \centering
    \begin{subfigure}[b]{0.49\linewidth}
      \includegraphics[width=\linewidth]{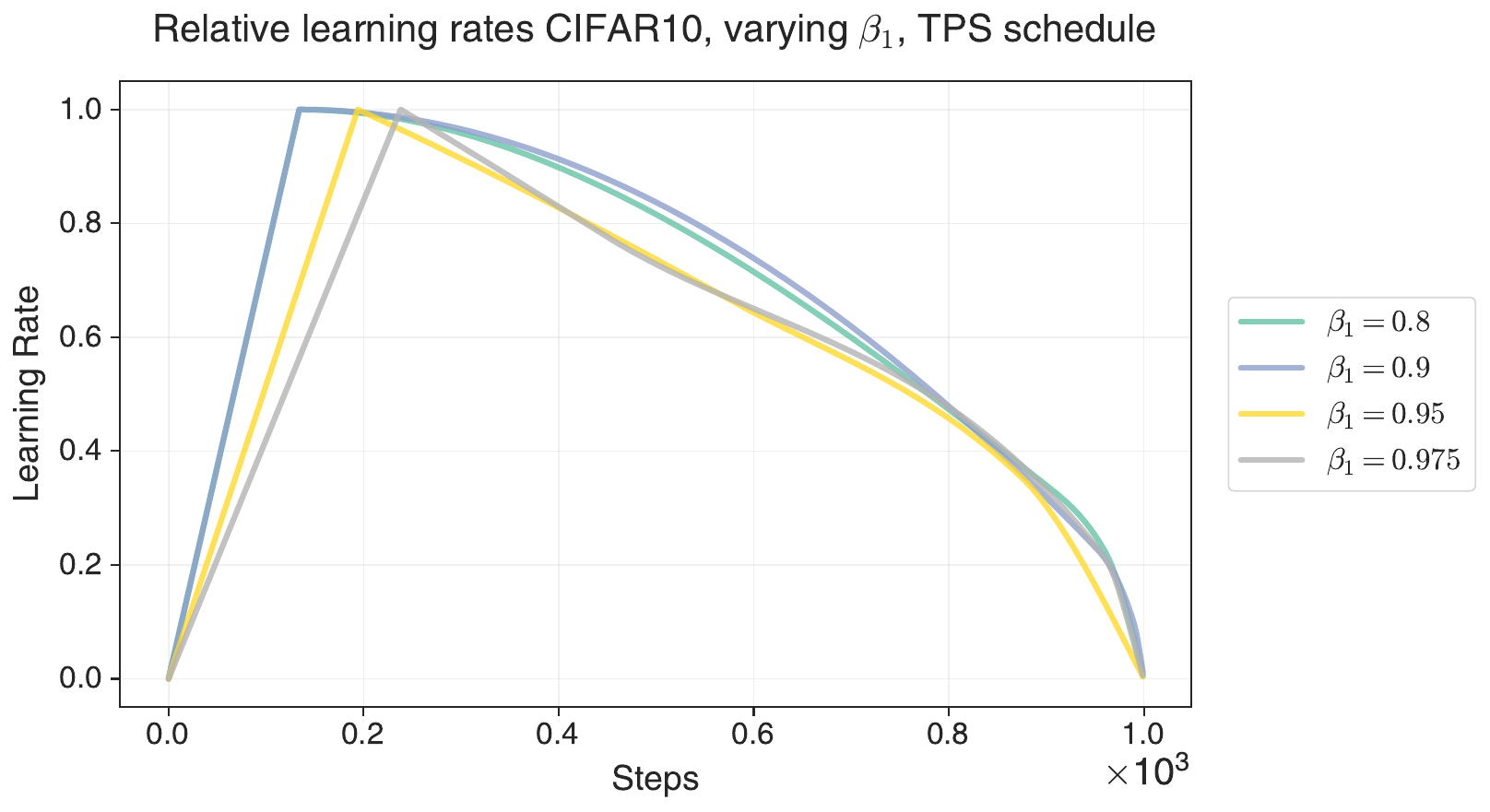}
    \end{subfigure}
    \begin{subfigure}[b]{0.49\linewidth}
      \includegraphics[width=\linewidth]{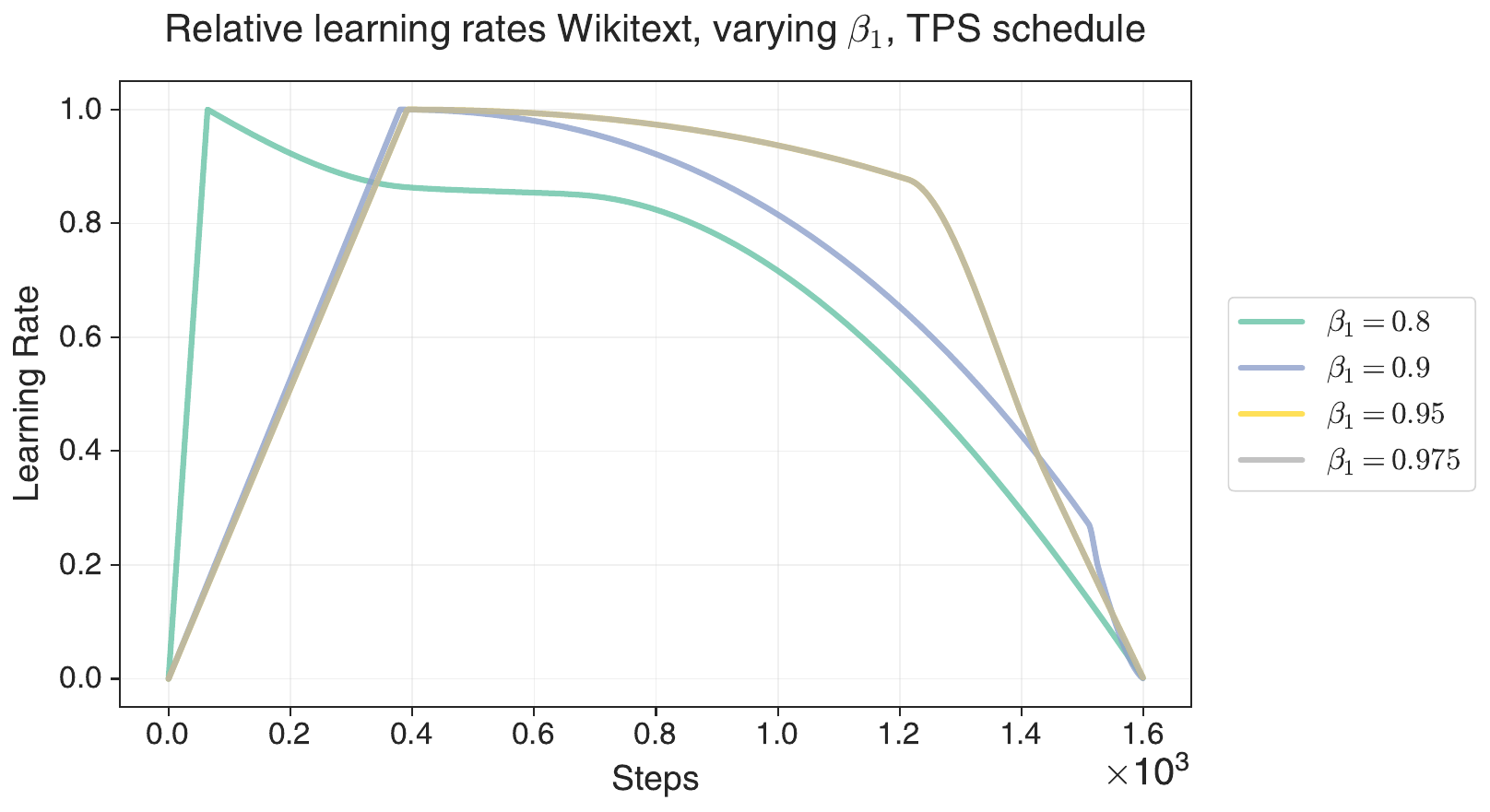}
    \end{subfigure}
    \begin{subfigure}[b]{0.49\linewidth}
      \includegraphics[width=\linewidth]{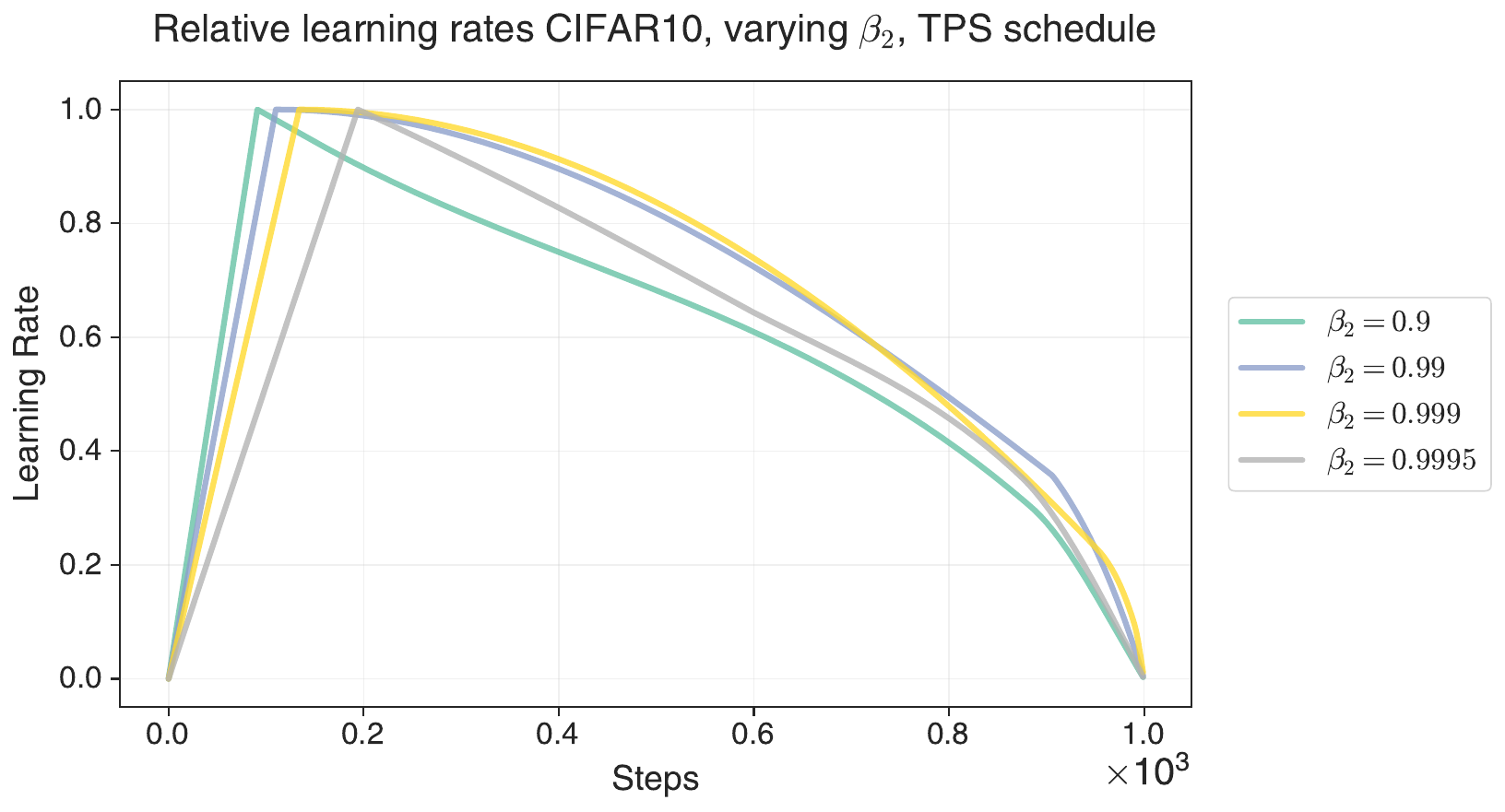}
    \end{subfigure}
    \begin{subfigure}[b]{0.49\linewidth}
      \includegraphics[width=\linewidth]{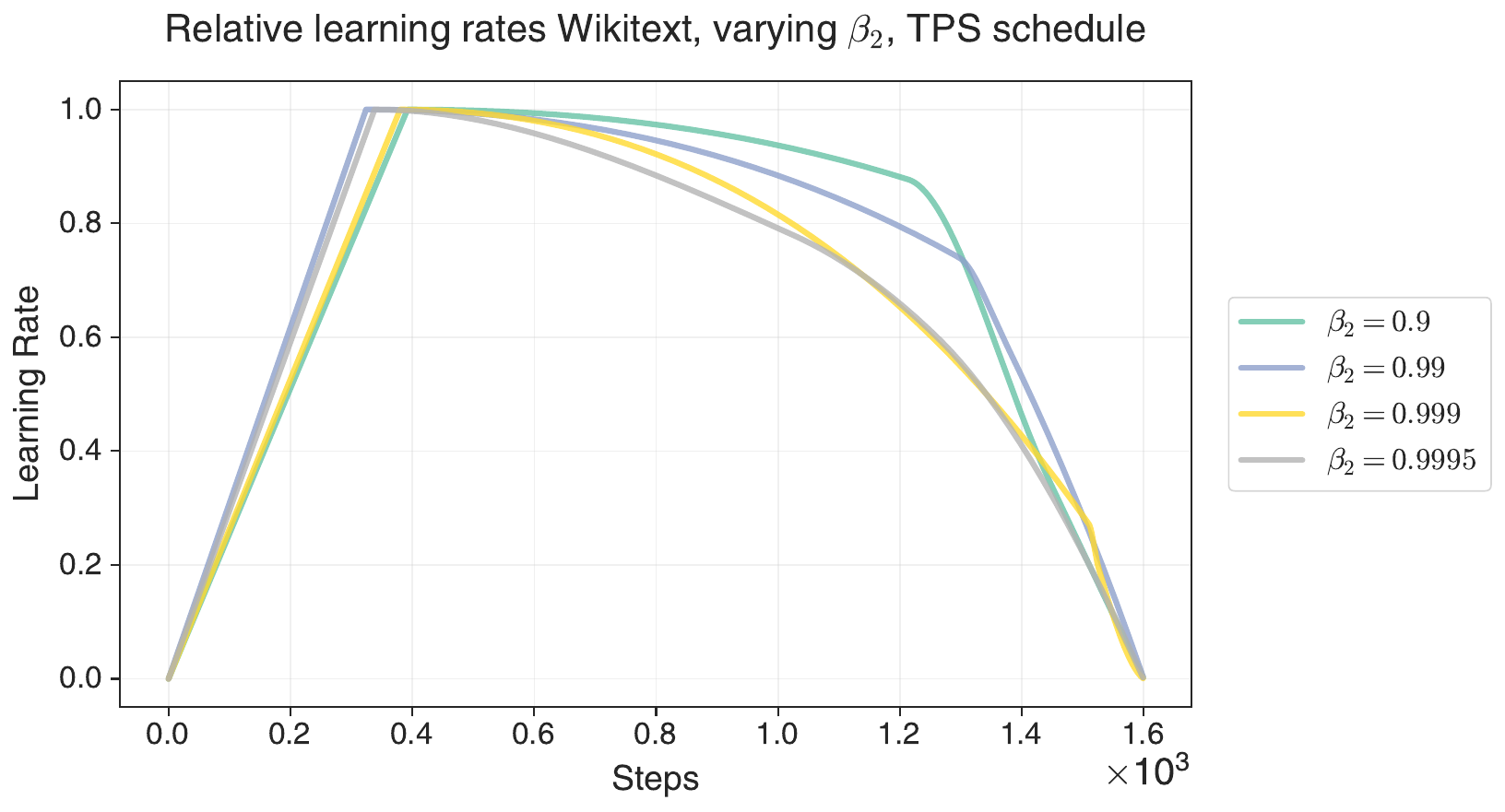}
    \end{subfigure}
    \caption{
    Optimal Shape varying $\beta_1$ and $\beta_2$ | Relative learning-rate schedules that minimise loss after tuning $\beta_1$ and $\beta_2$. The top row varies $\beta_1$ and the bottom row varies $\beta_2$. Plots on the left use \cifar, plots on the right use \wikitext. Low $\beta_1$ values give a short warmup and a brief high-rate plateau, while high $\beta_1$ values extend both. On \cifar, larger $\beta_2$ values also extend the warmup, whereas $\beta_2$ shows no clear link to schedule shape on \wikitext. 
    }
    \label{fig:opt_shapes_varying_beta}
\end{figure}

\begin{table}[h]
\centering
\begin{tabular}{@{}ccccc@{}}
\toprule
{$\beta_1$, Selection condition} & \multicolumn{4}{c}{$\beta_1$, Evaluation condition} \\
\cmidrule(l){2-5}
& 0.8 & 0.9 & 0.95 & 0.975 \\
\midrule
0.8 & $0.07 \pm 0.002$ & $0.072 \pm 0.002$ & $0.089 \pm 0.001$ & $0.123 \pm 0.002$ \\
0.9 & $0.068 \pm 0.002$ & $0.067 \pm 0.001$ & $0.083 \pm 0.001$ & $0.116 \pm 0.002$ \\
0.95 & $0.066 \pm 0.001$ & $0.066 \pm 0.001$ & $0.082 \pm 0.001$ & $0.114 \pm 0.001$ \\
0.975 & $0.068 \pm 0.004$ & $0.065 \pm 0.001$ & $0.081 \pm 0.001$ & $0.11 \pm 0.001$ \\
\bottomrule
\end{tabular}
\caption{Median \cifar training error with one standard deviation for every pair of $\beta_1$ values used at schedule selection (rows) and evaluation (columns). A larger $\beta_1$ in the search phase lowers the error. The lowest error appears when the schedule found at $\beta_1$ = 0.95 runs with $\beta_1$ = 0.8. In most rows a lower $\beta_1$ at evaluation further reduces the error, which suggests that a drop in momentum after a high-momentum search can improve performance.}
\label{tab:beta1_cifar}
\end{table}

\begin{table}[h]
\centering
\begin{tabular}{@{}ccccc@{}}
\toprule
{$\beta_1$, Selection condition} & \multicolumn{4}{c}{$\beta_1$, Evaluation condition} \\
\cmidrule(l){2-5}
& 0.8 & 0.9 & 0.95 & 0.975 \\
\midrule
0.8 & $28.8 \pm 0.96$ & $28.2 \pm 0.09$ & $33.04 \pm 0.06$ & $46.67 \pm 0.59$ \\
0.9 & $37.58 \pm 1.91$ & $27.01 \pm 0.92$ & $27.11 \pm 0.02$ & $30.97 \pm 0.06$ \\
0.95 & $39.27 \pm 1.85$ & $26.99 \pm 1.02$ & $26.91 \pm 0.04$ & $30.51 \pm 0.04$ \\
0.975 & $40.61 \pm 4.09$ & $26.92 \pm 0.92$ & $26.91 \pm 0.08$ & $30.47 \pm 0.07$ \\
\bottomrule
\end{tabular}
\caption{Median \wikitext training perplexity and one standard deviation for every pair of $\beta_1$ values used in schedule selection (rows) and evaluation (columns). Rows list the $\beta_1$ that guides the search; columns give the $\beta_1$ used to run the schedule. A higher $\beta_1$ in the search stage lowers perplexity, except for selection at $\beta_1$ = 0.975, which yields larger values. The best result appears when schedules found at $\beta_1$ = 0.95 or 0.975 run with $\beta_1$ = 0.95. Evaluation at $\beta_1$ = 0.8 shows the widest spread, which points to high uncertainty for low-momentum runs.}
\label{tab:beta1_wikitext}
\end{table}

\paragraph{Varying adamw $\beta_{1}$ and $\beta_{2}$.} We found only minor optimal shape variations when varying $\beta_{1}$ and $\beta_{2}$ in our
\cifar experiments (Figure \ref{fig:opt_shapes_varying_beta}, left column). In \wikitext, there appear to be some small variations; larger $\beta_{1}$
seemed to favor later decay, and larger $\beta_{2}$ seemed to favor earlier decay (Figure \ref{fig:opt_shapes_varying_beta}, right column).
From the re-evaluation experiments for $\beta_{1}$, we see that only $\beta_{1} = 0.8$ in \wikitext gives convincing evidence of a relationship with the learning rate schedule (Tables
\ref{tab:beta1_cifar} and \ref{tab:beta1_wikitext}). In both cases, smaller $\beta_{1}$ reached a lower training loss. In contrast, for $\beta_{2}$ our experiments do not provide any convincing evidence for the value of $\beta_{2}$ having much of an effect on what schedule shapes perform well in any setting.

\begin{table}[h]
\centering
\begin{tabular}{@{}ccccc@{}}
\toprule
{$\beta_2$, Selection condition} & \multicolumn{4}{c}{$\beta_2$, Evaluation condition} \\
\cmidrule(l){2-5}
& 0.9 & 0.99 & 0.999 & 0.9995 \\
\midrule
0.9 & $0.061 \pm 0.001$ & $0.062 \pm 0.001$ & $0.07 \pm 0.001$ & $0.072 \pm 0.001$ \\
0.99 & $0.062 \pm 0.001$ & $0.062 \pm 0.001$ & $0.066 \pm 0.001$ & $0.068 \pm 0.001$ \\
0.999 & $0.064 \pm 0.002$ & $0.064 \pm 0.001$ & $0.072 \pm 0.002$ & $0.07 \pm 0.002$ \\
0.9995 & $0.058 \pm 0.001$ & $0.059 \pm 0.001$ & $0.066 \pm 0.001$ & $0.065 \pm 0.001$ \\
\bottomrule
\end{tabular}
\caption{Median \cifar training error $\pm$ SD for every pair of $\beta_2$ values at schedule selection (rows) and evaluation (columns). Rows list the $\beta_2$ used to pick the schedule; columns give the value used to run it. Higher $\beta_2$ at selection tends to cut error, yet the lowest error comes from schedules tuned at $\beta_2$ = 0.9995 and run at $\beta_2$ = 0.9. 
}
\label{tab:beta2_cifar}
\end{table}

\begin{table}[h]
\centering
\begin{tabular}{@{}ccccc@{}}
\toprule
{$\beta_2$, Selection condition} & \multicolumn{4}{c}{$\beta_2$, Evaluation condition} \\
\cmidrule(l){2-5}
& 0.9 & 0.99 & 0.999 & 0.9995 \\
\midrule
0.9 & $26.23 \pm 0.02$ & $26.4 \pm 0.29$ & $26.87 \pm 6.03$ & $26.77 \pm 11.37$ \\
0.99 & $26.32 \pm 0.01$ & $26.29 \pm 0.35$ & $26.59 \pm 0.85$ & $26.71 \pm 2.26$ \\
0.999 & $26.32 \pm 0.01$ & $26.56 \pm 0.38$ & $26.74 \pm 1.42$ & $26.76 \pm 1.21$ \\
0.9995 & $26.36 \pm 0.01$ & $26.42 \pm 0.27$ & $26.75 \pm 2.11$ & $26.74 \pm 2.77$ \\
\bottomrule
\end{tabular}
\caption{Median \wikitext perplexity $\pm$ SD for every pair of $\beta_2$ values used at schedule selection (rows) and evaluation (columns). A higher $\beta_2$ at selection lowers the median, yet the best result appears when a schedule tuned at $\beta_2$ = 0.9995 runs with $\beta_2$ = 0.9. Selection at $\beta_2$ = 0.9 shows the widest spread, confirming that low $\beta_2$ during tuning adds noise. 
}
\label{tab:beta2_wikitext}
\end{table}

\begin{figure}[tb]
    \centering
    \begin{subfigure}[b]{0.49\linewidth}
      \includegraphics[width=\linewidth]{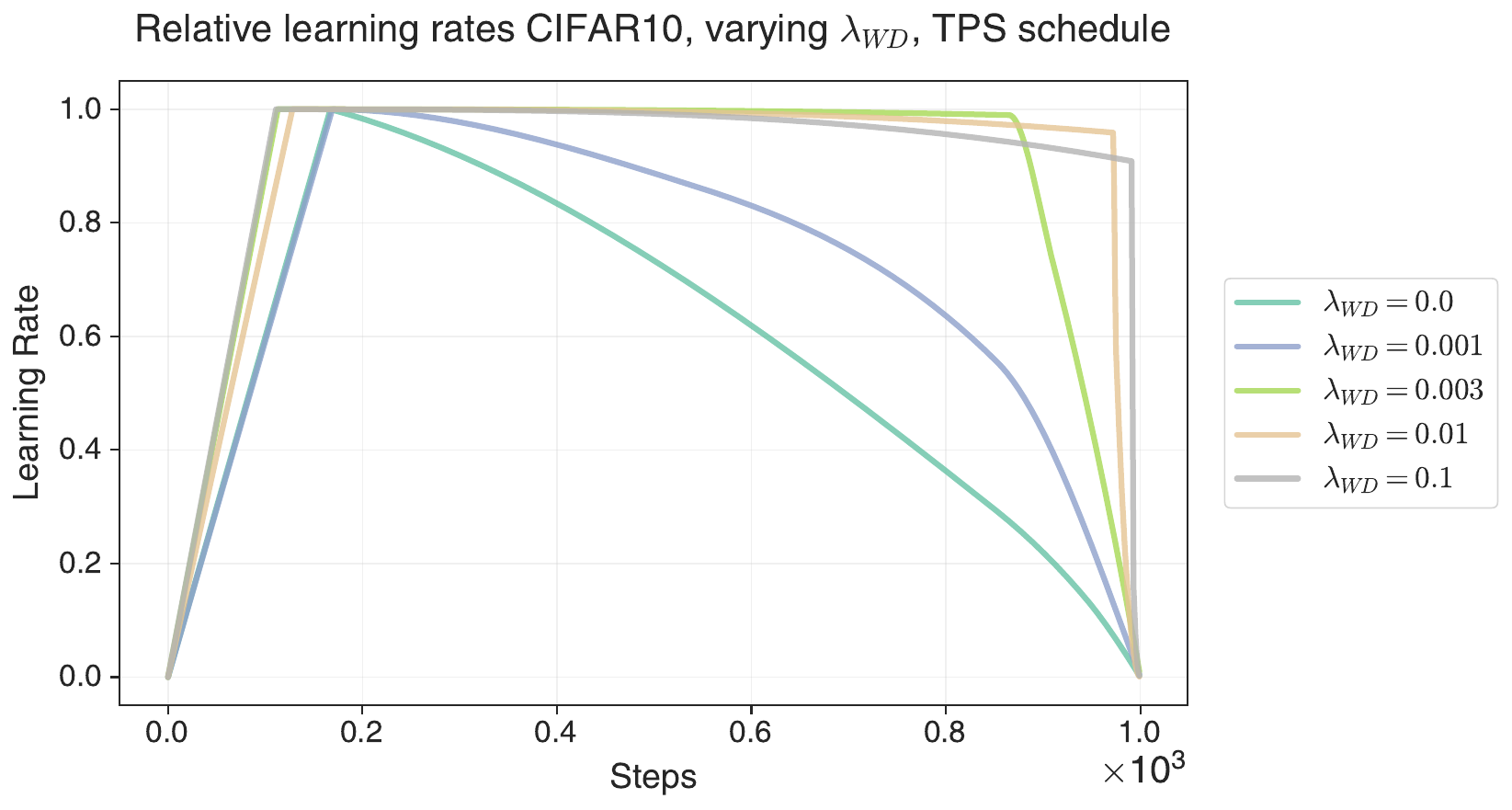}
    \end{subfigure}
    \begin{subfigure}[b]{0.49\linewidth}
      \includegraphics[width=\linewidth]{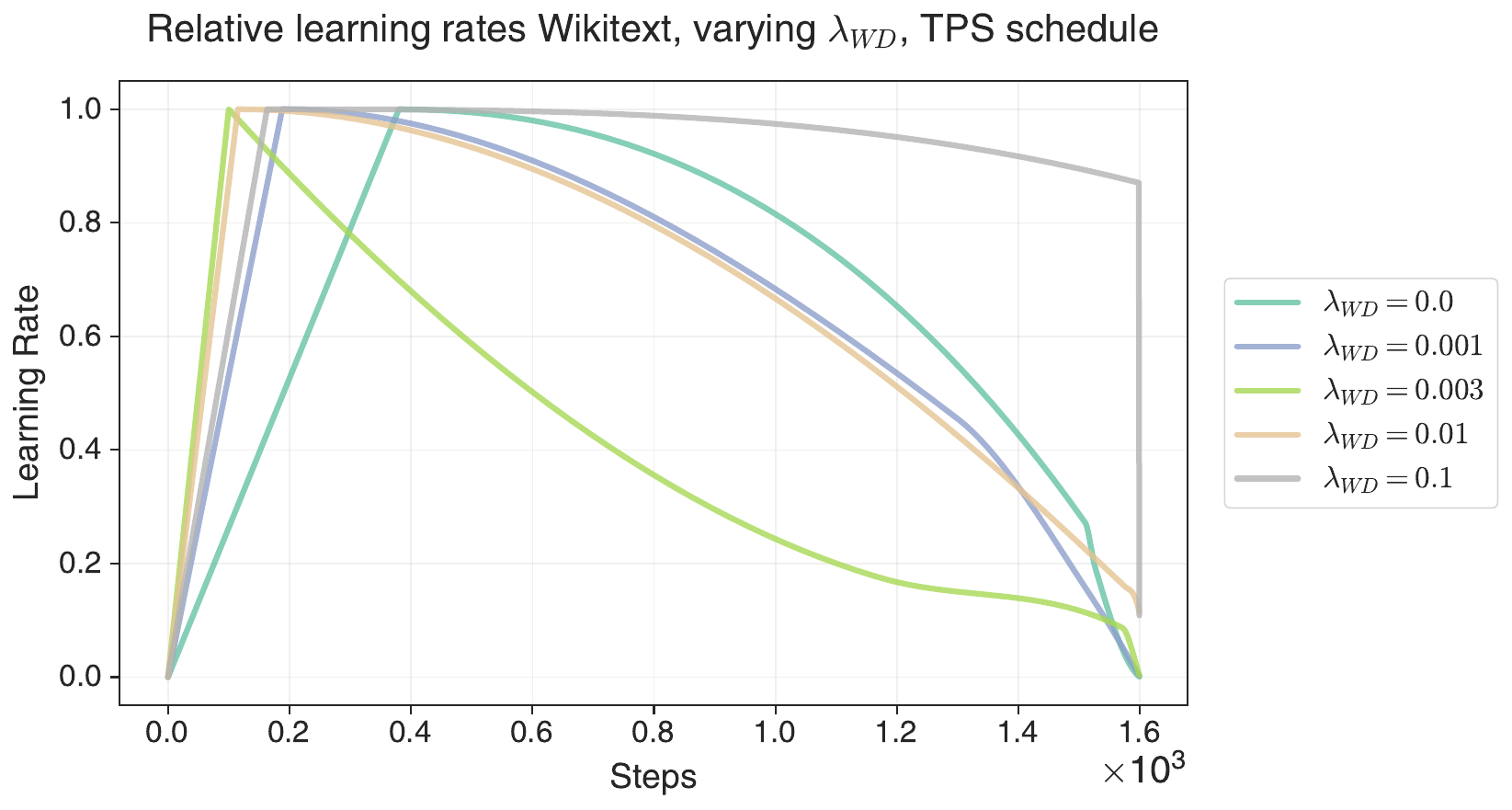}
    \end{subfigure}
    \caption{
    Optimal Shape varying Weight Decay | Relative learning rate schedules that yield the lowest loss after tuning weight decay. Left panel shows \cifar; right panel shows \wikitext. Small decay values lengthen the warmup, while large decay values keep the rate high until the final training steps. Both tasks trace the same pattern. 
    }
    \label{fig:opt_shapes_varying_weight_decay}
\end{figure}

\paragraph{Varying $\lambda_{WD}$.} Interestingly, varying the (decoupled) weight decay strength revealed a strong relationship between $\lambda_{WD}$ and
the learning rate schedule. In both workloads, increasing the weight decay favored schedules that decayed later (Figure \ref{fig:opt_shapes_varying_weight_decay}). 
Our re-evaluation experiments generally showed that the schedules selected
for a particular value of $\lambda_{WD}$ tend to be optimal for that specific
value (Tables \ref{tab:wd_cifar} and \ref{tab:wd_wikitext}), as we would expect if changing the weight decay strength truly changed what schedule shapes performed well.
Note that of the conditions we searched, $\lambda_{WD} = 0$ gives the best overall
performance (as we would expect since we are comparing \emph{training} metrics rather than validation or test metrics).

\begin{figure}[h]
   \centering
    \begin{subfigure}[b]{0.49\linewidth}
      \includegraphics[width=\linewidth]{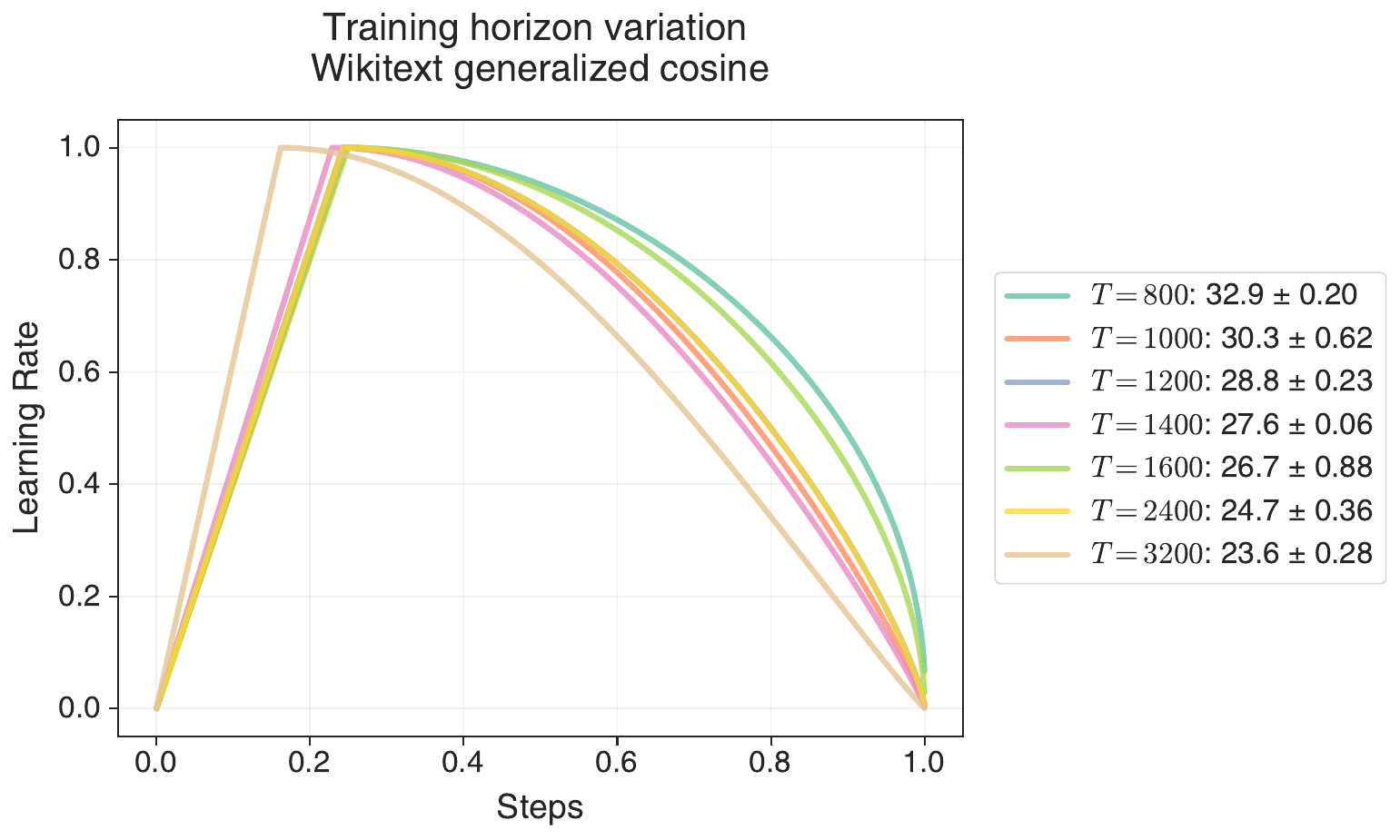}
    \end{subfigure}
    \begin{subfigure}[b]{0.49\linewidth}
      \includegraphics[width=\linewidth]{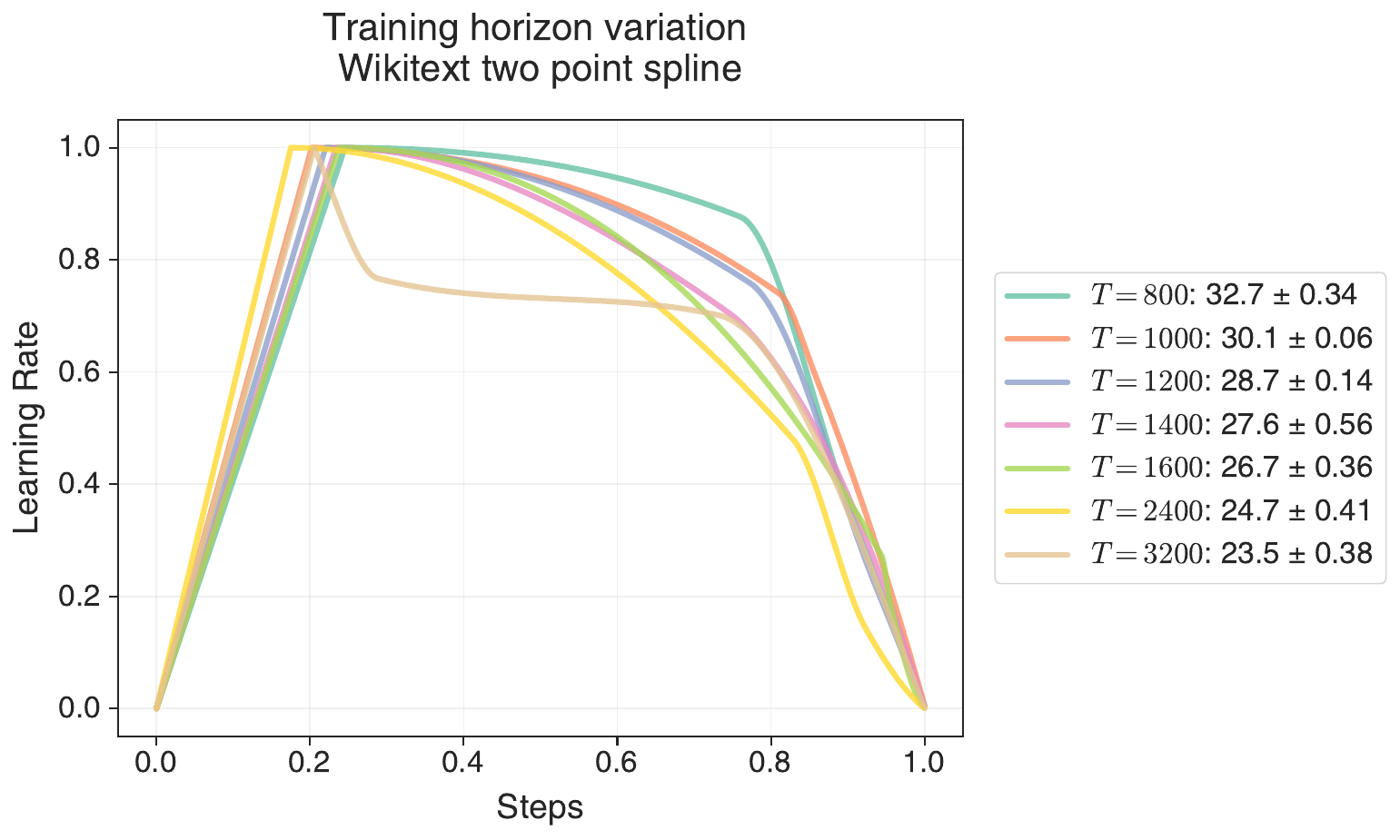}
    \end{subfigure}
    \caption{Optimal learning rate shapes for \cosgen (left) and \tps (right) for wikitext workload trained with varying training horizon. Performance improves with more steps, which favors a more steady decay. Warmup fraction remains consistent at this scale.
        }
    \label{fig:opt_shapes_varying_horizon}
\end{figure}

\paragraph{Varying the training horizon.} For the \wikitext workload, we varied the training horizon---the total number of training steps $T$---for the \cosgen and \tps schedules.
The base learning rates were similar in all settings, corresponding to two adjacent values of our search grid with no clear pattern.
Plotting the
schedule shapes against the fraction of training steps reveals a trend towards gentler decay for larger number of training steps (Figure \ref{fig:opt_shapes_varying_horizon}).
We also see that warmup fraction seems stable across horizons, which suggests that setting warmup fraction may be a more robust strategy than
searching over/fixing the number of warmup steps. Training horizon experiments are much less interesting on \cifar because training for even a bit longer than we did in our experiments quickly saturates the training error at zero errors, making it hard to stay in the optimization-limited regime or learn very much.

\begin{table}[h]
\centering
\begin{tabular}{@{}ccccc@{}}
\toprule
{$\lambda_{WD}$, Selection condition} & \multicolumn{4}{c}{$\lambda_{WD}$, Evaluation condition} \\
\cmidrule(l){2-5}
& 0.0 & 0.001 & 0.01 & 0.1 \\
\midrule
0.0 & $0.071 \pm 0.001$ & $0.127 \pm 0.003$ & $0.532 \pm 0.004$ & $0.9 \pm 0.0$ \\
0.001 & $0.067 \pm 0.001$ & $0.123 \pm 0.002$ & $0.477 \pm 0.003$ & $0.9 \pm 0.0$ \\
0.01 & $0.101 \pm 0.001$ & $0.149 \pm 0.001$ & $0.406 \pm 0.004$ & $0.9 \pm 0.002$ \\
0.1 & $0.109 \pm 0.001$ & $0.164 \pm 0.001$ & $0.415 \pm 0.002$ & $0.839 \pm 0.004$ \\
\bottomrule
\end{tabular}
\caption{Median \cifar training error $\pm$ SD for every pair of weight decay values used in schedule selection (rows) and later evaluation (columns). An evaluation decay of 0.01 or 0.1 lifts the error across all rows, while decay 0 or 0.001 yields the lowest values.
}
\label{tab:wd_cifar}
\end{table}

\begin{table}[h]
\centering
\begin{tabular}{@{}ccccc@{}}
\toprule
{$\lambda_{WD}$, Selection condition} & \multicolumn{4}{c}{$\lambda_{WD}$, Evaluation condition} \\
\cmidrule(l){2-5}
& 0.0 & 0.001 & 0.01 & 0.1 \\
\midrule
0.0 & $26.75 \pm 1.14$ & $29.2 \pm 0.78$ & $122.39 \pm 15.95$ & $5336.9 \pm 21.36$ \\
0.001 & $27.08 \pm 0.72$ & $27.05 \pm 0.35$ & $101.69 \pm 12.31$ & $4169.88 \pm 43.03$ \\
0.01 & $27.8 \pm 0.08$ & $27.61 \pm 0.06$ & $69.48 \pm 4.34$ & $443.55 \pm 21.18$ \\
0.1 & $29.16 \pm 0.79$ & $30.93 \pm 3.52$ & $165.99 \pm 33.78$ & $392.74 \pm 10.19$ \\
\bottomrule
\end{tabular}
\caption{Median \wikitext training perplexity $\pm$ SD for every pair of weight-decay values used in schedule selection (rows) and later evaluation (columns). Lower decay during the search phase gives smaller medians. At evaluation, decay 0 or 0.001 holds perplexity down, whereas 0.01 or 0.1 drives it up. 
}
\label{tab:wd_wikitext}
\end{table}

\clearpage
\newpage
\section{Discussion}

\subsection{Lessons about search methodology}

Our search methodology was designed to be simple, but it is not particularly efficient. Crucially, the search always works better with more seeds
and more search points. The choices we made in our experiments generally seemed to yield plausibly near-optimal schedules. In the case of \cifar, our random search procedure seems sufficient at the small scale of
our experiments; on \wikitext we were also generally able to draw conclusions, but there was more variability---in part due to the reduced number of seeds
and search points, and in part due to the increased variance in the stability of training runs with good median final train perplexities.

Our results (such as those presented in Figure \ref{fig:base_lr_heatmap}) validated our choice to separately optimize the base learning rate for each schedule shape.
By tuning the base learning rate for each shape, we isolated the effect of the shape from the strong base learning rate
effects (that could appear implicitly through shapes that stay close to their peak longer).
The efficiency of the random search can obviously be improved in various ways, but we have not even really optimized the allocation of computational resources across search points, seeds, and evaluation, let alone explored more sophisticated search strategies.

Although generally sufficient, for the \snm family (which was maximally flexible but correspondingly hard to search) our search procedure wasn't quite as effective as it needed to be. This was in part because
warmup and decay seem to be integral parts of any successful learning rate schedule; schedules with these properties compose a small fraction of the configuration space of the \snm family.

Searching more flexible families (perhaps ones even more flexible than the \snm family) could benefit from improved techniques. Random search could be augmented with better priors, or
multi-round searches that incorporate feedback could be used, such as Bayesian optimization or evolutionary algorithms. These methods
could also improve search efficiency in the more easily searchable families,
which could be of practical interest for more expensive workloads. However, although more efficient, these kinds of techniques can make the results harder to interpret since they could potentially introduce a misleading bias towards different parts of the search space, depending on the exact details.

Though we applied our search methodology to learning rate schedules, most optimizers contain other scalar hyperparameters that could benefit from scheduling. The most obvious
is momentum \citep{lee2022trajectory, ferbach2025dimension}, but Adam $\beta_{2}$ and $\lambda_{WD}$ may benefit from scheduling as well.
A natural extension of our work is to search flexible schedule families on these other parameters as well.

\subsection{Practical takeaways}

The most important, if somewhat obvious, takeaway from our work is that without optimizing the base learning rate, trying to optimize the schedule shape isn't very meaningful. Anyone attempting to search over or tune schedule shapes should make sure to devote sufficient resources to tuning the base learning rate. Also, when trying some new learning rate schedule family, the base learning rate must be tuned anew in order to minimize false negative results.

Our results confirm that, at a high level, the common practice of a warmup period followed by monotonic decay is indeed a consistently effective strategy. The best members of the
\snm family, which does not guarantee either phase, showed these features. This result suggests that near-optimal learning rate schedule searches
could have revealed these rules of thumb earlier on in the history of deep learning.

Our analysis of the linear regression workload represents, to the best of our knowledge, the first optimal schedule for a non-trivial, high-dimensional
problem in that class.
It is interesting to note that the optimal
schedule for linear regression is relatively ``smooth'', despite no such constraints induced by the optimization procedure; each of the $1000$
learning rates at each step was individually optimized.
The schedule favors no warmup as well as a relatively flat schedule with a sharp decay at the end of training.
The shape of the optimal
schedule in linear regression depends on the training horizon, as well as the spectrum of the data covariance. We hypothesize that changing these
choices might change the decay profile, but shouldn't make warmup suddenly become useful. 

Interestingly, the optimal schedules for \cifar and \wikitext are quite different from the optimal schedule for the linear regression workload,
namely in that warmup is useful across families, workloads, and workload variations. Comparing our results on linear regression and neural net training indicate extreme caution is necessary when applying principles derived from studies of convex optimization to the non-convex, non-linear setting of optimization in deep learning, specifically when trying to
understand useful principles for early and intermediate time training.

Our results on \cifar and \wikitext suggest that, if tuning resources are available, it is indeed worth considering schedules beyond popular cosine
decay schedule. The ability to change the shape of the decay gave benefits in both train and test set metrics.
In our workloads, these gains tended to be small but significant.

The differences in final training metrics between the various more flexible families, and between top members of each family, seem to be small. This
result suggests that it might not be worth searching \emph{even more flexible} schedule families. Even the \cosgen family, with one more
parameter than \cosstd, captured significant gains in the \cifar case. Another popular and slightly slightly more general cosine family than \cosstd uses a non-zero final learning rate (\shortcosstd). When this final learning rate is tuned well enough, we saw some gains over \shortcosstd, but still didn't get as good results as \shortcosgen (Appendix \ref{app:non_zero_decay}).
If computational resources are more abundant, then the \tpl or \tps families
likely have more than enough flexibility to capture schedules very close to optimal.

Our ability to find near-optimal schedules may be useful for research into automatic learning rate selectors. By measuring different quantities during training, we could hope to predict the (known) optimal learning rate
schedule as a function of these quantities. This might relate to simple measurements like the loss trajectory, gradient norms, or simple Hessian
statistics that can be measured cheaply during training. We can do even better in the linear regression setting where we can solve for the optimal schedule, and compute many of these quantities analytically.

The generality of our conclusions is limited primarily by our small number of workloads. Applying our methodology to a larger variety of workloads
will give more evidence of which features of optimal schedule shapes are general versus specific. Nonetheless, even this initial study has clarified
some mysteries and gives a path to better understanding of learning rate schedules.

\section*{Acknowledgements} 
We thank Shankar Krishnan, Sourabh Medapati, Dougal Maclaurin and Vincent Roulet for their discussions on the methods, support in experimentation, and detailed feedback on the manuscript.

\bibliography{main}

@article{loshchilov2016sgdr,
  title={{SGDR}: Stochastic gradient descent with warm restarts},
  author={Loshchilov, Ilya and Hutter, Frank},
  journal={arXiv preprint arXiv:1608.03983},
  year={2016}
}

@inproceedings{smith2017cyclical,
  author={Smith, Leslie N.},
  booktitle={2017 IEEE Winter Conference on Applications of Computer Vision (WACV)}, 
  title={Cyclical Learning Rates for Training Neural Networks}, 
  year={2017},
  volume={},
  number={},
  pages={464-472},
  doi={10.1109/WACV.2017.58}}

@article{kingma2014adam,
  title={Adam: A method for stochastic optimization},
  author={Kingma, Diederik P and Ba, Jimmy},
  journal={arXiv preprint arXiv:1412.6980},
  year={2014}
}

@incollection{bengio2012practical,
  title={Practical recommendations for gradient-based training of deep architectures},
  author={Bengio, Yoshua},
  booktitle={Neural Networks: Tricks of the Trade: Second Edition},
  pages={437--478},
  year={2012},
  publisher={Springer}
}

@article{bergstra2012random,
  title={Random search for hyper-parameter optimization},
  author={Bergstra, James and Bengio, Yoshua},
  journal={The journal of machine learning research},
  volume={13},
  number={1},
  pages={281--305},
  year={2012},
  publisher={JMLR. org}
}

@misc{bousquet2017critical,
      title={Critical Hyper-Parameters: No Random, No Cry}, 
      author={Olivier Bousquet and Sylvain Gelly and Karol Kurach and Olivier Teytaud and Damien Vincent},
      year={2017},
      eprint={1706.03200},
      archivePrefix={arXiv},
      primaryClass={cs.LG},
      url={https://arxiv.org/abs/1706.03200}, 
}

@Article{mockus1978bayesopt,
  author       = {Mockus, Jonas and Tie\v{s}is, Vytautas and \v{Z}ilinskas, Antanas},
  title        = {The Application of {B}ayesian Methods for Seeking the Extremum},
  volume       = {2},
  number       = {117-129},
  pages        = {2},
  year         = {1978},
  journal = {Towards Global Optimization},
  publisher    = {Amsterdam: Elsevier},
}

@article{snoek2012practical,
  title={Practical bayesian optimization of machine learning algorithms},
  author={Snoek, Jasper and Larochelle, Hugo and Adams, Ryan P},
  journal={Advances in neural information processing systems},
  volume={25},
  year={2012}
}

@article{goyal2017accurate,
  title={Accurate, large minibatch sgd: Training imagenet in 1 hour},
  author={Goyal, Priya and Doll{\'a}r, Piotr and Girshick, Ross and Noordhuis, Pieter and Wesolowski, Lukasz and Kyrola, Aapo and Tulloch, Andrew and Jia, Yangqing and He, Kaiming},
  journal={arXiv preprint arXiv:1706.02677},
  year={2017}
}

@article{jin2021autolrs,
  title={Autolrs: Automatic learning-rate schedule by bayesian optimization on the fly},
  author={Jin, Yuchen and Zhou, Tianyi and Zhao, Liangyu and Zhu, Yibo and Guo, Chuanxiong and Canini, Marco and Krishnamurthy, Arvind},
  journal={arXiv preprint arXiv:2105.10762},
  year={2021}
}

@inproceedings{maclaurin2015gradient,
  title={Gradient-based hyperparameter optimization through reversible learning},
  author={Maclaurin, Dougal and Duvenaud, David and Adams, Ryan},
  booktitle={International conference on machine learning},
  pages={2113--2122},
  year={2015},
  organization={PMLR}
}

@article{baydin2017online,
  title={Online learning rate adaptation with hypergradient descent},
  author={Baydin, Atilim Gunes and Cornish, Robert and Rubio, David Martinez and Schmidt, Mark and Wood, Frank},
  journal={arXiv preprint arXiv:1703.04782},
  year={2017}
}

@article{jaderberg2017population,
  title={Population based training of neural networks},
  author={Jaderberg, Max and Dalibard, Valentin and Osindero, Simon and Czarnecki, Wojciech M and Donahue, Jeff and Razavi, Ali and Vinyals, Oriol and Green, Tim and Dunning, Iain and Simonyan, Karen and others},
  journal={arXiv preprint arXiv:1711.09846},
  year={2017}
}

@article{shallueEtAlBatchScience2019,
  author  = {Christopher J. Shallue and Jaehoon Lee and Joseph Antognini and Jascha Sohl-Dickstein and Roy Frostig and George E. Dahl},
  title   = {Measuring the Effects of Data Parallelism on Neural Network Training},
  journal = {Journal of Machine Learning Research},
  year    = {2019},
  volume  = {20},
  number  = {112},
  pages   = {1--49},
  url     = {http://jmlr.org/papers/v20/18-789.html}
}

@article{chen2022rex,
  title={Rex: Revisiting budgeted training with an improved schedule},
  author={Chen, John and Wolfe, Cameron and Kyrillidis, Tasos},
  journal={Proceedings of Machine Learning and Systems},
  volume={4},
  pages={64--76},
  year={2022}
}

@article{krizhevsky2009learning,
  title={Learning multiple layers of features from tiny images},
  author={Krizhevsky, Alex},
  year={2009},
  publisher={Toronto, ON, Canada}
}

@article{vaswani2017attention,
  title={Attention is all you need},
  author={Vaswani, Ashish and Shazeer, Noam and Parmar, Niki and Uszkoreit, Jakob and Jones, Llion and Gomez, Aidan N and Kaiser, {\L}ukasz and Polosukhin, Illia},
  journal={Advances in neural information processing systems},
  volume={30},
  year={2017}
}

@misc{merity2016pointer,
      title={Pointer Sentinel Mixture Models},
      author={Stephen Merity and Caiming Xiong and James Bradbury and Richard Socher},
      year={2016},
      eprint={1609.07843},
      archivePrefix={arXiv},
      primaryClass={cs.CL}
}

@article{gilmer2021loss,
  title={A loss curvature perspective on training instability in deep learning},
  author={Gilmer, Justin and Ghorbani, Behrooz and Garg, Ankush and Kudugunta, Sneha and Neyshabur, Behnam and Cardoze, David and Dahl, George and Nado, Zachary and Firat, Orhan},
  journal={arXiv preprint arXiv:2110.04369},
  year={2021}
}

@misc{cohen2024adaptivegradientmethodsedge,
      title={Adaptive Gradient Methods at the Edge of Stability}, 
      author={Jeremy M. Cohen and Behrooz Ghorbani and Shankar Krishnan and Naman Agarwal and Sourabh Medapati and Michal Badura and Daniel Suo and David Cardoze and Zachary Nado and George E. Dahl and Justin Gilmer},
      year={2022},
      eprint={2207.14484},
      archivePrefix={arXiv},
      primaryClass={cs.LG},
      url={https://arxiv.org/abs/2207.14484}, 
}

@article{lee2022trajectory,
  title={Trajectory of mini-batch momentum: Batch size saturation and convergence in high dimensions},
  author={Lee, Kiwon and Cheng, Andrew and Paquette, Elliot and Paquette, Courtney},
  journal={Advances in Neural Information Processing Systems},
  volume={35},
  pages={36944--36957},
  year={2022}
}

@article{agarwala2024high,
  title={High dimensional analysis reveals conservative sharpening and a stochastic edge of stability},
  author={Agarwala, Atish and Pennington, Jeffrey},
  journal={arXiv preprint arXiv:2404.19261},
  year={2024}
}

@inproceedings{devlin2019bert,
  title={Bert: Pre-training of deep bidirectional transformers for language understanding},
  author={Devlin, Jacob and Chang, Ming-Wei and Lee, Kenton and Toutanova, Kristina},
  booktitle={Proceedings of the 2019 conference of the North American chapter of the association for computational linguistics: human language technologies, volume 1 (long and short papers)},
  pages={4171--4186},
  year={2019}
}

@article{qiu2025scaling,
  title={Scaling Collapse Reveals Universal Dynamics in Compute-Optimally Trained Neural Networks},
  author={Qiu, Shikai and Xiao, Lechao and Wilson, Andrew Gordon and Pennington, Jeffrey and Agarwala, Atish},
  journal={arXiv preprint arXiv:2507.02119},
  year={2025}
}

@article{hoffmann2022training,
  title={Training compute-optimal large language models},
  author={Hoffmann, Jordan and Borgeaud, Sebastian and Mensch, Arthur and Buchatskaya, Elena and Cai, Trevor and Rutherford, Eliza and Casas, Diego de Las and Hendricks, Lisa Anne and Welbl, Johannes and Clark, Aidan and others},
  journal={arXiv preprint arXiv:2203.15556},
  year={2022}
}

@article{ferbach2025dimension,
  title={Dimension-adapted Momentum Outscales SGD},
  author={Ferbach, Damien and Everett, Katie and Gidel, Gauthier and Paquette, Elliot and Paquette, Courtney},
  journal={arXiv preprint arXiv:2505.16098},
  year={2025}
}

@article{dvoretzky1956,
    author = {A. Dvoretzky and J. Kiefer and J. Wolfowitz},
    title = {{Asymptotic Minimax Character of the Sample Distribution Function and of the Classical Multinomial Estimator}},
    volume = {27},
    journal = {The Annals of Mathematical Statistics},
    number = {3},
    publisher = {Institute of Mathematical Statistics},
    pages = {642 -- 669},
    year = {1956},
    doi = {10.1214/aoms/1177728174},
    URL = {https://doi.org/10.1214/aoms/1177728174}
}

\appendix
\section*{Appendix}

\section{Linear regression workload details}
\label{app:lin_reg_workload}

\subsection{Workload definition}

\label{app:lin_reg_workload_def}

The linear regression workload consists of a linear model on a $\P$ dimensional parameter vector $\th$, optimized on $\D$ data points
according to the following loss function:
\begin{equation}
\Lo(\th) = \frac{1}{2\D}||\z||^{2},~\z\equiv \J\th-\y_{tr}
\end{equation}
Here $\J$ is the $\D\times\P$ dimensional Jacobian, and $\y_{tr}$ are the $\D$ targets.

We train the model using SGD with learning rate schedule $\lr_{t}$, randomly sampling $\B$ out of the $\D$ datapoints independently every step. The dynamical
equations can be written as
\begin{equation}
\th_{t+1}-\th_{t} = -\frac{\lr_{t}}{\B} \J^{\tpose}\pmat_{t}\z_{t}.
\end{equation}
Here $\pmat_{t}$ is be a sequence of random, i.i.d.
diagonal matrices with exactly $\B$ random $1$s on the diagonal, and $0$s everywhere else, which represents the sampling process.

The dynamics of this model can be written in terms of the residuals $\z$ alone:
\begin{equation}
\z_{t+1}-\z_{t}  = -\frac{\lr_{t}}{\B}\ntk\pmat_{t}\z_{t}.
\label{eq:lin_reg_sgd}
\end{equation}
where $\ntk = \J\J^{\tpose}$ is the empirical neural tangent kernel.

This is the dynamics we directly simulate. We initialize $\z_{t}$ as an i.i.d.
random vector whose elements are drawn from $\mathcal{N}(0,1)$. We initialize $\ntk$ using a fixed diagonal matrix $\lmat$ and a uniformly sampled $\D\times\D$
orthogonal matrix $\U$:
\begin{equation}
\ntk = \U \lmat \U^{\tpose}
\end{equation}
We used a uniform distribution on $\lmat$, with eigenvalues given by $\lam_{k} = 2k/(\D+1)$ for $k$ from $1$ to $\D$. This ensured that the average eigenvalue of
$\ntk$ is $1$. For all our simulations, we set $\D = 500$ and trained for $1000$ steps.

\subsection{Average loss curves in the high-dimensional limit}

In the limit of large $\D$, the loss curves of the above linear regression problem concentrate around their average \citep{lee2022trajectory}. The average is
described by tracking the dynamics of the variances in each eigenmode of $\ntk$ \citep{agarwala2024high}.
We define:
\begin{equation}
\pvec_{t} = \diag(\U^{\tpose}\expect[\z_{t}\z_{t}^{\tpose}]\U)
\end{equation}
where the expectation is taken over the initialization and sampling randomness.

In the high dimensional limit, the dynamics of $\pvec$ is described by
the following $\D$ dimensional linear recurrence relationship:
\begin{equation}
\pvec_{t+1}-\pvec_{t} = \left[(\Id-\lr_{t}\lmat)^2-\Id+\frac{1}{\D}(\bfr^{-1}-1)\lr_{t}^2\lmat\m{1}\m{1}^{\tpose}\lmat\right]\pvec_{t}
\label{eq:pvec_high}
\end{equation}
where $\bfr = \B/\D$.
The average loss can be written as
\begin{equation}
\Lo(\pvec) = \frac{1}{2\D}\m{1}^{\tpose}\lmat\pvec.
\label{eq:loss_from_pvec}
\end{equation}

Combined, Equations \ref{eq:pvec_high} and \ref{eq:loss_from_pvec} imply that the loss can be written as an analytic function of the learning rate schedule.
This will allow us to use standard optimization techniques to find optimal learning rate schedules for a given $\lmat$, batch size $\B$,
and number of steps $T$.

\subsection{Schedule descent}

\label{app:sched_descent}

In general, there are no analytic solutions to the optimal learning rate schedule for non-trivial $\lmat$. There is a limit where
Equation \ref{eq:pvec_high} can be approximated with an ODE, which gives rise to a calculus of variations problem to optimizer the learning rate schedule for lowest
loss after some fixed number of steps. However this problem is intractable when $\lmat$ has more than one unique eigenvalue.

Instead, we can directly optimize the schedule with gradient descent via Equations \ref{eq:pvec_high} and \ref{eq:loss_from_pvec}. More explicitly,
$\pvec_{T}$ is a function of the $T$ dimensional vector $\lr_{t}$, whose elements are the learning rate at each step $t$. Equation \ref{eq:pvec_high}
implies that the loss after $T$ steps can be written as $\Lo(\{\lr_{t}\})$, a loss function of the output of a neural network, whose layers are a quadratic function of $\lr_{t}$. This means that the standard machinery of
deep learning can be used to differentiate through $\lr_{t}$ to do gradient based minimization of $\Lo(\{\lr_{t}\})$.

We carried out a \emph{schedule space gradient descent} (schedule descent for short) algorithm as follows. We initialized by finding a good constant schedule $\lr_{t} = C$
using a logarithmically spaced grid search.

We then carried out standard gradient descent on the objective
$\log(\Lo(\pvec_{T}(\{\lr_{t}\})))$ with a constant learning rate schedule. We found that optimizing $\log(\Lo)$ sped up convergence dramatically without having
to use a (meta) learning rate schedule for the schedule descent.
We also added one additional update rule to schedule descent: if the final loss was above a certain
threshold (set to $10$ in our setup), then instead of taking a gradient descent step, the optimizer multiplies all learning rates uniformly by $0.3$. This helped
us deal with the fact that the optimal schedules are often close to ones that lead to divergent training, at which point no gradient signal can be obtained from the
current schedule.

\begin{figure}[tb]
    \centering
    \begin{subfigure}[b]{0.65\linewidth}
      \includegraphics[width=\linewidth]{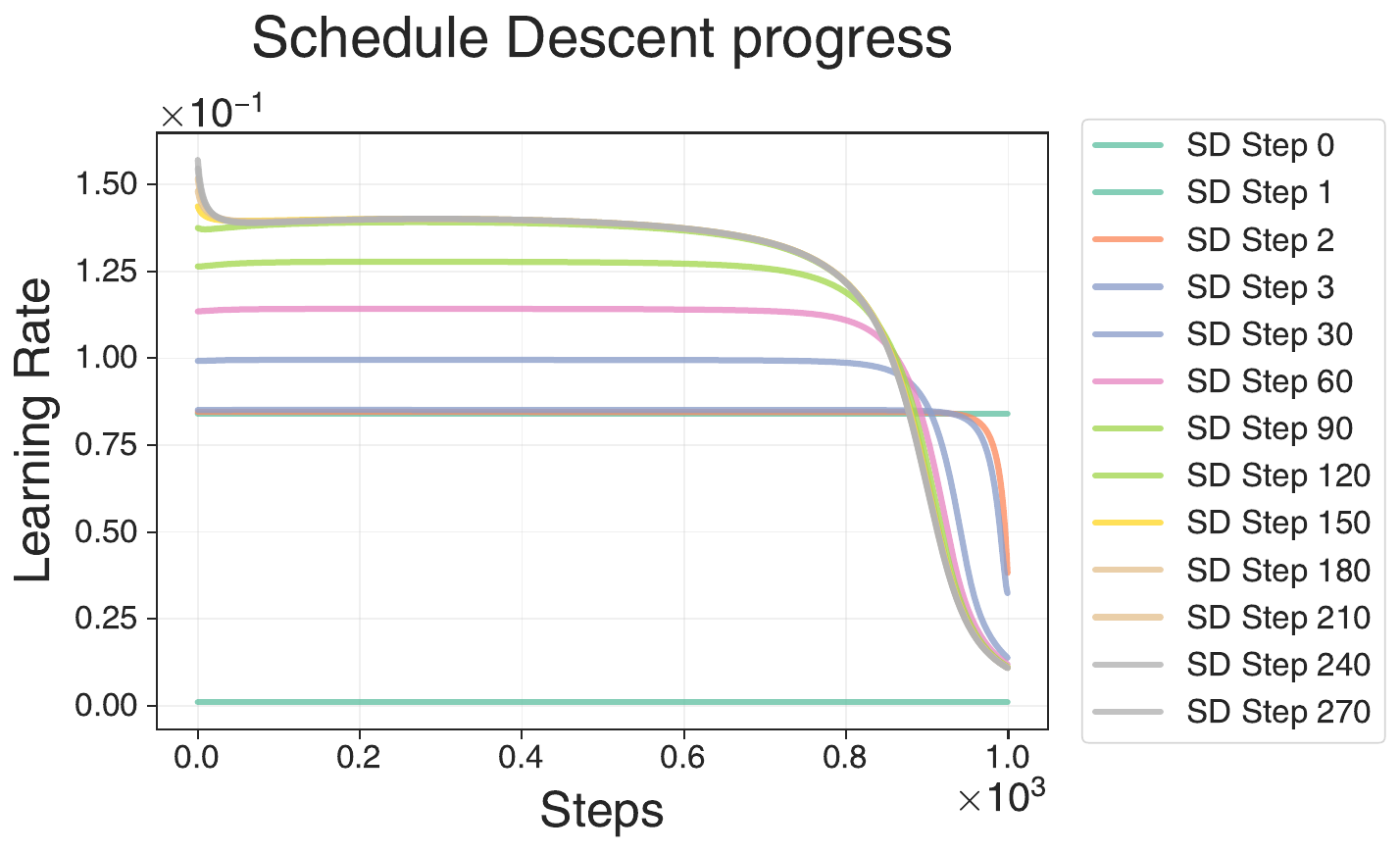}
    \end{subfigure}

    \caption{
    Progress of schedule descent optimizer on linear regression workload. Each curve represents the state of the schedule at different steps of the
    optimization of the loss at the final training point $T = 1000$. The first step finds a good constant learning rate schedule; after that the algorithm runs standard gradient descent with a fallback if the schedule leads to
    diverging training. Final loss, and the corresponding schedule seemed to converge after around $300$ steps. The schedule is represented by a $T$
    dimensional vector of discrete learning rates, continuity was not enforced beyond the initial condition.
    }
    \label{fig:lin_reg_sched_descent}
\end{figure}

We found that the schedule progression was generally smooth (Figure \ref{fig:lin_reg_sched_descent}). Decay at the end quickly developed, which
allowed the median base learning rate to increase as well. It only took around $300$ steps for the loss to stop changing appreciably and for the schedule
shape to converge. One remarkable feature is that the schedule appears to be relatively smooth, even though the $1000$ discrete learning rates were
simultaneously optimized with no restriction. We only enforced ``smoothness'' in the constant initial condition.

We used the schedule obtained from our search as the optimal schedule in our study of linear regression; as we showed in the main text, it indeed had the lowest
average loss out of all schedules we studied.

\subsection{Loss curves from random search}

\label{ref:app:lin_reg_compare}

One interesting feature of the loss curves of the schedules found using our searches in the various families is that many of the best schedules give
non-monotonic loss curves (Figure \ref{fig:ave_traj_lin_reg}). This suggests that many good schedules temporarily exceed the \emph{edge of stability} --- that is, they use learning rates larger
than the largest convergent constant learning rate. This can be beneficial, since exceeding the edge of stability helps the optimizer make more progress on smaller
eigenmodes.

As long as the larger eigenmodes are not unstable for too long, they
can be converged during the learning rate decay phase where the learning rate drops below the edge of stability. Here these large eigenmodes converge more
quickly than the smaller eigenmodes, so progress on the objective is not harmed by the brief foray into instability. This is similar to the fact that in
the non-convex setting, there are stable high frequency oscillations in the large eigendirections for a significant portion of training \citep{cohen2024adaptivegradientmethodsedge}.

One interesting feature of the linear regression workload is that it favors no warmup. This provides another line of evidence that warmup is indeed a useful
feature due to non-convexity and curvature dynamics in deep learning workloads \citep{gilmer2021loss}, and can't be explained with convex
reasoning alone.

\subsection{Theory and experiment comparison}

\begin{figure}[tb]
    \centering
    \begin{subfigure}[b]{0.7\linewidth}
      \includegraphics[width=\linewidth]{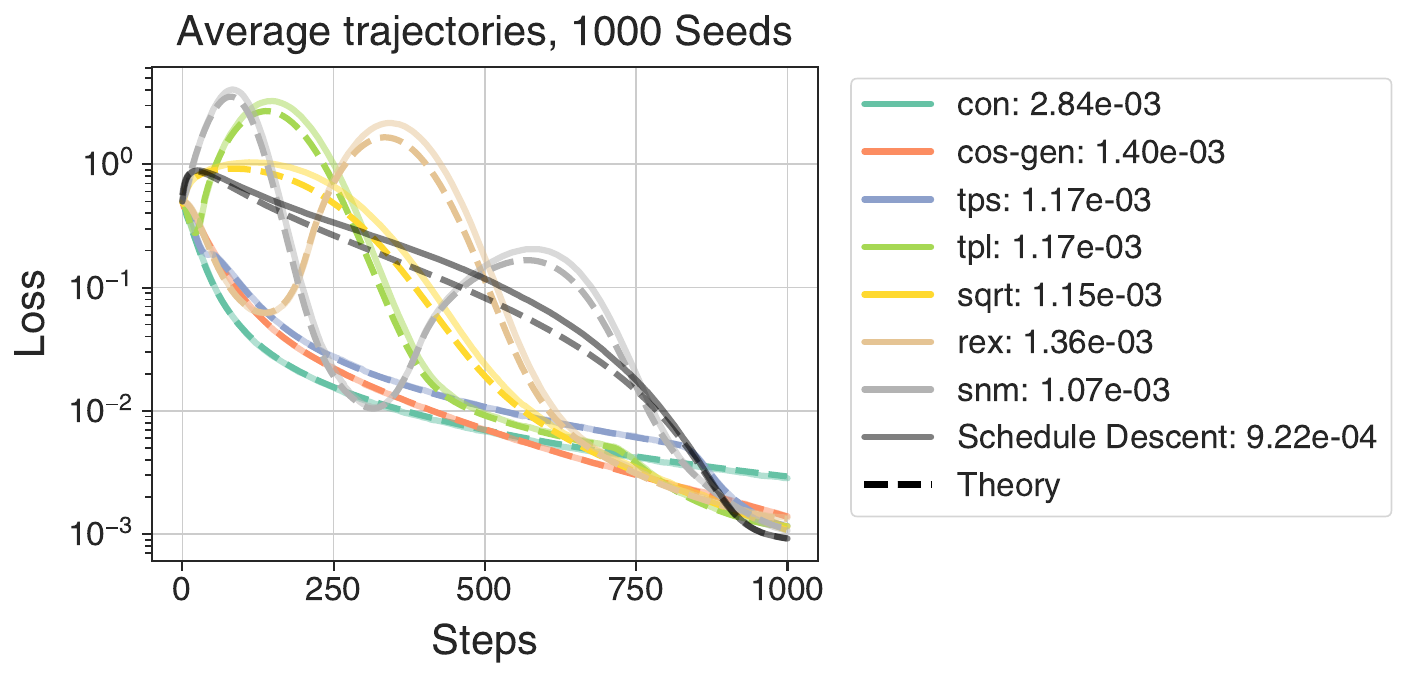}
    \end{subfigure}
    \caption{
    Training loss curves for linear regression workload.
    When averaged over $1000$ seeds, theory predicts early time and late time behavior well, but shows some systematic deviations at intermediate times where dynamics is near/above edge of stability, and loss curves can become non-monotonic.
    }
    \label{fig:ave_traj_lin_reg}
\end{figure}

Given that Equation \ref{eq:pvec_high} represents a high-dimensional limit of the dynamics of the linear regression problem, we can ask: how well does this reflect
the average in practice? To validate our approach, we answered this question by comparing the learning trajectories of the best schedules from each family,
averaged over $1000$ seeds, to the theoretical predictions of the loss for different schedules (Figure \ref{fig:ave_traj_lin_reg}). For all schedules, there
was good agreement between the theory and the empirical averages. For schedules with smaller maximum learning rates (like the constant schedule or $\shortcosgen$,
the match was good through the whole trajectory.

However, the better schedules spent time at or even slightly above the edge of stability, and showed worse agreement with theory at intermediate times. The
theory captured, for example, non-monotonicity in the loss curves, but there were quantitative differences at intermediate times. However these differences vanished
at late times. We speculate that for these trajectories, there were a small number of eigendirections (corresponding to the largest few eigenvalues) which dominated
these parts of the trajectory, which leads to a breakdown of some of the assumptions needed to apply the full high-dimensional version of the theory.
There is indeed a more general version of Equation \ref{eq:loss_from_pvec} which is exact at finite dimensions but not tractable to simulate and optimize as it contains
$\D^{2}$ coupled linear equations \citep{agarwala2024high}.

These discrepancies vanish at late times because the learning rate decay causes the large eigenvalues to converge to near-zero values, compared to the small
eigenvalues which dominate the loss at late times (whose dynamics is well suited to the high-dimensional approximation). We believe simulating the full system
(which requires analysis of a $\D^{2}\times\D^{2}$ matrix) would lead to very minute differences in the optimal schedule shape, none of which affect our
conclusions. We leave detailed analysis of the finite-size deviations to future work.

\subsection{Matching optimal schedule with the \snm family}

\label{app:match_optimal_with_snm}

In Figure \ref{fig:snm_comparison_lin}, we found a member of the \shortsnm family which best-approximated our optimal learning rate schedule from schedule descent.
We found this member by minimizing the MSE loss between the schedule values by optimizing the parameters of the \shortsnm family member.

Because our \shortsnm implementation used a Numpy based code that was not easily convertible to a differentiable JAX form, we ended up implementing an approximate
GD algorithm using sampling. We performed discrete coordinate descent, using small changes in one parameter at a time to give a discrete approximation to the
gradient.

It is likely that there can be small improvements made to the \snm member in terms of minimizing its $L^{2}$ distance to the optimal schedule; however the point still
stands that there exist members of the \snm family which much more closely
approximate the optimal schedule. Indeed it is not even clear that minimizing the $L^{2}$ distance to the optimal
schedule is what minimizes the loss; nonetheless, this new \snm member had significantly lower loss than the schedule found via our search procedure, and was
quantitatively close to the performance of the optimal schedule.

\clearpage
\section{Experimental Details}
\label{app:experimental_details}

In this section, we describe model architecture detail and how each schedule family parameterizes the shape of the schedules used in our study. 
Full implementation details are available at 
\url{https://github.com/google/init2winit/tree/master/init2winit/projects/optlrschedule}.

\subsection{Model Architecture}
\label{app:model_arch}

The \cnn architecture in Table~\ref{tab:cnn_architecture} targets \cifar classification. It begins with a convolutional layer of 32 filters (3x3), followed by ReLU and max pooling operations. A second convolutional layer employs 64 filters (3x3), again using ReLU and max pooling. After pooling, the output is flattened and processed by a dense layer of 256 units with ReLU activation. A final dense layer then outputs logits corresponding to the 10 \cifar classes.

\begin{table}[htbp]
\centering
\caption{Detailed architecture of the \cnn model used for \cifar classification. The model comprises two convolutional blocks, each followed by ReLU activation and max pooling, and two fully connected layers.}
\label{tab:cnn_architecture}
\begin{tabular}{lll}
\toprule
\textbf{Layer} & \textbf{Details} & \textbf{Output Shape} \\
\midrule
Input & Image (32 $\times$ 32 $\times$ 3) & (32, 32, 3) \\
\midrule
Conv2D & 32 filters, kernel size = (3, 3), stride = 1 & (30, 30, 32) \\
Activation & ReLU & (30, 30, 32) \\
Max Pooling & window size = (2, 2), stride = 2 & (15, 15, 32) \\
\midrule
Conv2D & 64 filters, kernel size = (3, 3), stride = 1 & (13, 13, 64) \\
Activation & ReLU & (13, 13, 64) \\
Max Pooling & window size = (2, 2), stride = 2 & (6, 6, 64) \\
\midrule
Flatten & Flatten feature maps to vector & (2304) \\
Dense & Fully connected, 256 units & (256) \\
Activation & ReLU & (256) \\
Dense & Fully connected, 10 units (logits) & (10) \\
\bottomrule
\end{tabular}
\end{table}

\begin{table}[htbp]
\centering
\caption{Architecture of the \transformer language model used for \wikitext training. The model follows a standard decoder-only \transformer structure with multiple self-attention and feedforward layers. }
\label{tab:transformer_architecture}
\begin{tabular}{lll}
\toprule
\textbf{Layer Type} & \textbf{Configuration} & \textbf{Output Shape} \\
\midrule
Input Embedding & Token and position embeddings & (batch, 128, 256) \\
\midrule
Transformer Block (repeated $N$ times) &  & \\
\quad Multi-Head Self-Attention & $h$ heads, each with $d_k$, $d_v$ & (batch, 128, 256) \\
\quad LayerNorm + Residual & Pre-norm & (batch, 128, 256) \\
\quad Feedforward Network & Two linear layers with activation & (batch, 128, 256) \\
\quad LayerNorm + Residual & Pre-norm & (batch, 128, 256) \\
\midrule
Output Projection & Linear layer to vocab size & (batch, 128, 10,000) \\
\bottomrule
\end{tabular}
\end{table}

Table~\ref{tab:transformer_architecture} summarizes the architecture of the \transformer model applied to \wikitext. It consists of a stack of standard decoder-style \transformer blocks. Each block includes multi-head self-attention, layer normalization, and feedforward layers. Token and positional embeddings are used at the input, and final predictions are made through a linear projection to the vocabulary dimension.

\subsection{Schedule Families}
\label{app:schedule_families}

Figure \ref{fig:schedule_families_v2} illustrates representative parameterizations from each schedule family.

The \snm family provides the highest degree of flexibility, parameterizing the schedule shape with seven parameters. It imposes no explicit constraints at the beginning or end of the learning rate schedule, nor does it require warmup or decay periods.

The \tps and \tpl families, each defined by five parameters, enforce linear warmup constraints. \tps schedules rely on spline interpolation after the warmup period, whereas \tpl employs linear interpolation.

The \cosgen family explores the exponent parameter within \cosstd, also incorporating a linear warmup period. \cosstd fixes the exponent at 1, varying only the warmup duration. \rex explores variations of the beta parameter within the REX schedules \citep{chen2022rex}. Finally, the \sqrtdecay schedule applies a square-root decay to the learning rate.

\subsection{Hyperparameter Search Space}
\label{app:hyperparameter_search_space}

This section describes the search spaces for hyperparameters associated with each schedule family and workload. 

For base learning rates, linear regression experiments used $16$ evenly log-spaced values from $[0.01, 1.0]$; all other experiments used $16$ evenly log-spaced values from $[0.001, 0.1]$.
All experiments used AdamW as the optimizer. Unless specified otherwise, weight decay was set to $0$. Section \ref{sec:workload_variations} describes the ablation study in detail.

For each schedule family, the hyperparameters and their respective search ranges are presented in Table \ref{tab:combined-hparams}.

\begin{table}[h!]
\centering
\caption{Hyperparameter search spaces for all schedule families.}
\label{tab:combined-hparams}
\begin{tabular}{lll}
\toprule
\textbf{Schedule Family} & \textbf{Hyperparameter} & \textbf{Search Range} \\
\midrule
\con & warmup\_steps (fraction) & $[0, 0.25]$ \\
\midrule
\cosstd & warmup\_steps (fraction) & $[0, 0.25]$ \\
 & exponent & $1.0$ \\
 & alpha & $[0, 2.0]$ \\
\midrule
\sqrtdecay & warmup\_steps (fraction) & $[0, 0.25]$ \\
 & alpha & $[0, 2.0]$ \\
\midrule
\rex & warmup\_steps (fraction) & $[0, 0.25]$ \\
 & beta & $[10^{-8}, 32]$ \\
\midrule
\tps & x0 & $[0.01, 0.25]$ \\
 & y1 & $[0.1, 1.0]$ \\
 & delta\_x1 & $[0.0, 1.0]$ \\
 & delta\_x2 & $[0.0, 1.0]$ \\
 & delta\_y2 & $[0.0, 1.0]$ \\
\midrule
\tpl & x0 & $[0.01, 0.25]$ \\
 & y1 & $[0.1, 1.0]$ \\
 & delta\_x1 & $[0.0, 1.0]$ \\
 & delta\_x2 & $[0.0, 1.0]$ \\
 & delta\_y2 & $[0.0, 1.0]$ \\
\midrule
\snm & y\_start & $[0.0, 1.0]$ \\
 & y\_end & $[0.0, 1.0]$ \\
 & x\_peak & $[0.0, 1.0]$ \\
 & y1 & $[0.0, 1.0]$ \\
 & delta\_x1 & $[0.0, 1.0]$ \\
 & y2 & $[0.0, 1.0]$ \\
 & delta\_x2 & $[0.0, 1.0]$ \\
\bottomrule
\end{tabular}
\end{table}

\subsection{Noise characteristics of \cifar measurements}

\label{app:noise_character}

\begin{figure}[h!]
    \centering
    \begin{subfigure}[b]{0.24\linewidth}
      \includegraphics[width=\linewidth]{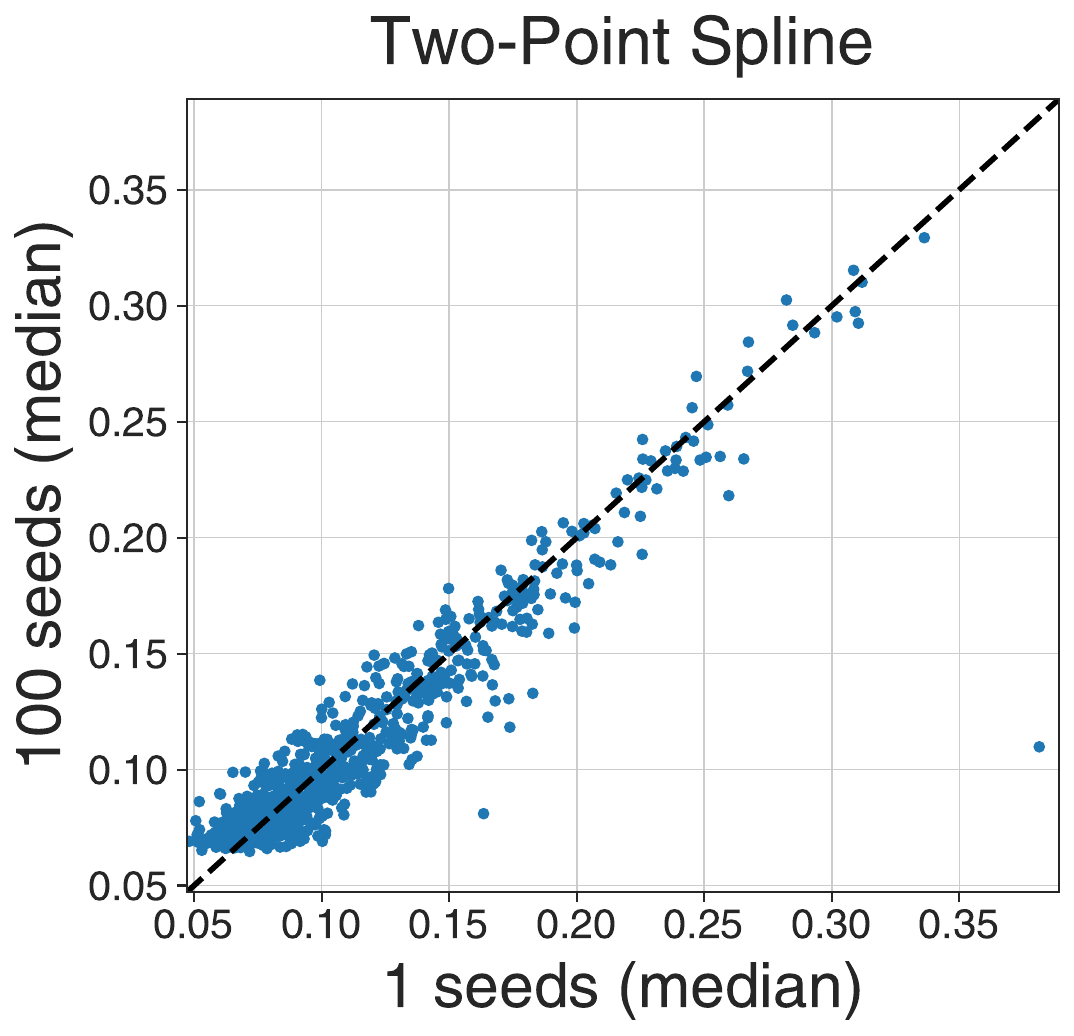}
    \end{subfigure}
    \begin{subfigure}[b]{0.24\linewidth}
      \includegraphics[width=\linewidth]{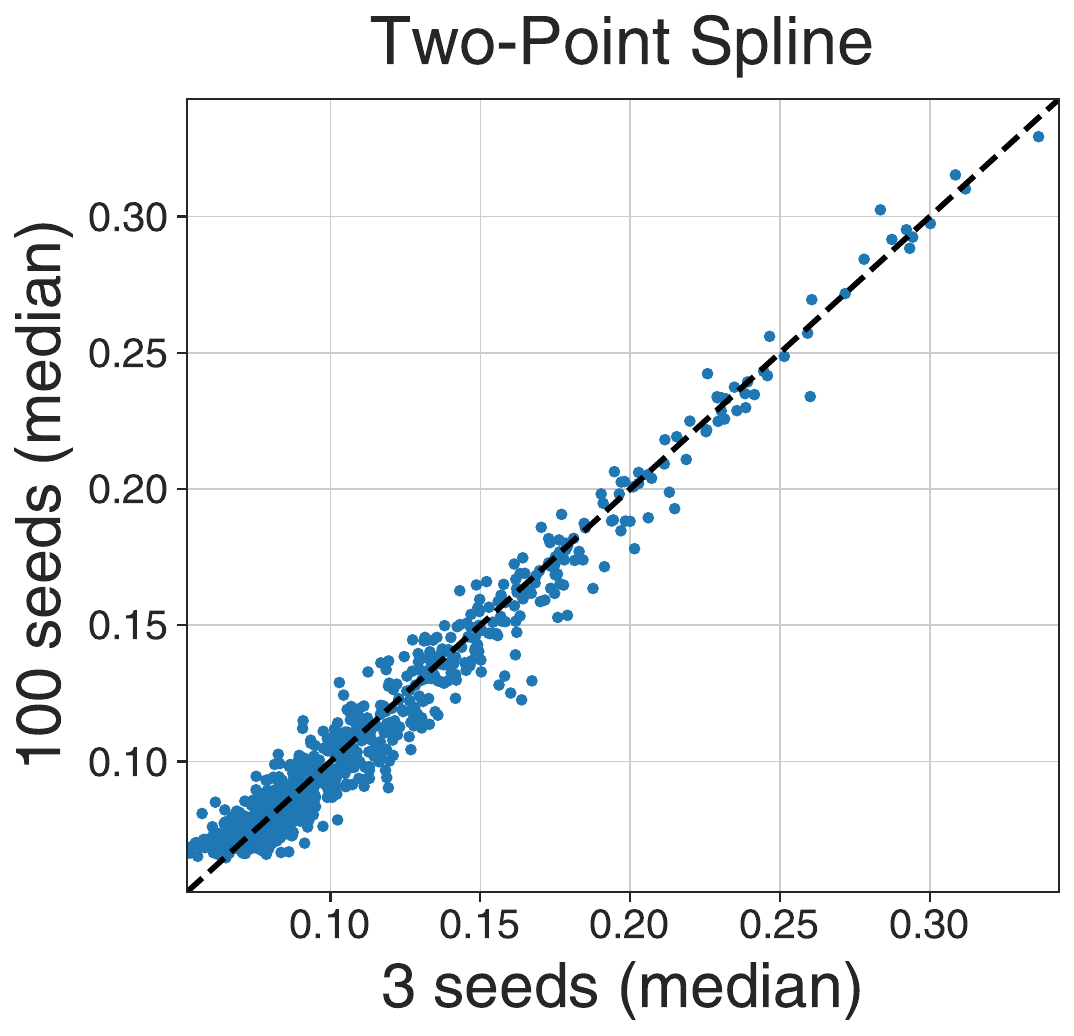}
    \end{subfigure}
    \begin{subfigure}[b]{0.24\linewidth}
      \includegraphics[width=\linewidth]{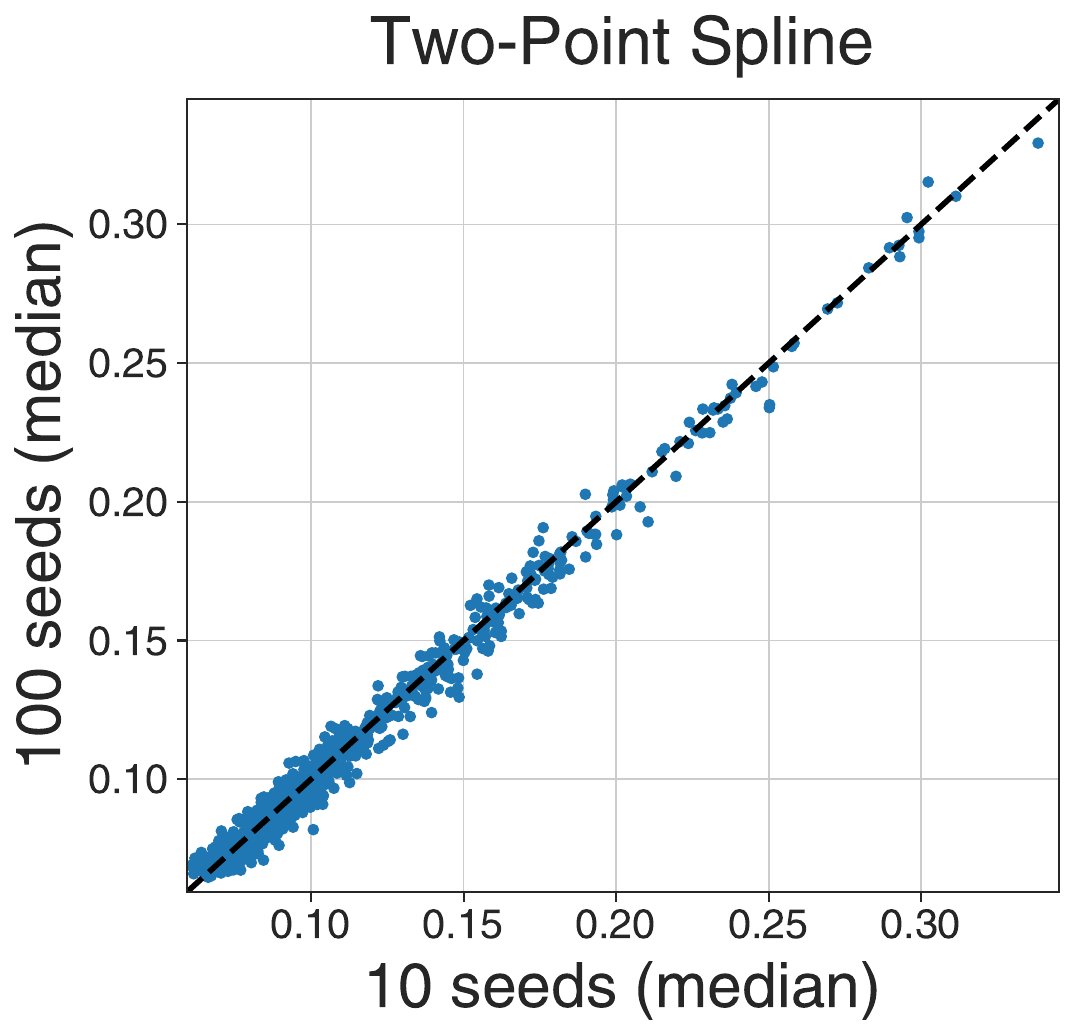}
    \end{subfigure}
    \begin{subfigure}[b]{0.24\linewidth}
      \includegraphics[width=\linewidth]{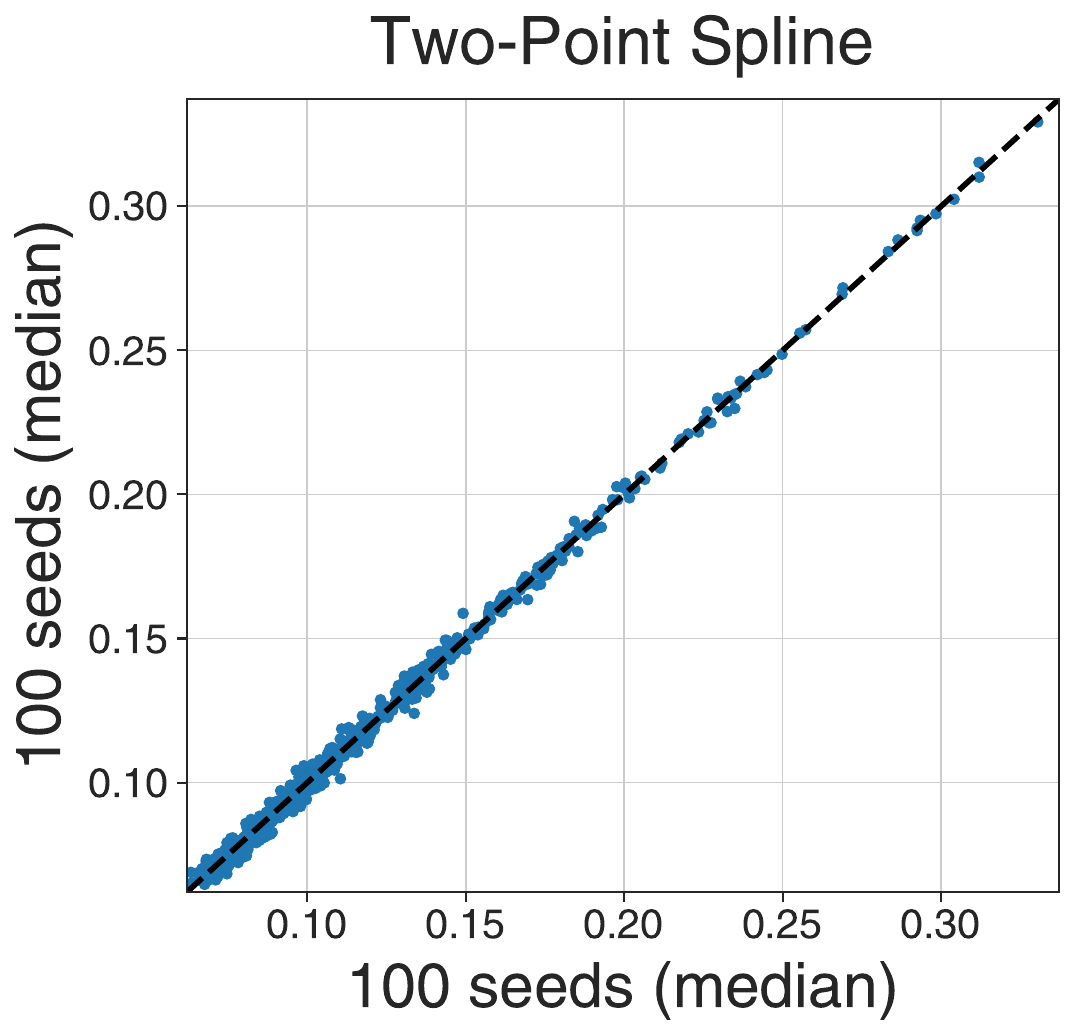}
    \end{subfigure}
    \caption{
    Median \cifar training error under the \tps schedule family at four seed counts. Each plot compares the median from 1, 3, 10, or 100 seeds on the horizontal axis with the reference median from 100 seeds on the vertical axis. A single seed produces wide scatter. Three seeds reduce the spread but still drift. Ten seeds lie close to the diagonal and roughly match the reference, showing that ten seeds give a stable estimate.
    }
    \label{fig:tps_seed_cifar10}
\end{figure}

To choose the number of seeds used in our search and evaluation protocol, we conducted preliminary experiments using the \tps schedule family. This allowed us to test on
a schedule family with a large number of parameters. We trained with two independent sets of
$100$ seeds, and used the median training error of one set as a reference. We then compared this reference to medians computed using subsets from the second set of seeds --- $1$, $3$, $10$, and $100$ seeds (Figure \ref{fig:tps_seed_cifar10}).
We found that there was a high amount of variation between the reference and the $1$ and $3$ seed measurements, particularly at low error. This was an issue because our protocol
requires us to use the small-number-of-seed experiments to suggest candidates for further
measurement. We deemed that the $10$ seed measurements displayed small enough variability to be sufficient for our purposes. In addition, the low discrepancy between the $100$ seed
measurements motivated our use of $100$ seeds for the evaluation step.

\begin{figure}[h]
    \centering
    \begin{subfigure}[b]{0.24\linewidth}
      \includegraphics[width=\linewidth]{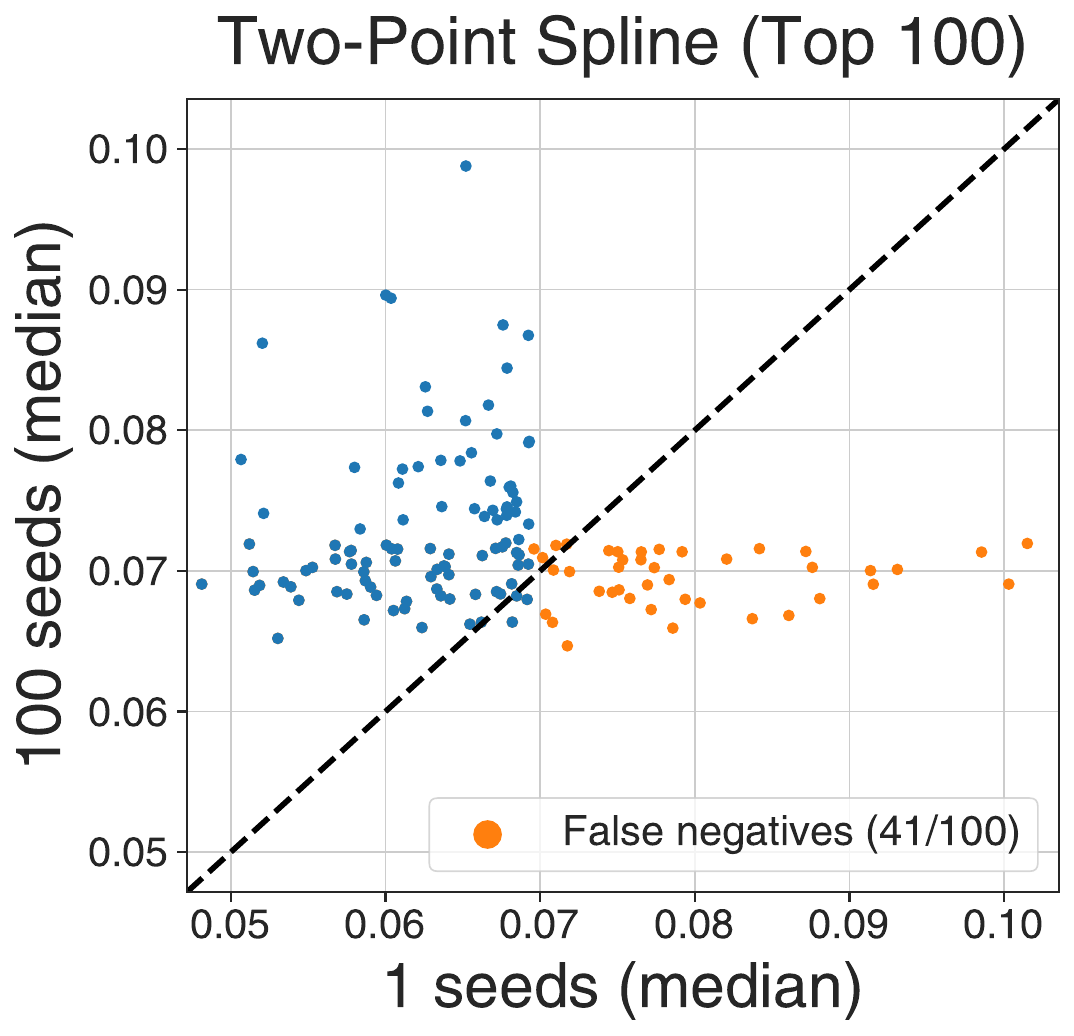}
    \end{subfigure}
    \begin{subfigure}[b]{0.24\linewidth}
      \includegraphics[width=\linewidth]{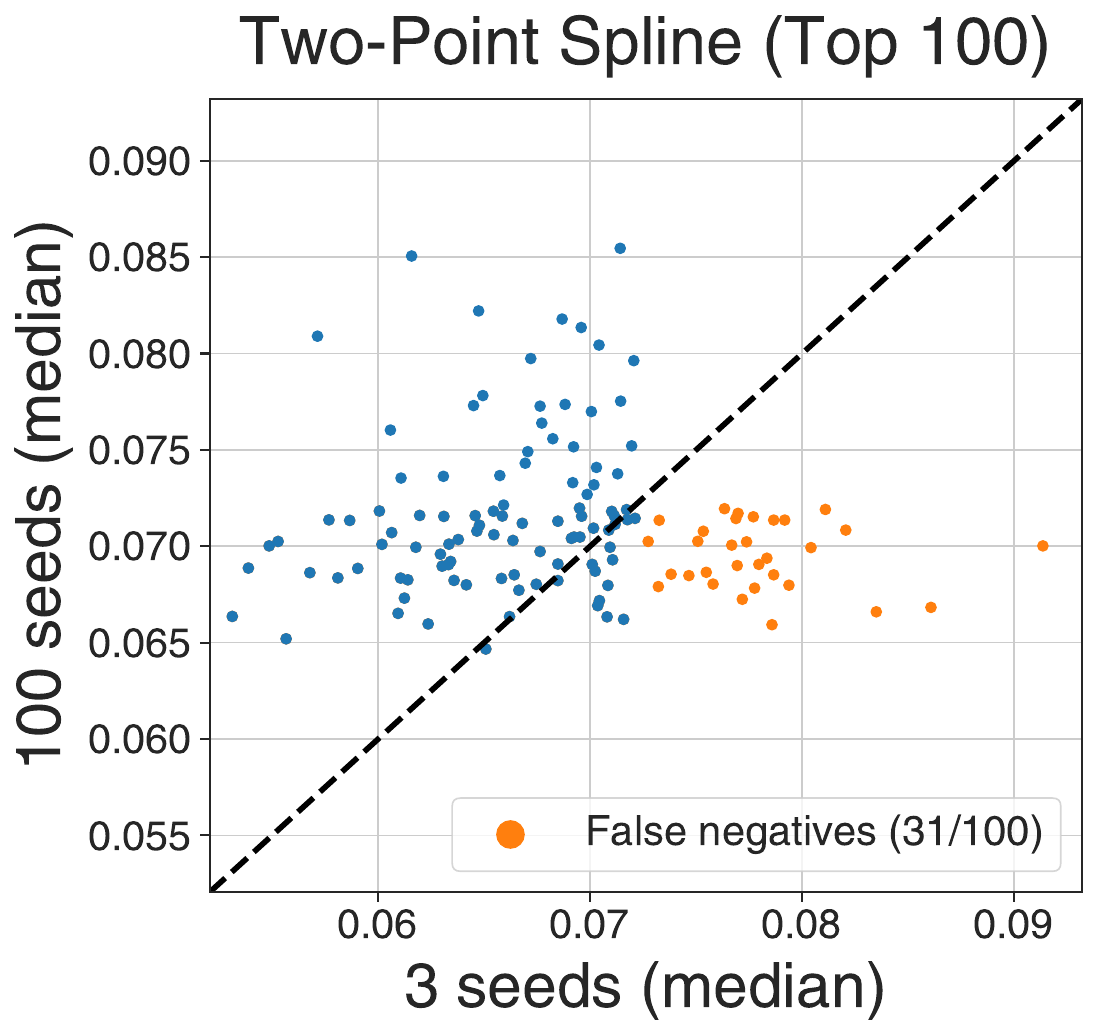}
    \end{subfigure}
    \begin{subfigure}[b]{0.24\linewidth}
      \includegraphics[width=\linewidth]{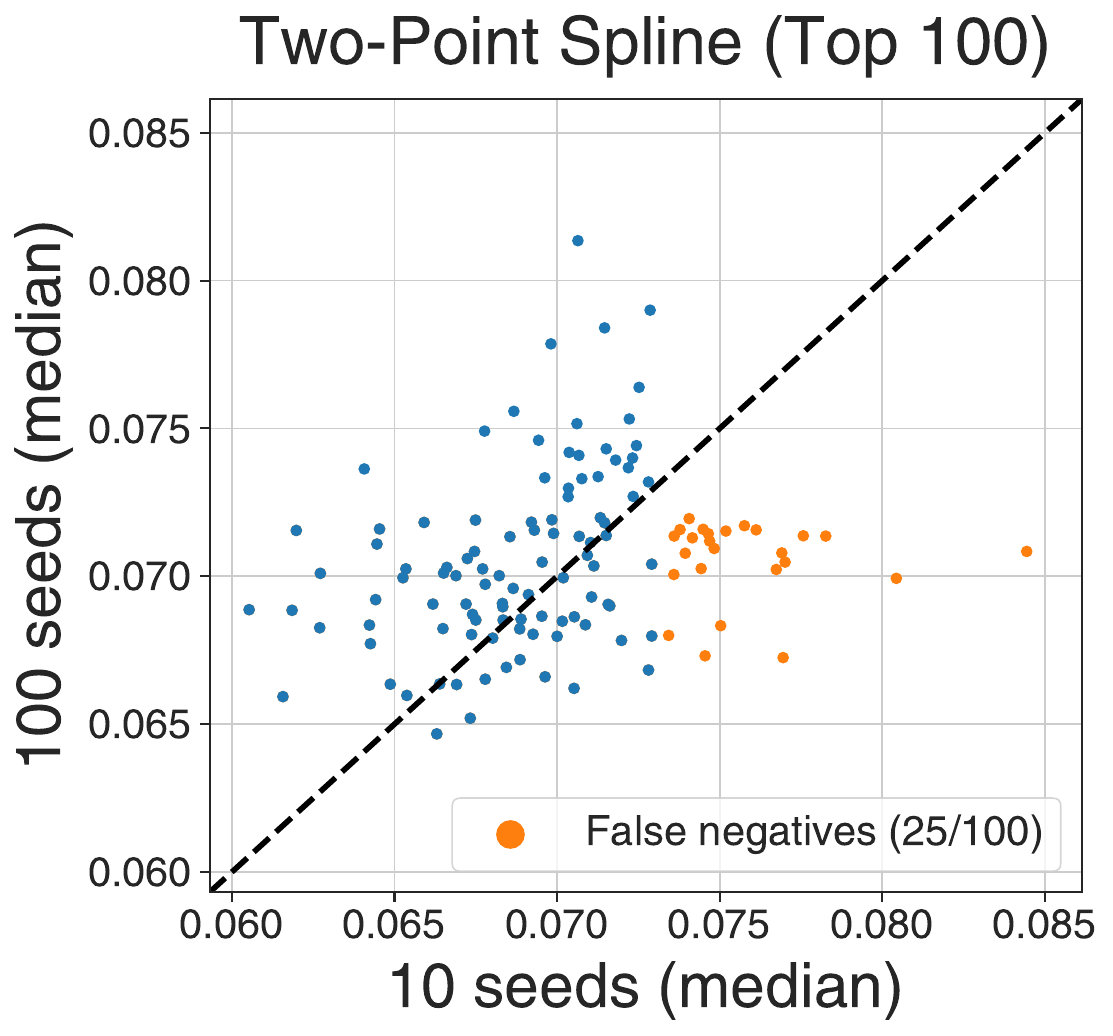}
    \end{subfigure}
    \begin{subfigure}[b]{0.24\linewidth}
      \includegraphics[width=\linewidth]{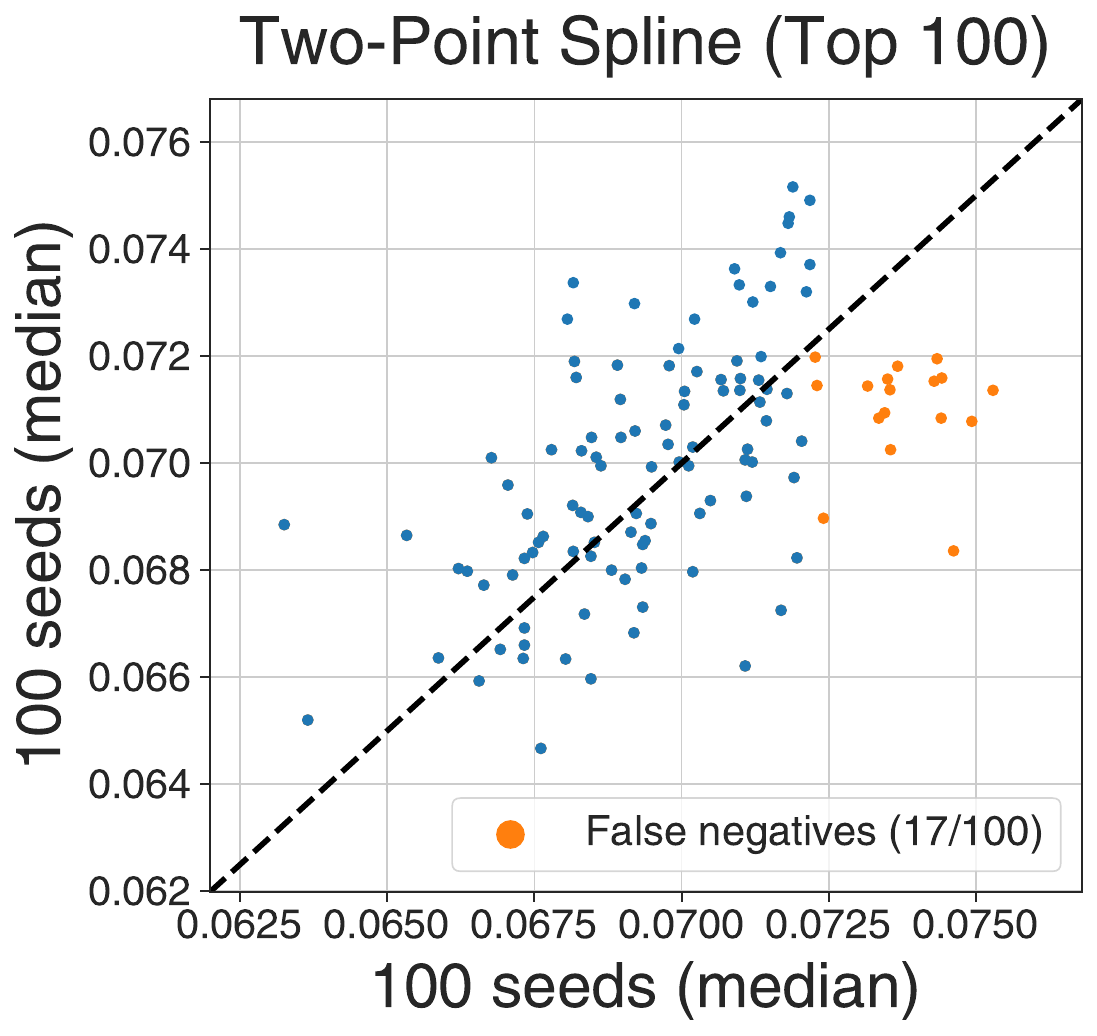}
    \end{subfigure}
    \caption{
    To test our selection procedure, we selected the top $100$ schedules according to the reference measurement ($100$ seed median, $y$-axis) and smaller seed data ($x$-axis, various number of seeds).
    When a single seed was used for measurement, we found that $41$ of the $100$ schedules with lowest error according to the reference were not in the top $100$ of the single-seed measurement (left). In comparison, using a $100$ seed measurement, there were $17$ out of
    the top $100$ references which were not in the top $100$ of the new measurement. We found that using $10$ seeds for the measurement maintained similar false negative rates to
    the $100$ seed measurement, and captures most of the best performing seeds. This motivated our choice of $10$ seeds for the initial phase of the measurement protocol.
    }
    \label{fig:tps_top_100_seed_cifar10}
\end{figure}

We further tested our selection procedure by ranking the top $100$ schedules (by lowest error)
for each metric --- the $100$ seed reference, as well as the $4$ measurements from the other
$100$ seed set. Then we took the union of the top $100$ reference seeds with each of the top
$100$ measurements individually and plotted their errors in each measurement (Figure \ref{fig:tps_top_100_seed_cifar10}).
The resulting measurement allowed us to understand the distribution of top performing schedules in each measurement.
We defined a false negative rate by asking how many seeds ranked in the top $100$ of the reference were \emph{not} in the top $100$ of the second measurement. Comparing the two
independent $100$ seed measurements gave us a baseline false negative rate of $0.17$; this represents an empirical measurement of the baseline level of uncertainty given that
our final evaluations are over $100$ seeds.
We found that
with $1$ measurement seed the false negative rate was $0.41$. We saw that using $10$ seeds for the measurment gave
a false negative rate of $0.25$, which we deemed close enough to the noise floor. Looking at the individual datapoints, we also see that the very best schedules in both measurements
are in the top $100$ for each measure, suggesting that our selection procedure (for this workload and schedule family) is well tuned to balance computational cost with
measurement precision.

\clearpage
\section{Additional results}
\label{app:results}

\subsection{Effects of non-zero final decay value}

\label{app:non_zero_decay}

In all of our schedules except \snm, we set the final learning rate to $0$ at the end of training. Some training suggestions
use a non-zero final learning rate, such as \citet{hoffmann2022training} which uses a cosine learning rate schedule with decay to $10\%$ of the peak
value. In this section, we explore the effects of allowing non-zero decay adding non-zero decay, and show that it doesn't 
change our results much.

We defined two additional schedules: \shortcosy and \shorttpsy which added a free parameter representing the fraction of the peak LR to decay to the \cosstd and \tps schedules respectively.
We then performed the same search procedure as for the other schedule families on both workloads.

\begin{figure}[h]
   \centering
    \begin{subfigure}[b]{0.48\linewidth}
      \includegraphics[height=0.53\linewidth]{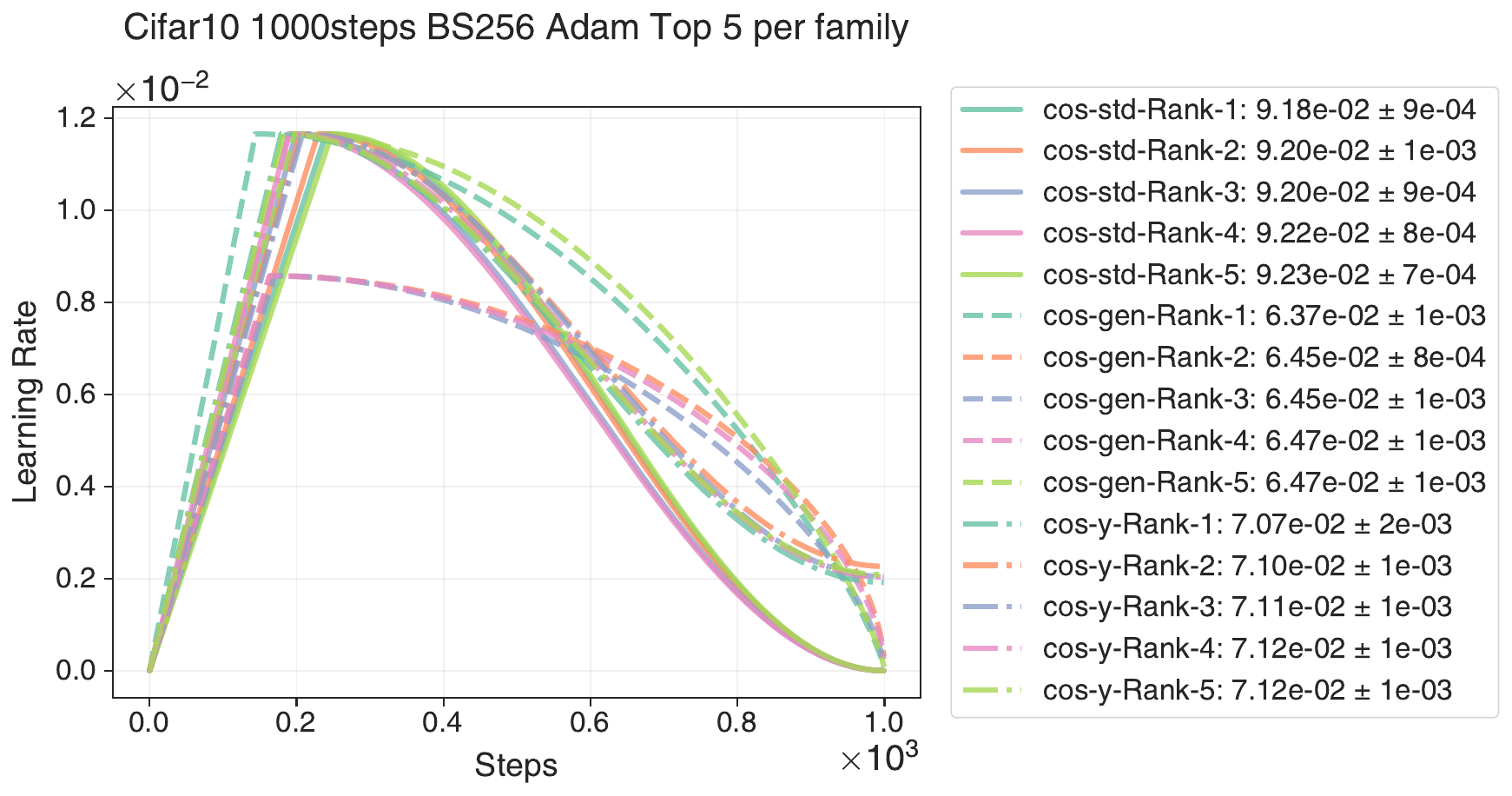}
    \end{subfigure}
    \begin{subfigure}[b]{0.48\linewidth}
      \includegraphics[height=0.53\linewidth]{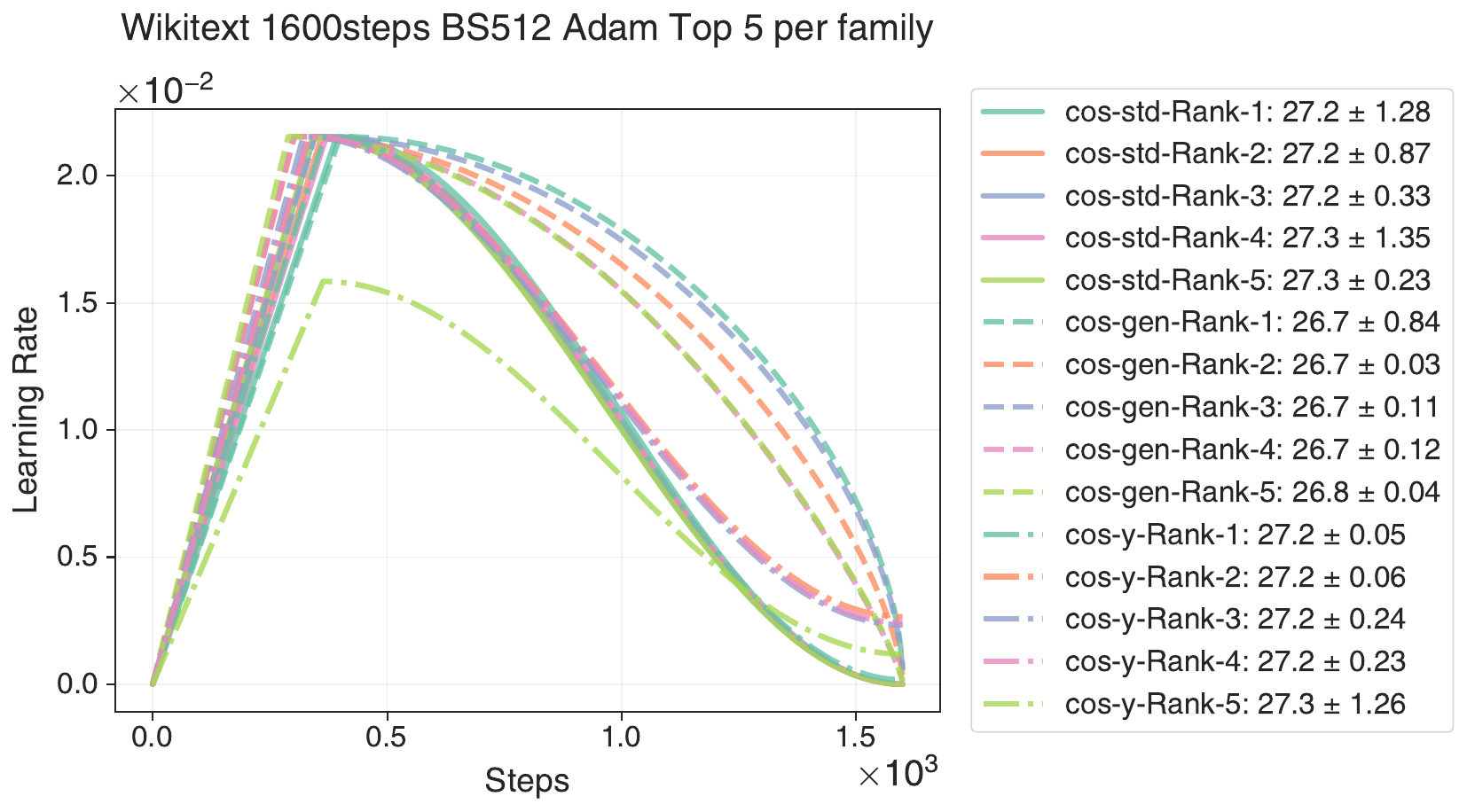}
    \end{subfigure}
    \caption{
     Top-$5$ schedules from search for \shortcosstd, \shortcosgen, and \shortcosy families. For CIFAR10 non-zero final learning rate closes most, but not all of the gap between \shortcosstd and \shortcosgen (left). For wikitext, optimal $y$ value is close to $0$ and there is no gain.
    }
    \label{fig:top_5_cos_y}
\end{figure}

We found that a nonzero final base learning rate improved \shortcosstd on CIFAR10 (Figure \ref{fig:top_5_cos_y}, left), and
closed most (but not all) of the gap with \shortcosgen.
For CIFAR10, \shortcosy had a similar performance to \shortcosstd. This suggests that a tunable final base learning
rate does indeed improve performance over \shortcosstd, but at least on our workloads tuning the power of the cosine
is a better use of an additional parameter than tuning the final base learning rate.

In contrast, for the \tps schedule we found that adding a tunable final base learning rate did not signficantly change
performance. The top $5$ \shorttpsy schedule all had a final decay value near $0$, and consequencyly did not show significant
differences in performance from \shorttps. This suggests that \tps, and most likely, the other schedules with similar final
optimal shapes do not benefit from non-zero final learning rate.

\begin{figure}[h]
   \centering
    \begin{subfigure}[b]{0.48\linewidth}
      \includegraphics[height=0.53\linewidth]{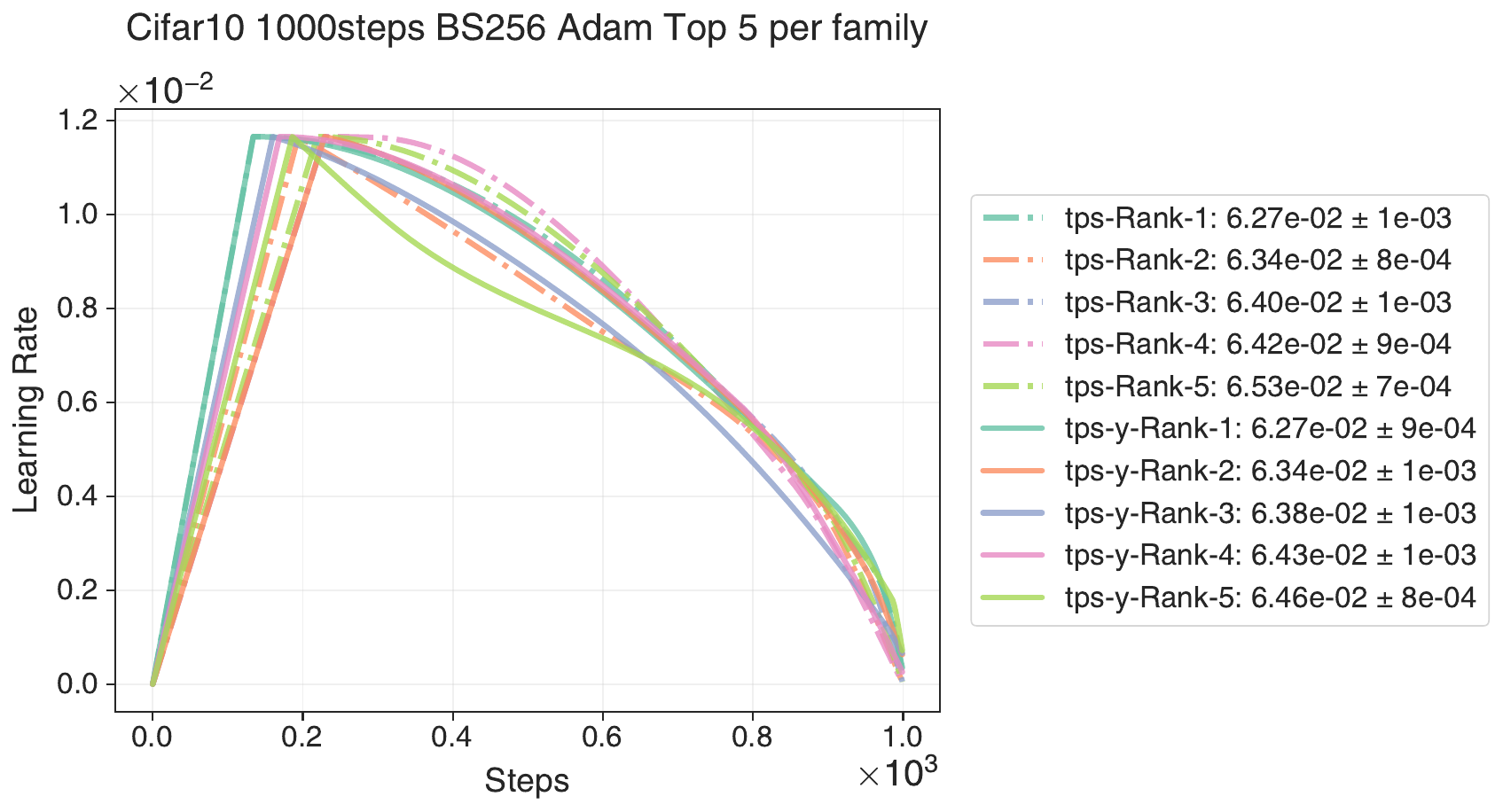}
    \end{subfigure}
    \begin{subfigure}[b]{0.48\linewidth}
      \includegraphics[height=0.53\linewidth]{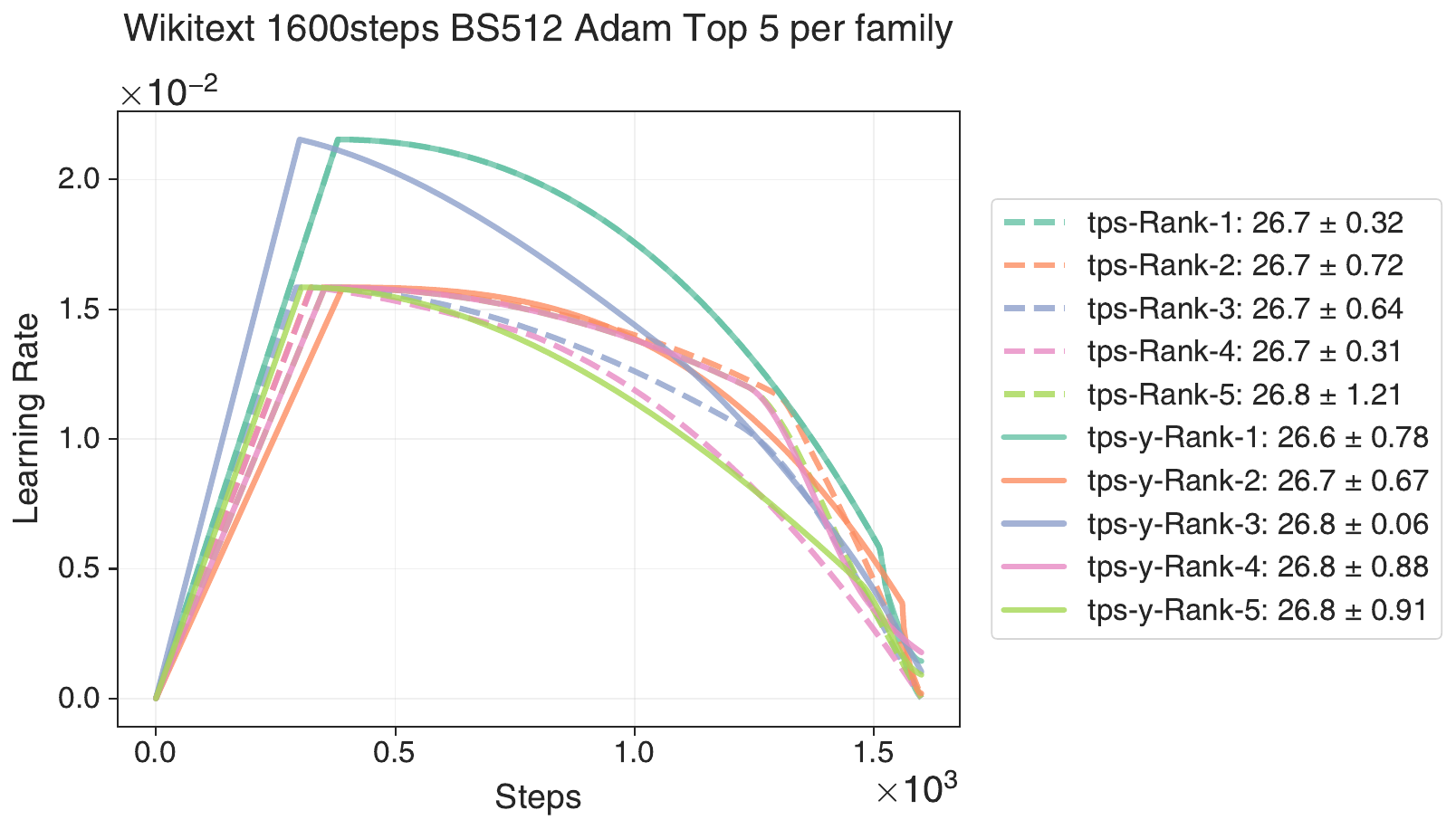}
    \end{subfigure}
    \caption{
     Top-$5$ schedules from search for \shorttps and \shorttpsy families. Optimal final learning rate values are close to $0$ and there are no appreciable differences between the families.
    }
    \label{fig:top_5_tps_y}
\end{figure}

\subsection{\snm with increased samples}

\label{app:snm_ecdf_extra}

\begin{figure}[tb]
    
    \begin{subfigure}[b]{\linewidth}
    \centering
      \includegraphics[width=0.6\linewidth]{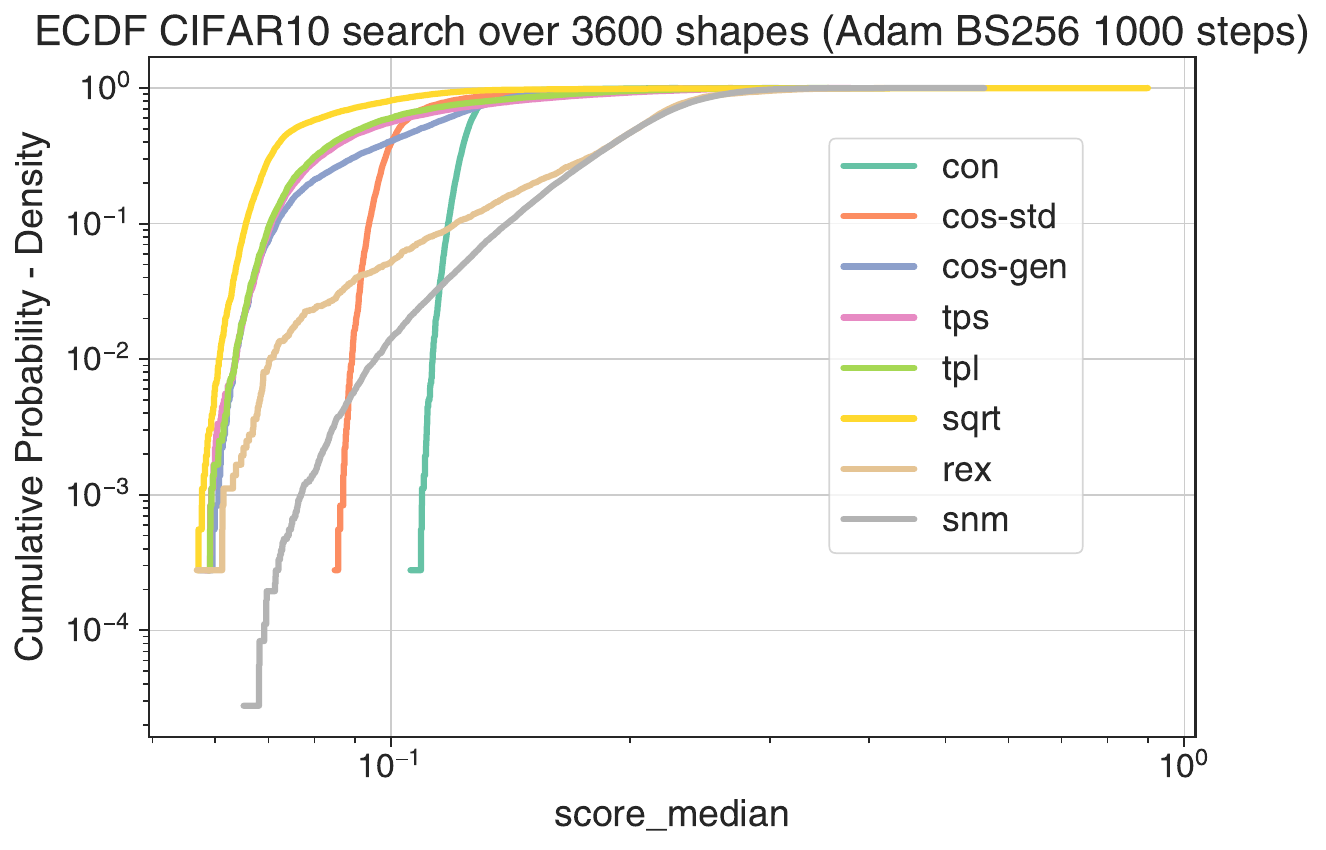}
    \end{subfigure}
    \caption{
    ECDF (Empirical cumulative distribution) of the best score found by random search in each schedule family. The data is the same as Figure \ref{fig:ecdf}, left panel, except
    the \snm family was searched with $36000$ families ($10\times$ more than the rest).
    The extra samples find schedules with better error than the original experiments, but \snm lags behind the other families (higher error). This supports our claim that the \snm
    family was not sufficiently searched/optimized relative to the other families.
    }
    \label{fig:ecdf_36k}
\end{figure}

To confirm the finding that the \snm family is not efficiently searched compared to the other families, we performed an additional search
over $36000$ shapes for \snm in the \cifar workload. Plotting the ECDfs, we found that the additional samples did improve the best schedule found, but only by a small amount,
leaving a gap with the other schedule families (Figure \ref{fig:ecdf_36k}). This suggests that there is room for improvement in the \snm family (as suspected since it is more flexible
than all our other families), but that our search procedure is unable to efficiently probe the search space and requires improvement either of the search space geometry/search distribution,
or needs new techniques alltogether.

\end{document}